\documentclass[default,iicol]{sn-jnl}

\jyear{2024}%

\usepackage{amsthm}
\usepackage{amsmath}
\usepackage{bm}
\usepackage{subcaption}
\usepackage{multirow}
\usepackage{multicol}
\usepackage{booktabs}
\usepackage{fontawesome}
\usepackage{listings}
\usepackage[utf8]{inputenc}
\usepackage{comment}
\usepackage{epsfig}
\usepackage{graphicx}
\newcommand{\cls}[1]{#1}

\usepackage{dblfloatfix}
\usepackage{calc}
\usepackage{tabularx}

\let\originalleft\left
\let\originalright\right
\renewcommand{\left}{\mathopen{}\mathclose\bgroup\originalleft}
\renewcommand{\right}{\aftergroup\egroup\originalright}

\usepackage{amsmath}
\usepackage{amssymb}  

\usepackage{mathtools}  
\usepackage{bm}

\makeatletter
\newcommand{\spx}[1]{%
	\if\relax\detokenize{#1}\relax
	\expandafter\@gobble
	\else
	\expandafter\@firstofone
	\fi
	{^{#1}}%
}
\makeatother

\newcommand{\genericdel}[4]{%
	\ifcase#3\relax
	\ifx#1.\else#1\fi#4\ifx#2.\else#2\fi\or
	\bigl#1#4\bigr#2\or
	\Bigl#1#4\Bigr#2\or
	\biggl#1#4\biggr#2\or
	\Biggl#1#4\Biggr#2\else
	\left#1#4\right#2\fi
}

\let\set\cbr

\usepackage{stmaryrd}  

\usepackage[OMLmathsfit]{isomath}  
\usepackage{upgreek}

\DeclareMathAlphabet{\mathbbmsl}{U}{bbm}{m}{sl}
\DeclareMathAlphabet{\mathbbmb}{U}{bbm}{b}{it}
\DeclareMathAlphabet{\mathbbmssit}{U}{bbmss}{m}{it}

\let\vec\relax
\let\set\relax
\newcommand{\vec}[1]{\bm{#1}}
\newcommand{\set}[1]{\mathbbmsl{#1}}

\newcommand{\nunder}[2][5]{\mathrlap{\mkern\the\numexpr#1/2mu\relax\underline{\phantom{\mathrm{#2}\mkern-#1mu}}}\mathrm{#2}}

\newcommand{\rvarstyle}[1]{\uppercase{#1}}
\newcommand{\rvar}[1]{\rvarstyle{#1}}

\DeclareMathOperator{\softmax}{softmax}

\DeclareMathOperator*{\argmax}{arg\,max}
\let\P\relax
\DeclareMathOperator{\P}{P}
\DeclareMathOperator*{\E}{\mathrm{I\kern-.282em E}}
\DeclareMathOperator*{\D}{\mathrm{I\kern-.282em D}}

\newcommand{\enbbracket}[1]{{\mathinner{\left\llbracket{#1}\right\rrbracket}}}

\let\oldhat\hat
\renewcommand{\hat}[1]{\vphantom{#1}\smash[t]{\oldhat{#1}}}
\let\oldtilde\tilde
\renewcommand{\tilde}[1]{\vphantom{#1}\smash[t]{\oldtilde{#1}}}
\let\oldwidetilde\widetilde
\renewcommand{\widetilde}[1]{\vphantom{#1}\smash[t]{\oldwidetilde{#1}}}

\usepackage[nodisplayskipstretch]{setspace}

\newenvironment{talign*}
{\let\displaystyle\textstyle\csname align*\endcsname}
{\endalign}

\usepackage{dashbox}%

\newcommand{\myds}[1]
  {{\small\texttt{#1}}}
\newcommand{\mycls}[1]
  {{\small\texttt{#1}}}
\newcommand{\mydscls}[2]
  {\myds{#1}:\allowbreak\mycls{#2}} 
\newcommand{\myunicls}[1]
  {\mydscls{uni}{#1}}
\newcommand{\myMcls}[1]
  {\mydscls{M}{#1}}

\newcommand{\mytabsep}{\\[.5em]}
\begin{document}

\title[Multi-domain segmentation]{Weakly supervised training of universal visual concepts\\
for multi-domain semantic segmentation}


\author*{\fnm{Petra} \sur{Bevandić}*}\email{petra.bevandic@fer.hr}
\author{\fnm{Marin} \sur{Oršić}}\email{marin.orsic@gmail.com}
\author{\fnm{Josip} \sur{Šarić}}\email{josip.saric@fer.hr}
\author{\fnm{Ivan} \sur{Grubišić}}\email{ivan.grubisic@fer.hr}
\author{\fnm{Siniša} \sur{Šegvić}}\email{sinisa.segvic@fer.hr}

\affil{\orgdiv{Faculty of Electrical Engineering and Computing}, \orgname{University of Zagreb}, \orgaddress{\street{Unska 3}, \city{Zagreb}, \postcode{10000}, \country{Zagreb}}}


\abstract{Deep supervised models have 
  an unprecedented capacity to absorb 
  large quantities of training data.
  Hence, training on multiple datasets 
  becomes a method of choice
  towards strong generalization in usual scenes
  and graceful performance degradation in edge cases.
  Unfortunately, popular datasets 
  often have discrepant granularities.
  For instance, the Cityscapes road class
  subsumes all driving surfaces,
  while Vistas defines separate classes
  for road markings, manholes etc.
  Furthermore, many datasets have overlapping labels.
  For instance, pickups are labeled as trucks in VIPER,
  cars in Vistas, and vans in ADE20k.
  We address this challenge 
  by considering labels 
  as unions of universal visual concepts.
  This allows seamless and principled learning 
  on multi-domain dataset collections
  without requiring any relabeling effort.
  Our method improves within-dataset 
  and cross-dataset generalization,
  and provides opportunity 
  to learn visual concepts
  which are not separately labeled
  in any of the training datasets.
  Experiments reveal competitive
  or state-of-the-art performance
  on two multi-domain dataset collections
  and on the WildDash 2 benchmark.
}




\maketitle

\section{Introduction}

Semantic segmentation has made
great progress in recent years
across a wide variety of domains \cite{shelhamer17pami,bulo18cvpr, orsic20pr, cheng22cvpr}.
The progress is noticeable
not only in terms of improved 
performance on classic datasets 
\cite{cordts16cvpr, everingham10ijcv},
but also on recent increasingly
complex and challenging benchmarks 
\cite{neuhold17iccv, lin14eccv, gupta2019lvis}.
However, generalization 
beyond the training domain 
remains elusive \cite{zendel18eccv}.
This suggests that particular datasets
introduce unintended pieces of bias
that hampers our performance
in real-world applications.

Dataset bias can be reduced
by training models on multiple datasets \cite{fourure17neucom,lambert20cvpr,kim22eccv}.
This task is straightforward when
datasets share the same taxonomy.
On the other hand,
training on divergent taxonomies
such as Cityscapes and Vistas has to be
specifically
facilitated in some way.
One solution is to choose 
a specific common taxonomy
and to impose it  to all datasets
by relabelling all affected images
\cite{zendel18eccv,lambert20cvpr}.
If the chosen relabeling policy 
keeps only the superset classes, 
the procedure can be 
carried out automatically. 
For instance, fine-grained Vistas classes 
may simply be mapped onto 
more general Cityscapes classes
\cite{kreso20tits}. 
If, however, we wish to
keep the fine-grained classes,
we must relabel  all superclass labels.
For instance, all Cityscapes road pixels 
would have to be disambiguated by hand
into some of the Vistas leaf classes.
A middle-ground solution alleviates 
weaknesses of the two extreme approaches
by preserving only some fine-grained classes
in order to contain the relabeling
effort \cite{lambert20cvpr}.

We address two kinds of inconsistency
between taxonomies: 
discrepant granularity and 
overlapping classes.
Discrepant granularity occurs when 
a class from dataset A corresponds to 
several classes from dataset B  
\cite{liang2018dynamic}.
For instance the class 
\mycls{road} in Cityscapes
is further divided into 8 classes in Vistas:
 \mycls{road}, \mycls{bike\_lane},
 \mycls{crosswalk\_plain}, 
 \mycls{marking\_zebra}, 
 \mycls{marking\_other}, 
 \mycls{manhole}, \mycls{pothole}, and
 \mycls{service\_lane}.
Overlapping classes appear when 
visual concepts get inconsistently
grouped across datasets.
For instance, pickups are
grouped with trucks in VIPER
\cite{richter17iccv},
cars and vans in Vistas
\cite{neuhold17iccv},
and vans in ADE20k
\cite{zhou17cvpr}.
We illustrate these two types 
of labeling inconsistencies 
in Figure \ref{fig:intro}.

\newcommand{\szfi}{.213\columnwidth}
\label{intro}
\begin{figure}[h]
\centering
\begin{tabular}{@{}c@{\,}c@{\,}c@{}}
    \includegraphics[height=\szfi]{
    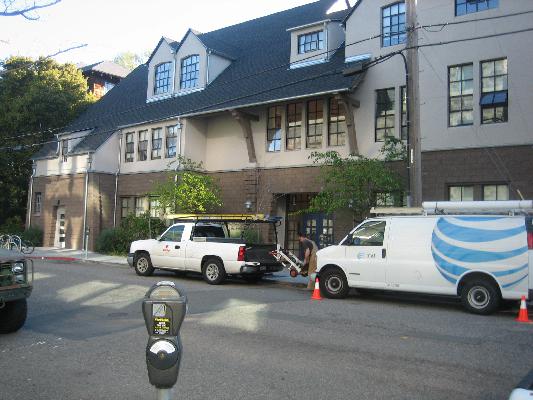}
    &
   \includegraphics[height=\szfi]{
    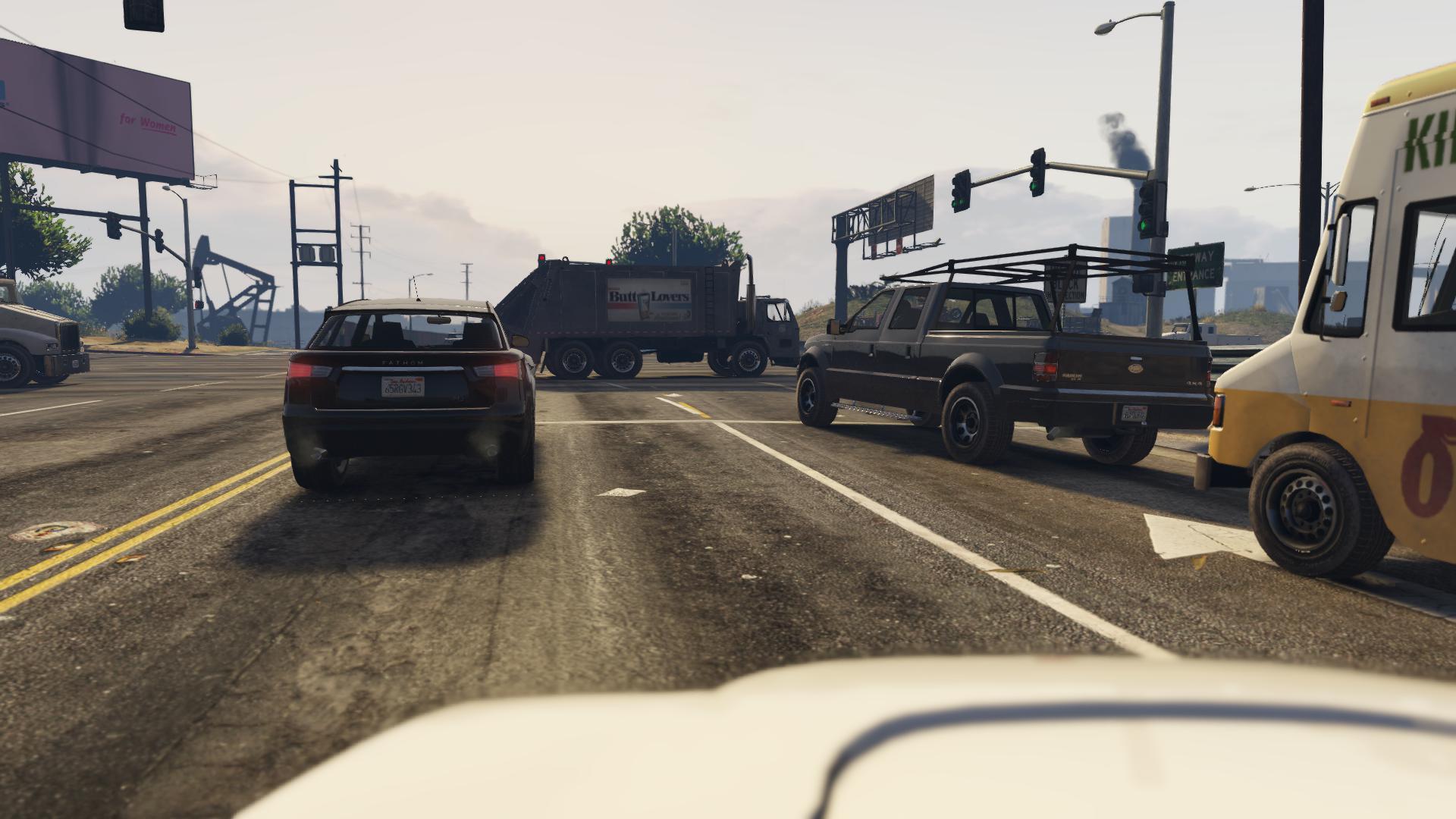}
    &
   \includegraphics[height=\szfi]{
    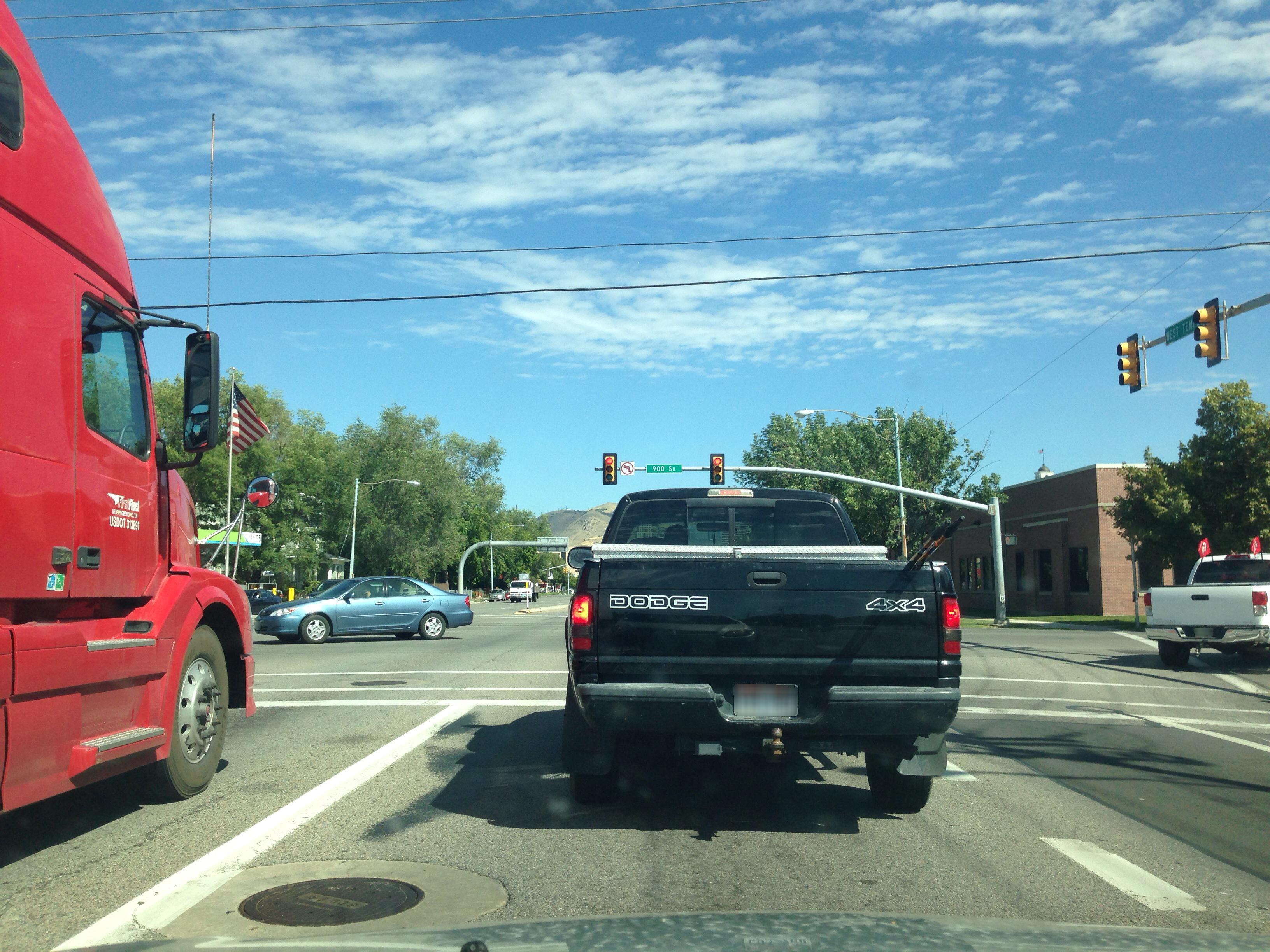}
    \\
        \includegraphics[height=\szfi]{
     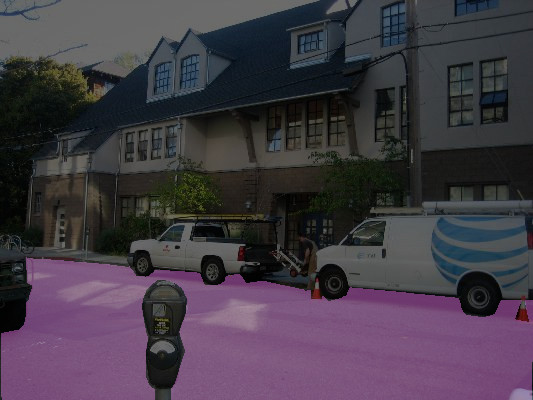}
    &
   \includegraphics[height=\szfi]{
    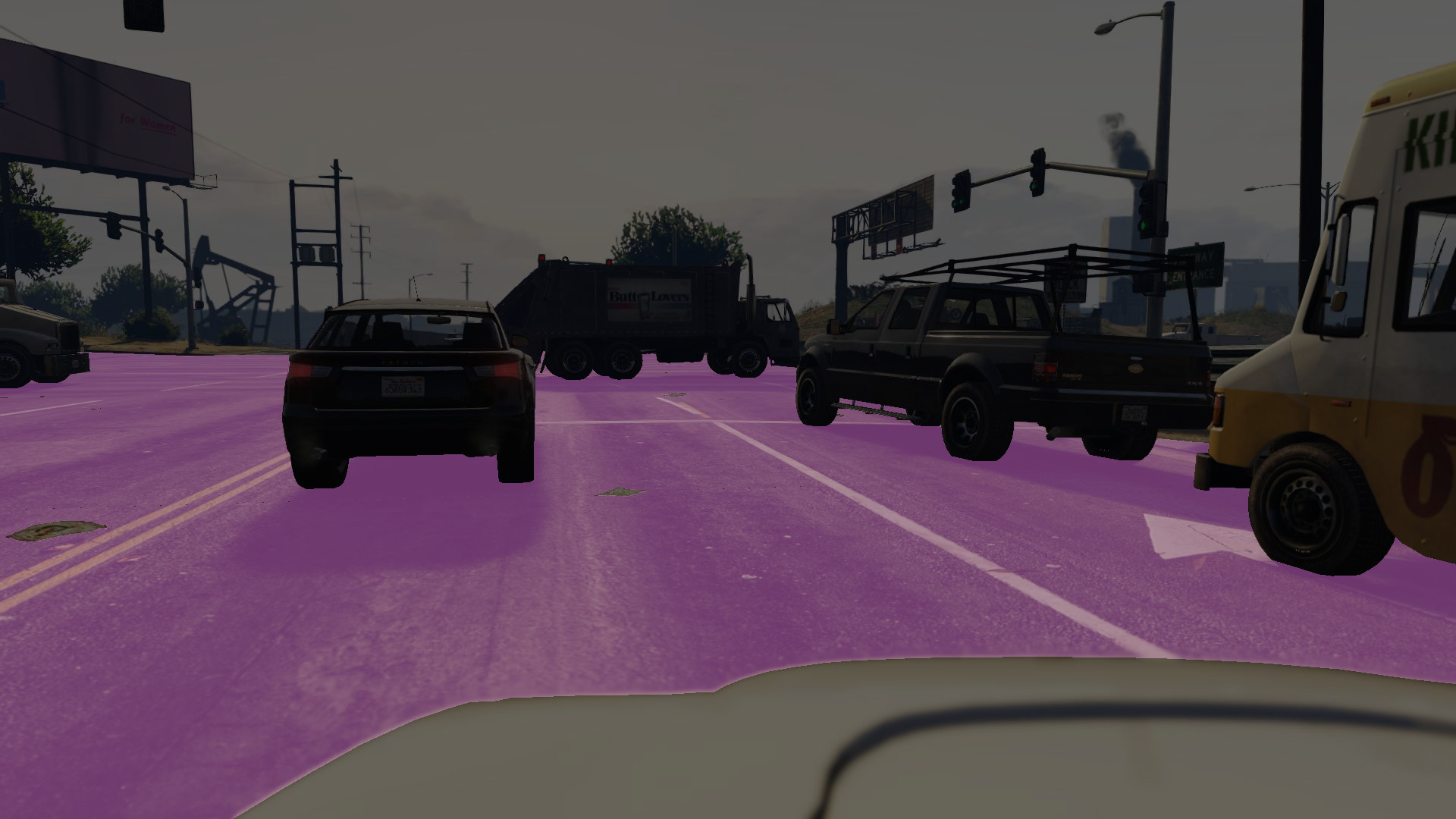}
    &
   \includegraphics[height=\szfi]{
    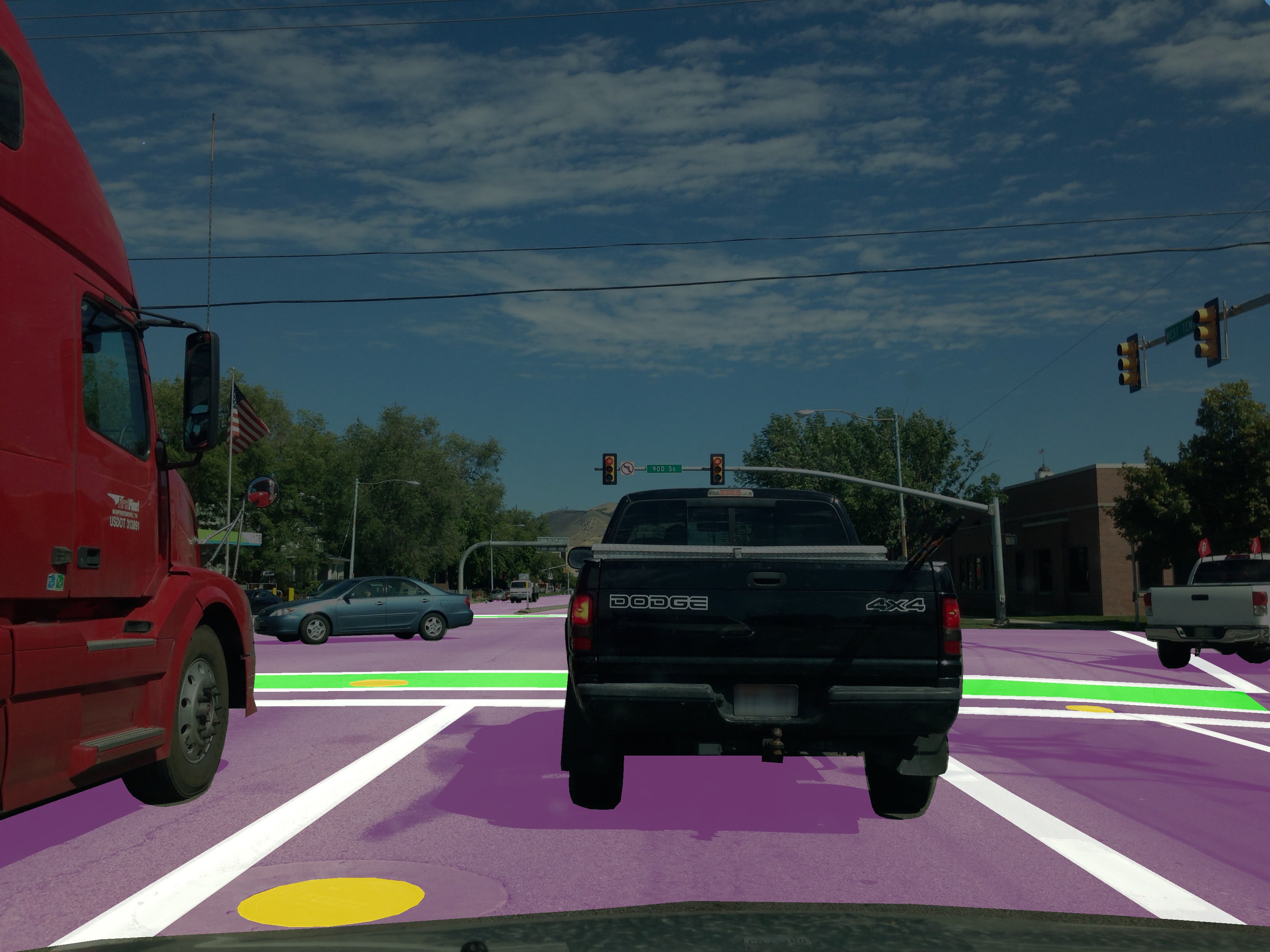}
    \\
        \includegraphics[height=\szfi]{
     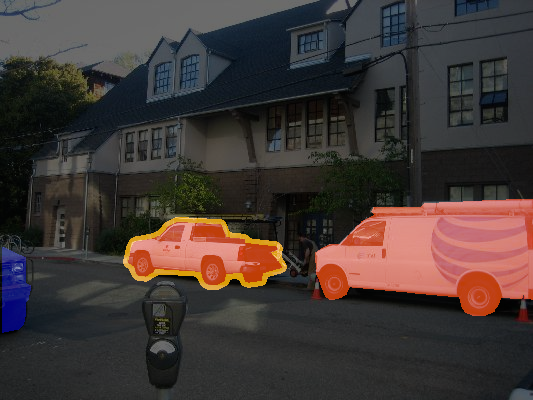}
    &
   \includegraphics[height=\szfi]{
    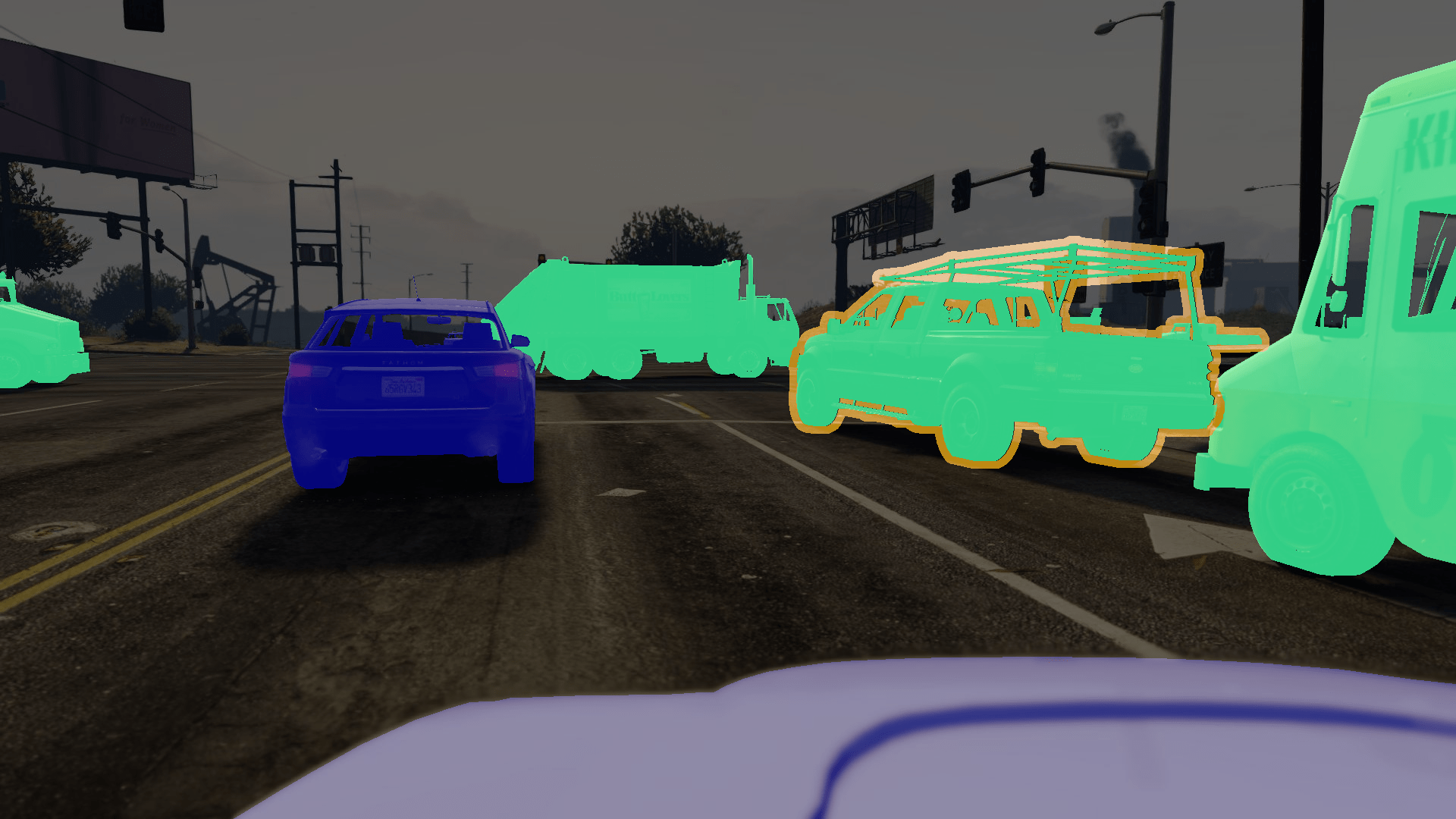}
    &
   \includegraphics[height=\szfi]{
    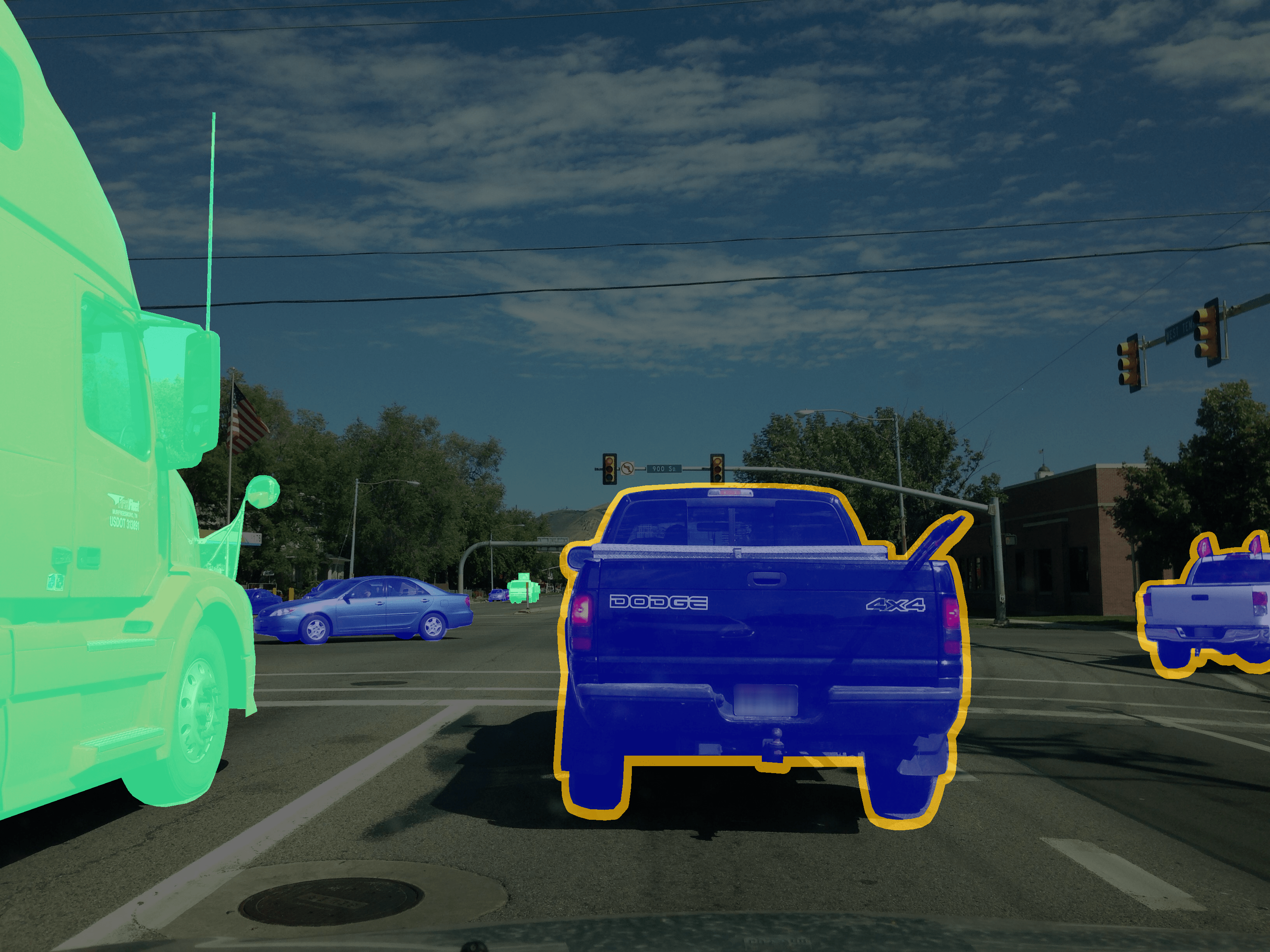}
  \end{tabular}
\caption{
  Common inconsistencies
  among popular datasets. 
  Images from ADE20k, Viper and Vistas (top) 
  are labeled with 
  discrepant granularity:
  class \mycls{road} 
  from ADE20k and VIPER 
  corresponds to 7 classes in Vistas:
  \mycls{marking-other}, 
   \mycls{manhole}, 
   \mycls{crosswalk}, 
   \mycls{road-other} etc
  (middle).
  These datasets have also
  overlapping labels (bottom):
  pickups are labeled as 
  \mycls{truck} in VIPER, 
  \mycls{car} in Vistas, and 
  \mycls{van} in ADE20k.
}
\label{fig:intro}
\end{figure}

We propose a principled method
for training on dataset collections
with inconsistent taxonomies. 
Our method expresses each dataset-specific label
as a union of disjunct visual concepts
which we denote as universal classes.
For instance, we resolve the problem 
from the bottom row of 
Figure \ref{fig:intro}
by defining universal classes
\myunicls{car}, \myunicls{van}, 
\myunicls{pickup} and \myunicls{truck}.
Figure 2 shows that \mydscls{VIPER}{truck}
can be mapped to 
\myunicls{truck} $\cup$ \myunicls{pickup}, 
\mydscls{Vistas}{car} to \myunicls{car} 
  $\cup$ \myunicls{van} 
  $\cup$ \myunicls{pickup},
and \mydscls{ADE20k}{van} to 
\myunicls{van} $\cup$ \myunicls{pickup}. 
Now the posterior probability 
of each dataset-specific label 
can be expressed as sum 
of the corresponding universal posteriors. 
This allows us to train universal models 
against particular taxonomies
through partial labels \cite{cour2011learning}.

Note that our training setup 
does not require any relabeling, 
which means that we do not
need to discard any existing classes.
Therefore, our universal models
capture the full expressiveness
of the dataset collection.
Our method outperforms all previous baselines, 
achieves the best overall result 
on the 2020 Robust Vision Challenge
and delivers competitive performance 
on the WildDash 2 benchmark 
\cite{Zendel_2022_CVPR}.
We encourage comparison 
with related and future 
cross-dataset training approaches
by publishing the source code
\cite{bevandic22github}
for a universal taxonomy 
that unifies a collection 
of ten popular datasets 
with dense groundtruth.

This paper consolidates 
our earlier reports 
\cite{orsic20arxiv,Bevandic_2022_WACV}
and brings several important additions.
First, we formalize the necessary conditions
for a visual concept to be treated
as a learnable universal class.
Second, we construct a universal taxonomy
that spans 10 datasets 
from MSeg and RVC collections.
Third, we extend a recent 
mask-level segmentation architecture 
\cite{cheng22cvpr}
for multi-dataset learning
against our universal taxonomy.
The resulting models deliver 
the first demonstration 
of universal mask-level prediction
in cross-dataset experiments.
Finally, we show that knowledge
of cross-dataset relations 
can improve performance
even when applied only during inference.
This insight can be used as a tool
to disambiguate hypotheses
about mutually inconsistent class relations
during automatic construction 
of universal taxonomies
\cite{bevandic22bmvc}.

Consolidated contributions 
of our work are as follows. 
First, we propose a principled procedure
for recovering a universal taxonomy
for a given dataset collection.
Second, we show that our universal taxonomies
can be learned by different architectures
on original dataset-specific labels
according to the logarithm 
of aggregated probability (NLL+, NLL max).
Finally, our experiments demonstrate
that our forms of weak supervision
deliver state-of-the-art performance,
can learn novel concepts,
as well as perform competitively 
with respect to noiseless learning 
on opportunistically relabeled data.

\section{Related work}

We consider prior work related to 
multi-dataset training 
of semantic segmentation models.
We focus on approaches that can learn
subclass logits from superclass labels.

\subsection{Semantic segmentation}

Standard semantic segmentation delivers
dense categorical predictions 
\cite{long2015fully,chen2017deeplab,zhao2017pyramid,cheng20cvpr}
with respect to the training taxonomy
such as PASCAL VOC \cite{everingham10ijcv} 
or Cityscapes \cite{cordts16cvpr}.
They must retain fine spatial details 
in order to detect small objects,
as well as incorporate 
a large receptive field 
in order to disambiguate 
non-discriminative regions 
through context 
\cite{galleguillos10}.
Furthermore, they should be efficient 
in order to support real-world applications 
and to preserve environment 
\cite{schwartz20cacm}.

Deep convolutional models increase 
receptive field through subsampling
\cite{chen2017deeplab, zhao2017pyramid}.
This strategy also brings improved efficiency 
and reduced training footprint \cite{bulo18cvpr},
however a special care is required 
in order to preserve small objects.
Another way to improve context
awareness is through pyramidal
inputs \cite{zhao18eccv,orsic20pr}.
Some architectures increase 
the receptive field
or improve efficiency through
special kinds of convolutions
\cite{Yu:2016:MCA,sandler18cvpr}.
Recent approaches trade-off efficiency
for improved recognition quality
by relying on transformers
\cite{cheng22cvpr}.

Spatial details can be recovered by 
blending deep layers 
with the shallow ones 
\cite{ronneberger2015u}.
This idea can be carried out 
throughout an efficient upsampling path 
with substantially less capacity 
than the downsampling backbone 
\cite{kreso20tits}.

Semantic segmentation models have usually been trained 
as per-pixel classifiers according to the standard 
cross-entropy loss, which may be inefficient
when dealing with
a large number of classes.
More recent work explores alternative
approaches to semantic segmentation.
For example, non-parametric models which rely on 
prototypes and distance-based classification 
have been shown
to work well on large taxonomies
\cite{zhou22cvpr}.
Improvements have also been achieved
by decoupling segment formation from semantic prediction
\cite{cheng2021nips, cheng22cvpr}.
We demonstrate  the flexibility of our universal taxonomomies
by applying them for cross-dataset training
of the Mask2Former architecture.

Recent work shows that single-dataset semantic segmentation 
may be improved through inclusion 
of hierarchical information training objectives
\cite{li22cvpr}.  Their approach enforces the positive constaints up 
and negative constraints down the class hierarchy. 
Similar to their approach, our method enforces
positive constraints down the hierarchy tree.
Unlike their approach, our method focuses on
cross-dataset training with non-leaf labels.

\subsection{Multi-dataset training}

Deep models can absorb huge quantities 
of labeled data \cite{sun17iccv}.
However, annotated datasets 
for dense recognition 
are quite scarce due to being 
very expensive to procure 
\cite{zlateski18cvpr}.
Hence, multi-dataset training 
becomes a prominent avenue 
towards improved generalization 
\cite{rvc22www}.
Multi-dataset training also improves 
the supervision quality
due to discouraging exploitation 
of dataset bias \cite{zendel18eccv}
and facilitating multi-domain inference 
\cite{lambert20cvpr}.
Thus, the inclusion of a negative domain 
into a specialized supervised setup 
may dramatically improve 
open-set performance  
\cite{chan21iccv, biase21cvpr, BEVANDIC22ivc}.

Simple multi-dataset training 
disregards relations between 
particular taxonomies 
by generating predictions 
with per-dataset heads 
\cite{fourure17neucom,kalluri19iccv,masaki21itsc}.
These approaches can be useful 
for pretraining and further finetuning.
However, they offer limited value 
in realistic applications 
due to being unable 
to deliver unified predictions.
This problem can be addressed 
by enforcing a consistent common taxonomy 
through manual relabeling 
across all datasets
\cite{lambert20cvpr, Zendel_2022_CVPR}.
However, manual relabeling 
is tedious and error-prone.
Moreover, future extensions 
would require further manual relabeling
both in existing and the new datasets.

Relabeling can be avoided 
by detecting relationships 
between particular taxonomies.
Several approaches propose automatic
merging of equivalent classes. 
This can be carried out 
either through optimization 
\cite{zhou2021simple}, 
or by merging classes with identical names 
\cite{lambert20cvpr,kim22eccv}.
However, these approaches are unable 
to exploit subset/superset relationships.
They, therefore, lead to 
implicit dataset detection
and model overfitting
in the presence of overlapping logits.

Very recent work proposes to learn
superclass logits on subclass labels
in a multi-label segmentation setup
\cite{kim22eccv}.
However, their work can not learn 
universal logits from superset labels.
This problem has been addressed 
by proposing prediction heads 
with distinct nodes 
for categories and classes 
\cite{liang2018dynamic, meletis18iv}.
However, hierarchical taxonomies 
complicate model training 
while failing to offer 
substantial advantages 
over flat universal taxonomies
such as the one proposed in this paper.


\subsection{Learning with partial labels}

Partial labels are a 
form of weak supervision
where a training label 
corresponds to
a set of classes
such that only one of the 
classes is correct.
Early work considers 
the problem of 
face identification 
in movie scenes
where partial labels can be
automatically extracted 
from screenplays 
\cite{cour2011learning}.
In this case,
a partial label
corresponds to the 
set of all characters 
present in the scene.
In our setup, 
partial labels are determined 
by semantic relations 
between the universal taxonomy
and the taxonomies 
of particular datasets.
For example,
cars and vans in
Cityscapes are labeled 
with the same class \mycls{car},
but our universal taxonomy
can discriminate 
between the two.
Hence, we consider
the Cityscapes car labels
as partial labels corresponding
to the set of universal classes
\{\myunicls{car}, \myunicls{van}\}.
The learning objective
can be formulated
by aggregating predictions 
that correspond to partial labels.
Recent work considers aggregation
through sum and max functions
but does not find them competitive 
for object detection \cite{zhao20eccv}. 
Log-sum-prob loss has been used 
to alleviate influence 
of labeling noise 
at semantic borders
\cite{zhu2019improving}.
Their method assumes that
the pixels are partially labeled
with classes found
in a $3\times3$ neighborhood.

\subsection{Learning on pseudo-labels}

Pseudo-labeling is a form of 
semi-supervised learning 
where a model trained on labeled data 
provides supervision for unlabeled data
\cite{mcclosky06naacl, lee13wrepl}.
Pseudo-labeling can be viewed 
as an alternative 
to weakly supervised learning
with respect to a given universal taxonomy. 
Recent work pseudo-labels 
object locations in images  
with the predictions of the heads 
trained on other datasets 
\cite{zhao20eccv}.
However, their setup assumes 
that the label spaces
of different datasets do not overlap.
Unlike them, we consider semantic segmentation
and datasets with overlapping classes.

Very recent work frames
universal semantic segmentation
as a regression problem
towards pseudo-labels provided by 
a pre-trained language model
\cite{yin22arxiv}. 
However, label semantics 
may vary across datasets. 
Indeed, another recent work 
finds that visual cues 
give rise to better 
cross-dataset semantic relations
than language-based representations
\cite{uijlings22eccv}.
Another recent work recovers
complex subset/superset relations
that would be difficult to find
with language embeddings
\cite{bevandic22bmvc}.

\section{Method}
\label{sec:method}

This section describes our approach 
for multi-dataset training
of universal visual concepts.
We introduce universal models,
propose a procedure 
for constructing a universal taxonomy 
over a given dataset collection,
formulate a suitable weakly supervised objective,
implement our method within the mask-level context, 
and discuss removal of untrainable classes. 
Most of our considerations are applicable 
in any categorical recognition context.
Still, we focus on dense prediction 
where our method has most practical value
due to huge cost of ground-truth annotations.

\subsection{Terminology and notation}

We use the following 
terminology and notations:
\begin{itemize}
\item 
  we typeset dataset-specific classes 
  in typewriter font as 
  \mydscls{Dataset}{class};
  we abbreviate Cityscapes as City
  and WildDash2 as WD;
\item 
  we consider a semantic class $c$ 
  as a set of all image pixels that 
  should be annotated as $c$;
\item 
  we express semantic relationships 
  according to set notation, e.g.:
  \mydscls{Vistas}{sky} = 
  \mydscls{City}{sky},
  \mydscls{City}{road} $\supset$ 
  \mydscls{Vistas}{manhole},
  \mydscls{VIPER}{truck} $\cap$ 
   \mydscls{Vistas}{car} = 
   \mydscls{WD}{pickup},
  \mydscls{City}{road} $\perp$ 
  \mydscls{City}{car} $\Longleftrightarrow$ 
   \mydscls{City}{road} $\cap$ 
   \mydscls{City}{car} = $\emptyset$;
\item 
  a taxonomy $\set S$ is a set of
  mutually disjoint semantic classes:
  $\forall c_i,c_j\in\set S
    \colon 
    c_i \perp c_j$;  
\item
  a universal taxonomy $\set U$
  encompasses the entire semantic range
  of the considered dataset collection:
  $\bigcup_{\cls u\in\set U} \cls u = 
   \bigcup_d \set S_d$;
  each universal class can intersect
  at most one class from each dataset:
  $\forall 
   \cls u \in \set U,
   \cls c \in \bigcup_d \set S_d
   \colon
  (\cls u \perp \cls c) 
   \vee
  (\cls u \subseteq \cls c)$;
\item 
  a union of taxonomies
  is a pseudo-taxonomy if its members
  have non-empty intersections:
  ${\set S}_\text{VIPER}
    \cup {\set S}_\text{Vistas}$
  is a pseudo-taxonomy since 
  \mydscls{VIPER}{truck} $\not\perp$ 
  \mydscls{Vistas}{car}
  (cf.~Fig.~\ref{fig:intro});
\item 
  a semantic segmentation dataset $\set D_d$ 
  consists of images and corresponding 
  dense labels: 
  $\set D_d = \{(\vec x^d, \vec y^d) \}$;
  the labels correspond 
  to semantic classes $c \in \set S_d$,
  where $\set S_d$ is the taxonomy
  of the dataset $\set D_d$.
\end{itemize}

\subsection{Universal models}

\begin{figure}[htb]
 \centering
 \includegraphics[width=0.45\textwidth]
 {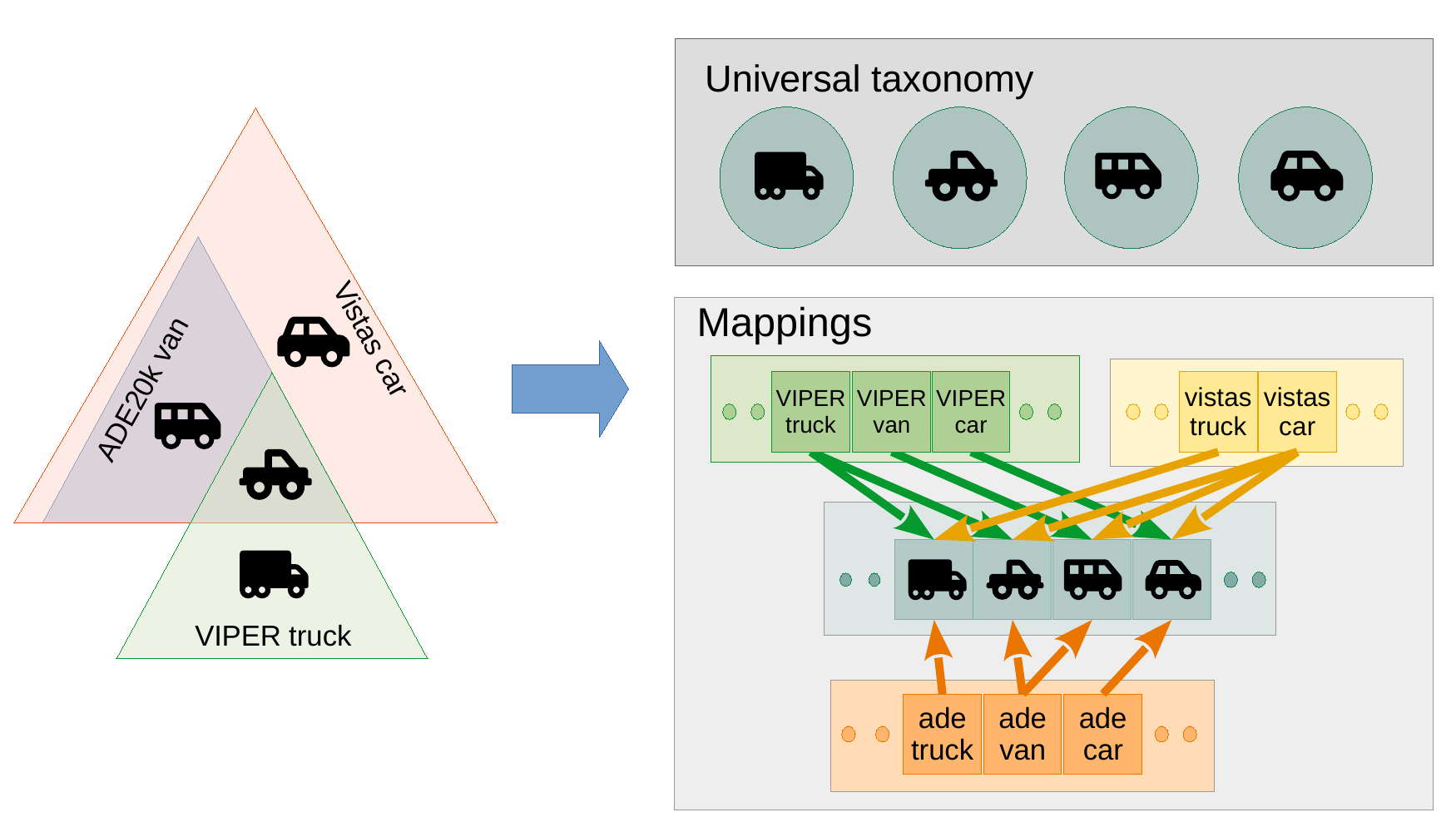}
 \caption{%
   We address the semantic overlaps
   from the bottom row 
   of Fig.\ \ref{fig:intro} 
   by mapping each label (left) to a set 
   of disjoint universal classes (right).
   Our universal taxonomy spans
   the entire semantic range
   of the considered dataset collection.
 }
 \label{fig:approach-taxonomy}
\end{figure}

We consider training a 
flat universal model
on multiple datasets with
incompatible labeling policies.
Our universal taxonomies ensure 
that each dataset-specific class 
can be mapped to a union 
of disjoint universal classes
as illustrated in Figure 
\ref{fig:approach-taxonomy}.

\begin{figure}[htb]
 \centering
 \includegraphics[width=0.45\textwidth]
  {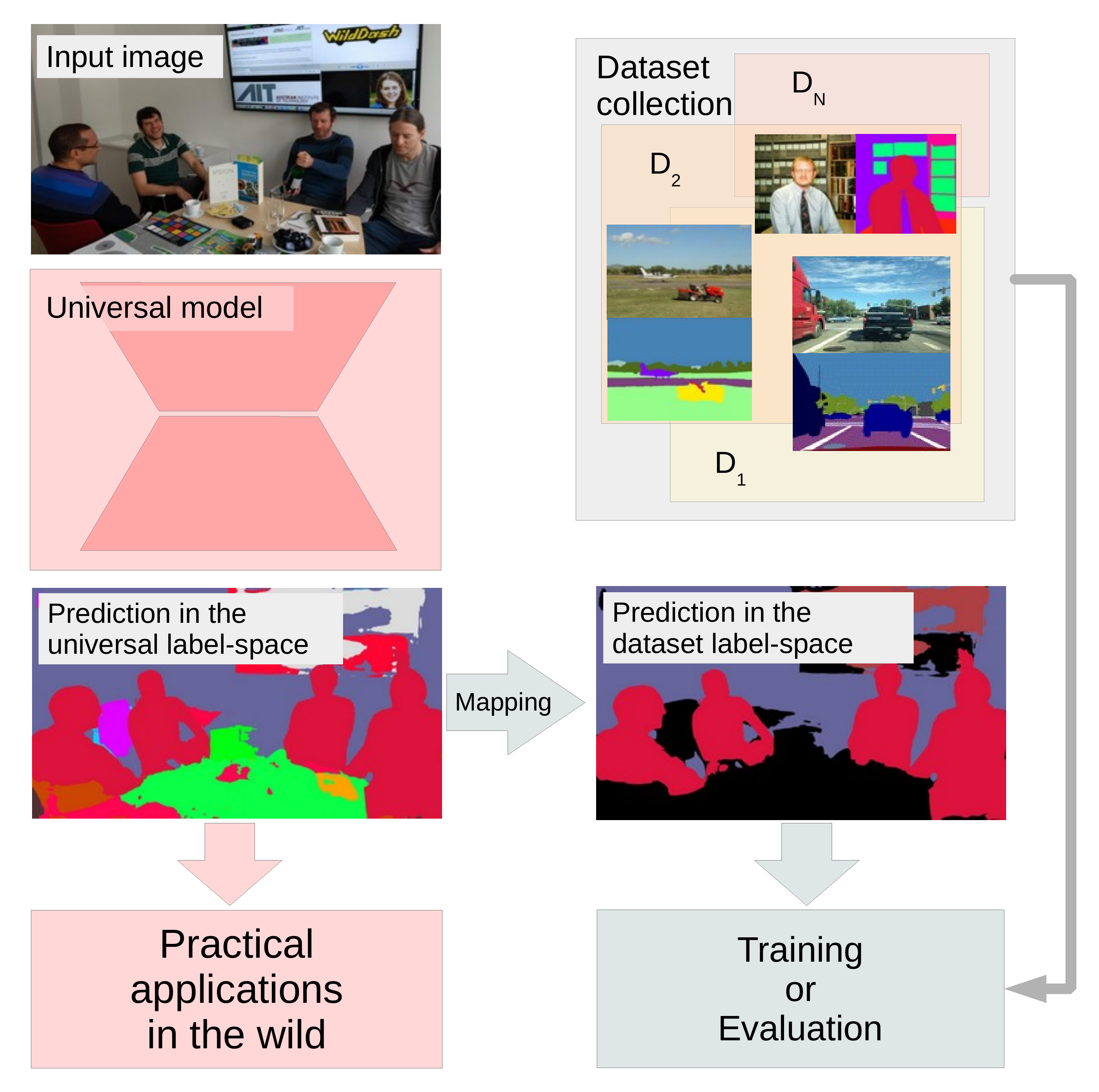}
 \caption{Our models allow 
   universal inference in the wild (left)
   as well as multi-dataset 
   training and validation
   on dataset-specific labels (right).  
  }
 \label{fig:approach-pipeline}
\end{figure}

Universal models output a distribution
over fine-grained universal classes
that amalgamate semantic knowledge
of all training datasets.
They are therefore very convenient
for practical applications in the wild
where we desire graceful performance degradation
in presence of anomalies and hard edge cases.
Furthermore, each dataset-specific posterior 
can be recovered as a sum 
of corresponding universal posteriors.
Thus, universal models can be trained 
by leveraging dataset-specific ground truth
as partial labels \cite{cour2011learning}.
Finally, they can be evaluated 
on dataset-specific test data.
These three ways to interact 
with a universal model are illustrated in  
Figure \ref{fig:approach-pipeline}.

\subsection{Creating a universal taxonomy}
\label{ss:method-taxonomy}
\begin{figure*}[t]
 \centering
 \includegraphics[width=0.95\textwidth]
  {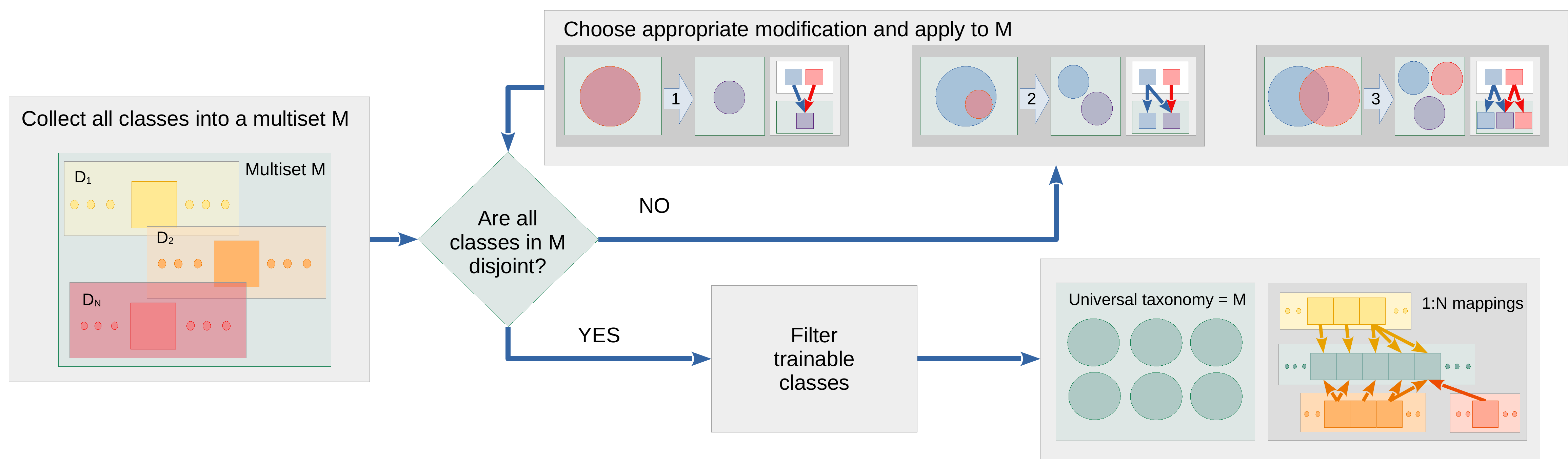}
 \caption{Construction of a universal taxonomy.
  We collect all dataset-specific classes 
  into the multiset $\set M$ (left).
  Then, we iteratively modify $\set M$ 
  according to the three resolution rules 
  from section \ref{ss:method-taxonomy} (top-right).
  The iteration continues until 
  all classes in $\set M$ are disjoint 
  (center).
  Finaly, we filter universal classes 
  that can not be trained 
  with the available supervision 
  (bottom-right).
 }
\label{fig:create-univ}
\end{figure*}
We propose a principled procedure
for recovering the
universal taxonomy 
for a given collection of datasets.
Figure \ref{fig:create-univ}
shows that we start 
the process with the multiset $\set M$
that contains each class from 
all dataset-specific taxonomies.
We iteratively transform $\set M$ 
according to rules that resolve 
three types of overlaps 
between classes. 
Concurrently we update the mappings
that connect dataset-specific classes 
to the remaining classes in $\set M$.
Initial mappings are identity functions.
The three rules for overlap resolution 
can be formulated as follows.
\begin{enumerate}
\item
  If two classes $c_i$ and $c_j$ 
  match exactly, then 
  we replace them with a new
  class $c'$ and remap 
  both $c_i$ and $c_j$ to $c'$.
  \\\emph{Example}: 
  Since \mydscls{WD}{sky} and 
  \mydscls{City}{sky} 
  are equivalent, 
  we merge them into \myMcls{sky},
  and define mappings:
  \mydscls{WD}{sky} 
    $\mapsto$ \myMcls{sky},
  \mydscls{City}{sky} 
    $\mapsto$ \myMcls{sky}.
\item
  If a class $c_i$ is 
  a superset of a class $c_j$,
  then  $c_i$ is removed from $\set M$, 
  a new class 
  $c_i'=c_i\setminus \cls c_j$ is added,
  while
  $c_i$ is remapped to $\{\cls c_j, \cls c_i'\}$.
  \\\emph{Example}: 
  \mydscls{KITTI}{car} is a superset of 
  \mydscls{ADE20k}{car}
  because it contains vans.
  We therefore add classes
  \myMcls{van} and \myMcls{car}, 
  and create mappings
  \mydscls{KITTI}{car} $\mapsto$ 
    \{\myMcls{car}, \myMcls{van}\}
  and
  \mydscls{ADE20k}{car} $\mapsto$ 
    \{\myMcls{car}\}.
\item
  If two classes overlap,
  $(\cls c_i \not\perp \cls c_j)$
  $\wedge$ 
  $(\cls c_i \setminus \cls c_j \neq \emptyset)$
  $\wedge$ 
  $(\cls c_j \setminus \cls c_i \neq \emptyset)$,
  then
  $c_i$ and $c_j$ are replaced with three
  new disjoint classes
  $\cls c_i' = \cls c_i\setminus \cls c_j$, 
  $\cls c_j' = \cls c_j\setminus \cls c_i$, 
  and $c' =\cls c_i \cap \cls c_j$. 
  Class $\cls c_i$ is remapped to 
  $\{\cls c_i', \cls c'\}$,
  while $c_j$ is remapped to
  $\{\cls c_j', \cls c'\}$.
  \\\emph{Example}: 
  \mydscls{VIPER}{truck} contains 
  trucks and pickups while 
  \mydscls{ADE20k}{truck} 
  contains trucks and trailers.
  We therefore replace 
  \mydscls{VIPER}{truck} and 
  \mydscls{ADE20k}{truck}
  with \myMcls{truck}, \myMcls{pickup} and \myMcls{trailer},
  and create the following mappings:
  \mydscls{VIPER}{truck} $\mapsto$ 
   \{\myMcls{truck}, \myMcls{pickup}\} and
  \mydscls{ADE20k}{truck} $\mapsto$ 
   \{\myMcls{truck}, \myMcls{trailer}\}.
\end{enumerate}

The process ends when all rules 
are no longer applicable.
At this moment, all remaining classes
within $\set M$ are disjoint.
Furthermore, they have 
equal or finer granularity 
than all dataset-specific classes. 
In other words, each dataset-specific class 
maps to a subset of classes from $\set M$.
Consequently, $\set M$ now
corresponds to a taxonomy 
that encompasses the entire semantic range 
of the considered dataset collection. 

\subsection{NLL+ loss}

We model the probability 
of universal classes
as per-pixel softmax 
over universal logits $\vec s$.
Let the random variable $\rvar U$
correspond to a universal prediction 
at a particular pixel,
and let $\vec p$ denote the softmax output.
Then, the posterior probability of a single
universal class $\cls u$
corresponds to:
\begin{align}
    \label{eq:probu}
  \P(\rvar U = \cls u \mid \vec x)
  = \softmax(\vec s_{u})\;.
\end{align}

Let the random variable $\rvar Y$
denote a dataset-specific prediction
at a particular pixel and let
$m_{\set S_d}: \set S_d \to 2^U$ 
denote our mapping 
from dataset-specific classes
to subsets of universal classes.
Then, we can express the posterior  
of a dataset-specific class $\cls y$ 
as a sum of the posteriors
of universal classes 
$u' \in m_{\set S_d}(\cls y)$
in the same pixel:
\begin{align}
  \label{eq:proby}
    \P(\rvar Y
      = \cls y\mid \vec x)
    =
    \sum_{\cls u' \in m_{\set S_d}(y)}
    \P(\rvar U
      \cls = u'\mid \vec x) \;.
\end{align}

If we substitute that sum into the standard
negative log likelihood for that pixel,
then we obtain negative log-likelihood 
over aggregated universal posteriors,
which we denote as NLL+:
\begin{align}
  \label{eq:nllplusdef}
  \mathcal{L}^\text{NLL+}
    &(\vec{x}, \cls y \mid m_{\set S_d})
  =
    - \ln 
    \P(\rvar Y
      = \cls y\mid \vec x)
  \nonumber
  \\
  &=
    - \ln 
    \sum_{\cls u' \in m_{\set S_d}(\vec y)}
    \P(\rvar U
       = \cls u'\mid \vec x)
    \;.
\end{align}

NLL+ loss exploits weak supervision 
by learning fine-grained logits 
on coarse-grained labels 
of particular datasets.
Experiments will show that such loss 
can learn visual concepts
that are not explicitly labeled 
in any of the datasets. 

To better understand 
training with the NLL+ loss, 
it is helpful to analyze 
its partial derivatives
with respect to universal logits $\vec s$.
We start by expressing the NLL+ loss
in terms of the logits:
\begin{align}
  \mathcal{L}&^\text{NLL+}
    (\vec{x}, \cls y \mid m_{\set S_d})=
  \nonumber \\
    &=
    - \ln 
    \sum_{\cls u' \in m_{\set S_d}(\cls y)}
    \P(\rvar U
      = \cls u'\mid \vec x)
  \nonumber \\
    &=
    - \ln 
    \frac
      {\sum_{\cls u' \in m_{\set S_d}
       (\cls y)}\exp{s_{u'}}}
      {\sum_{u \in \vec U} \exp{s_{u}}}
  \nonumber \\
    &= 
     \ln \sum_{u \in \set U} \exp{s_{u}}
     -
     \ln 
      \sum_{\cls u' \in m_{\set S_d}
       (\cls y)}\exp{s_{u'}}
     \;.
   \label{eq:loss-expand}
\end{align}

From this we get the following 
gradients of the loss 
with respect to the 
universal logit $s_v$:
\begin{align}
  \label{eq:partial-derivative}
  \frac
    {\partial \mathcal{L}}
    {\partial s_v} 
  &= 
  \frac{\partial}{\partial s_v}
   \ln \sum_{u \in \vec U} 
    \exp{s_{u}}
  -
  \frac{\partial}{\partial s_v}
   \ln 
    \!\!\!\!
    \sum_{\cls u' \in m_{\set S_d}(\cls y)}
    \!\!\!\!
    \exp{s_{u'}}
  \nonumber\\
  &=
  \frac{\exp{s_{v}} }{\sum_{u \in \vec U} 
    \exp{s_{u}}}
  - 
  \frac{\enbbracket{v \in m_{\set S_d}(\cls y)} \exp{s_{v}}}{\sum_{\cls u' \in m_{\set S_d}(\cls y)} \exp{s_{u'}}}  
  \nonumber\\
  &=
  P(\rvar U = \cls v\mid\vec x)
  - 
  P(\rvar U = \cls v\mid\rvar Y = \cls y, \vec x)
  \;.
\end{align}

If $v$ is not associated with 
the label $y$,
then the gradient is strictly positive
and exactly the same as in the 
standard case with crisp labels.
The NLL+ loss pushes 
incorrect logits to $-\infty$ 
just the same as the standard NLL.

If $v$ is a subset of $y$,
then the gradient is strictly negative 
since 
$\P(U = v \mid \vec{x}) \leq 
 \P(U = v \mid y, \vec{x})$.
Furthermore, the magnitude of the gradient 
will be proportional to 
$P (U = v \mid Y = y, x)$.
Thus, the gradients favour 
the universal class 
that is currently winning, 
and that class will become 
even more probable after the update. 
If the model succeeds to lock onto 
the correct universal class,
then the gradient of the correct logit 
will be the same as in the
standard supervised case.

\subsection{Mask-level recognition}

\begin{figure}[t]
\centering
 \includegraphics[width=0.40\textwidth]
   {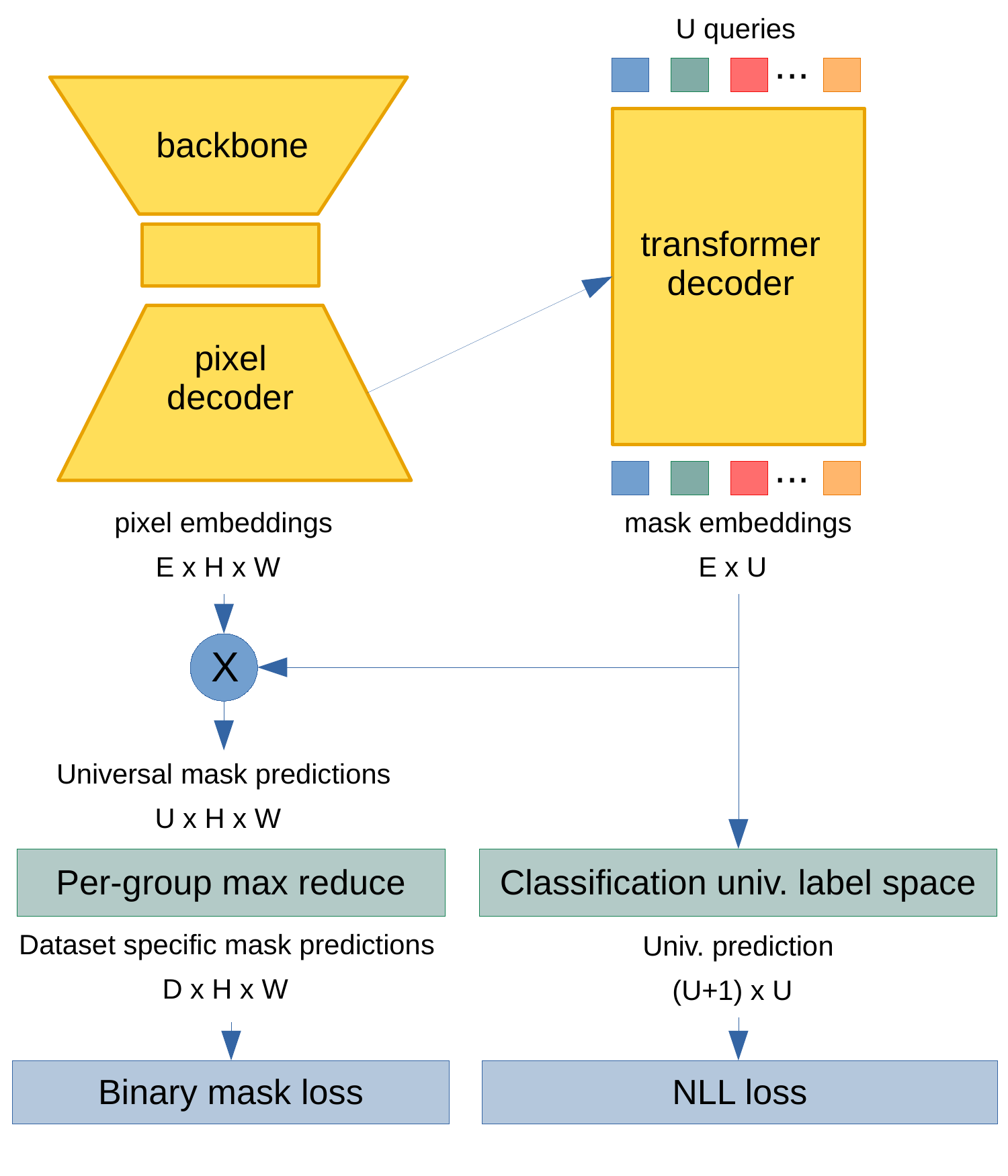}
\caption{We propose an extension 
  of the M2F architecture with fixed matching
  that can be trained with our universal taxonomy.
  The model assigns pixels to universal classes
  according to sigmoid-activated dot products 
  between mask embeddings and pixel-level embeddings. 
  We recover dataset-specific masks
  as maximum pixel-level assignments 
  over the associated universal classes
  (\ref{eq:nll-max}).
}
\label{fig:m2f}
\end{figure}

This section applies universal taxonomies
for multi-dataset training 
of recent semantic segmentation models 
based on mask-level recognition. 
We consider a simplified version 
of a recent architecture
that can be described as 
Mask2Former with fixed matching 
\cite{cheng2021nips,cheng22cvpr}.
Figure \ref{fig:m2f} shows that
the considered architecture
consists of a downsampling backbone, 
pixel decoder and transformer decoder. 
The pixel decoder upsamples 
the backbone features 
towards per-pixel embeddings
$E \times H \times W$.
The transformer decoder 
produces the matrix $E \times U$
that projects embeddings
onto universal multi-label logits.
Finally, the logits give rise
to universal semantic maps
through independent sigmoid activation.

The $U$ vectors of the projection matrix 
are classified into U universal classes
and one null class.
The resulting categorical distributions
are required to match 
the corresponding universal classes 
through standard NLL loss
so that each embedding corresponds
to a single universal class.
The binary mask predictions are trained 
through combination of dice and focal loss. 

Note that we can not score
universal masks through NLL+ 
due to sigmoid activation.
Instead, we propose 
to perform the aggregation
through the max function:
\begin{equation}
  m_y[i,j] = 
   \mathrm{max}_{u \in  m_{\set S_d}(\cls y)}
      m_u[i,j] \; .
  \label{eq:nll-max}
\end{equation}

\subsection{Determining trainable logits}
\label{ss:novel_classes}

\begin{figure*}[t]
\centering
 \includegraphics[width=0.99\textwidth]{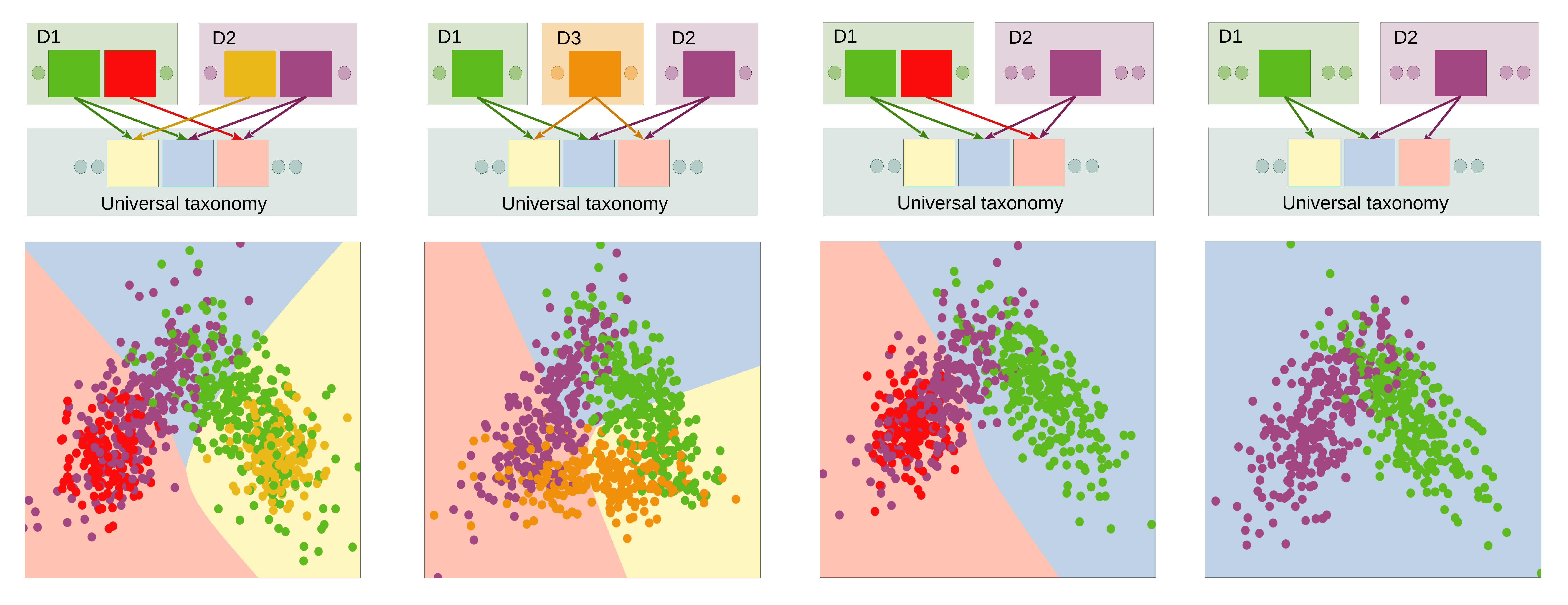}
\caption{Some universal taxonomies
  can not be learned with NLL+.
  The top row shows   
  dataset-specific taxonomies 
  and their mappings 
  to the universal taxonomy. 
  The bottom row shows the 
  training samples (coloured points)
  and the learned decision landscape
  (coloured regions).
  NLL+ succeeds to learn concepts  
  at intersections of
  dataset-specific classes
  (columns 1 and 2).
  However, if a tuple of universal classes
  is always labeled as 
  the same superset class,  
  NLL+ learns to predict
  only one of them
  (columns 3 and 4).}
\label{fig:toyoverlaps}
\end{figure*}
Learning with partial labels
can never learn a universal class 
which is always labeled together 
with some other universal class.
More formally, a universal class $u$ 
is not learnable 
if there is another universal class $u'$ 
that is a subset all dataset-specific classes
that $u$ is a subset of:
\begin{align}
	\exists u' \in \set U, \forall c\in \bigcup_d \set S_d, (u\subset c \Rightarrow u'\subset c) .
\end{align}
In such cases, no training label 
prefers classification into $u$ 
instead of into $c\setminus u$.
Hence, there is no advantage to assign 
a nontrivial probability to $u$.

We illustrate this failure mode
with an example that corresponds
to the rightmost toy problem
in Fig.\ \ref{fig:toyoverlaps}.
Classes from the example
will correspond to colours
from the toy problem.
Suppose we wished to recognize
universal classes
\myunicls{rider} (blue),
\myunicls{pedestrian} (yellow)
and 
\myunicls{bicycle} (pink)
by NLL+ learning from labels  
\mydscls{CamVid}{bicycle} (violet)
=
\myunicls{bicycle} 
$\cup$ 
\myunicls{rider}
and 
\mydscls{Pascal}{person} (green) =
\myunicls{rider} 
$\cup$ 
\myunicls{pedestrian}.
Unfortunately, this setup 
will simply learn to predict both  
\mydscls{camvid}{bicycle} and 
\mydscls{pascal}{person} 
as \mydscls{uni}{rider} since such solution
perfectly minimizes the loss.
Although, there are some good solutions 
that could recognize 
\myunicls{bicycle} 
and
\myunicls{pedestrian},
the NLL+ loss provides no 
incentive to find them
since the optimisation problem 
is underconstrained.
On the other hand, 
if we introduced labels 
that map to 
\myunicls{bicycle}
and 
\myunicls{pedestrian}
independently from 
\myunicls{rider}, 
the optimization would learn 
all three universal classes.
This behaviour is a common limitation 
of all forms of learning with partial labels
\cite{cour2011learning}.

Figure \ref{fig:toyoverlaps} demonstrates 
the ability of the NLL+ loss 
to learn separate concepts 
from overlapping labels
on four toy problems 
over 2D data.
Each label from row 1 is mapped 
to universal classes in row 2. 
We annotate the training samples
with the label colors,
and the decision surfaces
with the colours 
of the universal classes. 
The two rightmost toy problems show 
that some universal logits may die off 
due to insufficient supervision.
On the other hand, 
the two leftmost columns show
that NLL+ optimization succeeds 
whenever the partial labels provide 
enough learning signal. 
These two toy problems also show 
that NLL+ can learn 
to recognize universal concepts 
that are never labeled 
as a standalone class.

There is no need to keep 
universal classes which 
would die off during training
since that would only reduce 
the model efficiency. 
Following this realization, 
we introduce another processing step 
to our procedure 
for recovering the universal taxonomy 
(cf.\ Figure  \ref{fig:create-univ}).
The new processing step 
filters universal classes
that have no chance 
to succeed during training
due to always co-occurring 
with at least one of the siblings.

\begin{figure*}[t]
 \centering
 \begin{subfigure}{0.24\textwidth}
 \begin{subfigure}{0.95\textwidth}
 \centering
  \includegraphics[width=0.95\textwidth]{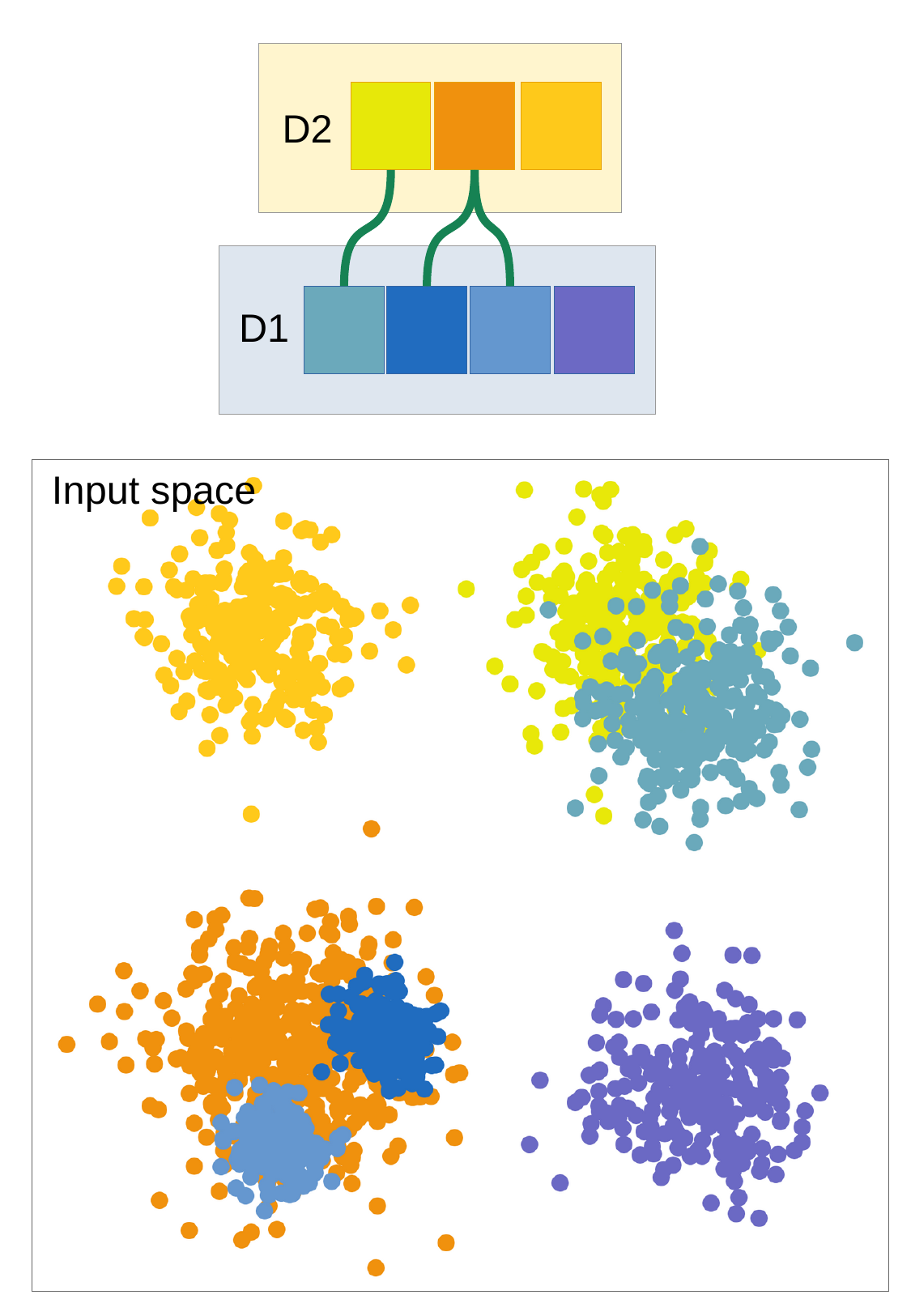}
 \caption{Training data}
 \label{fig:approaches-data}
 \end{subfigure}
 \end{subfigure}
 \centering
 \begin{subfigure}{0.24\textwidth}
 \centering
 \includegraphics[width=0.95\textwidth]{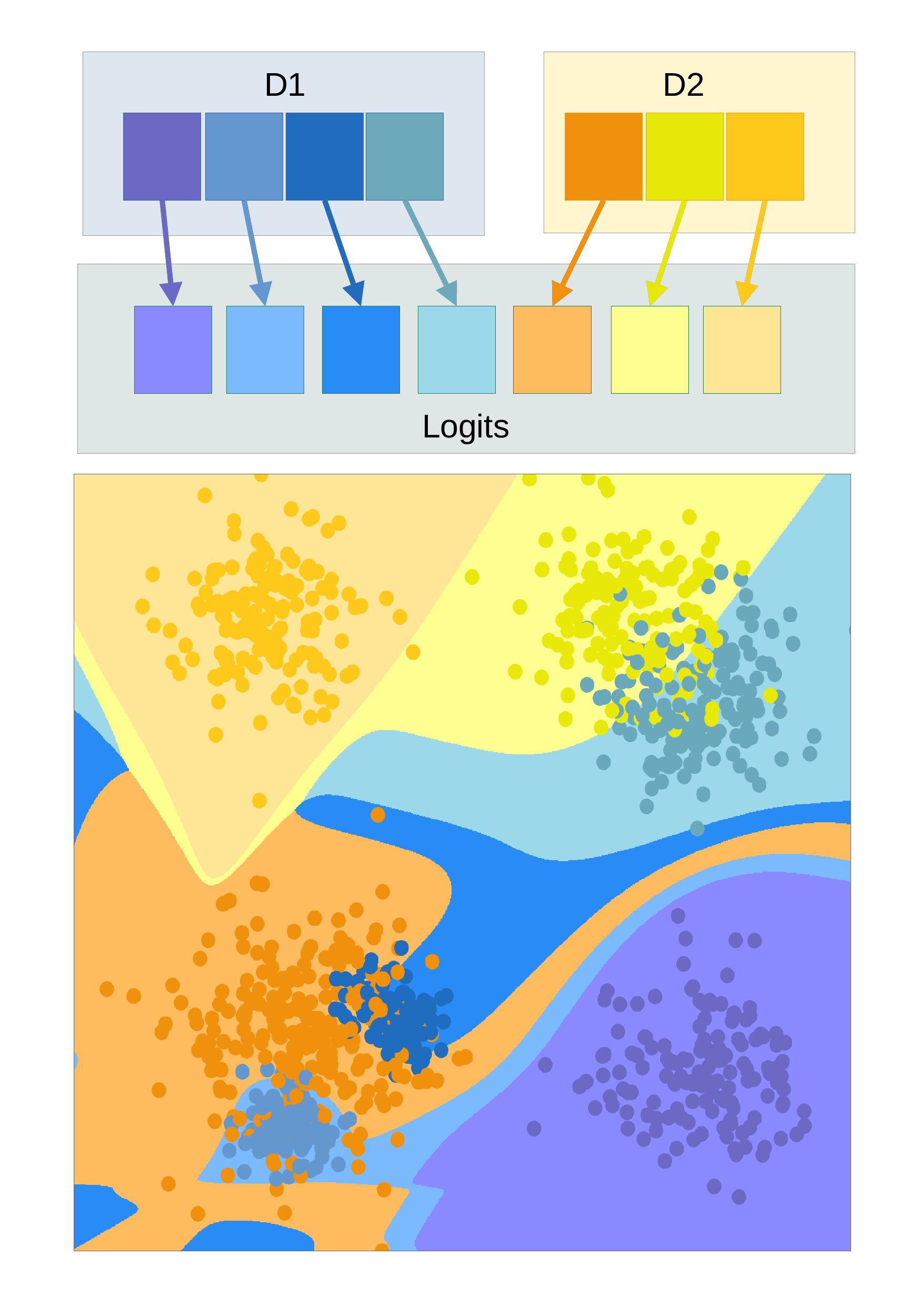}
 \caption{Naive concatenation}
 \label{fig:approaches-concat}
 \end{subfigure}
 \centering
 \begin{subfigure}{0.24\textwidth}
 \centering
 \includegraphics[width=0.95\textwidth]{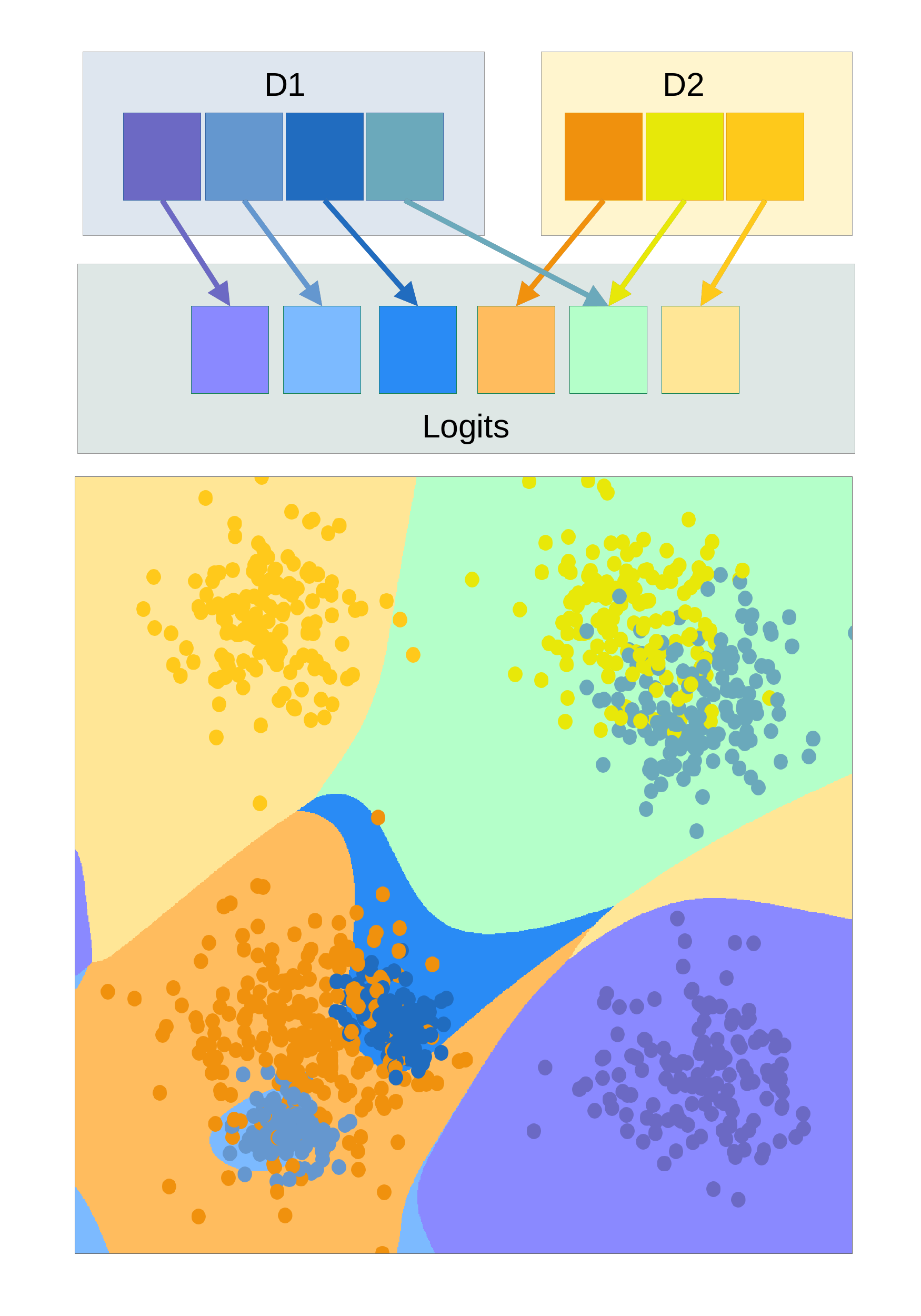}
 \caption{Partial merge}
 \label{fig:approaches-partial}
 \end{subfigure}
 \begin{subfigure}{0.24\textwidth}
 \centering
 \includegraphics[width=0.95\textwidth]{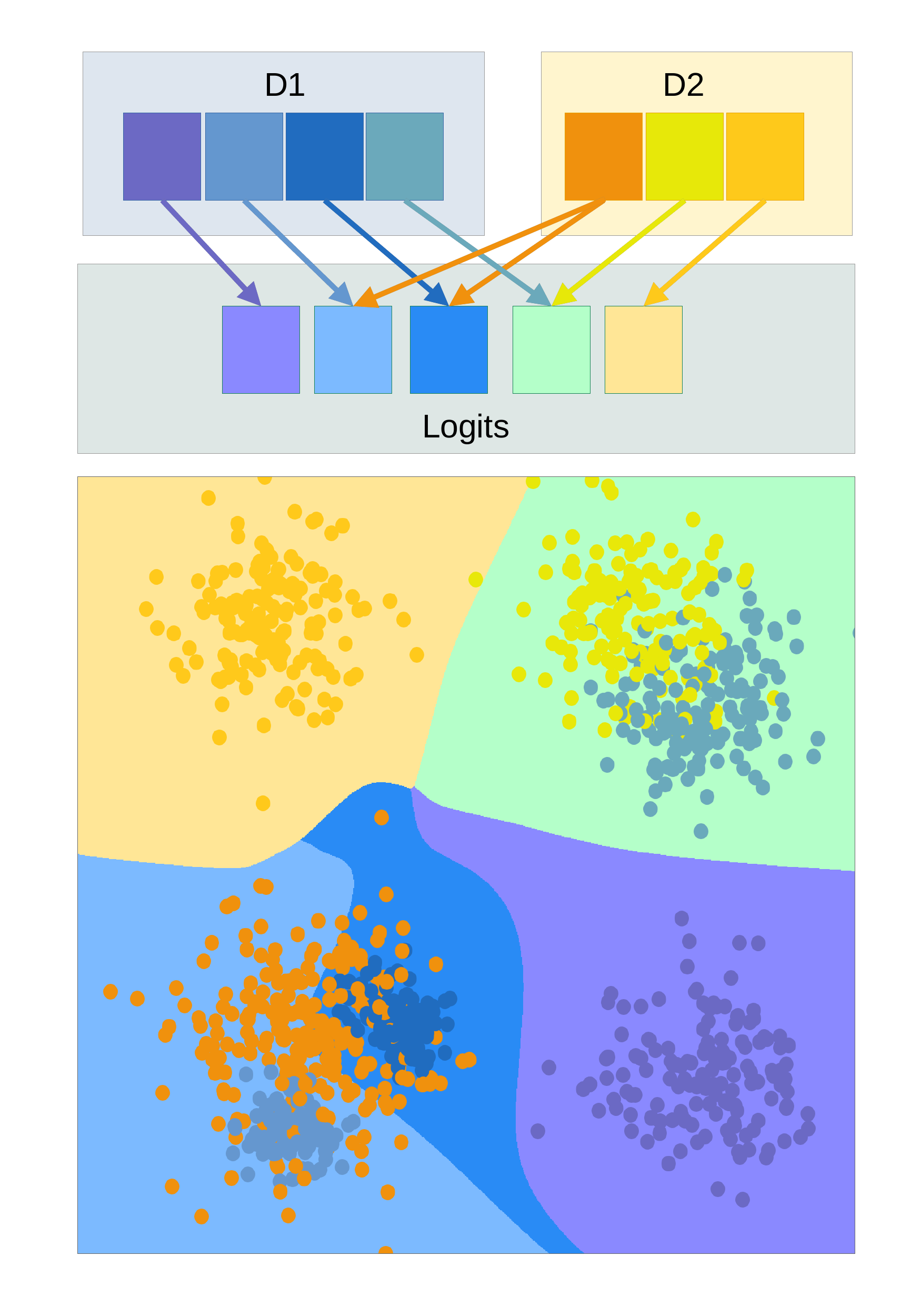}
 \caption{Universal taxonomy}
 \label{fig:approaches-nllplus}
 \end{subfigure}
\caption{
  Multi-dataset training on two toy datasets 
  with semantically related classes (a).
  Our models (b-d) involve shared features 
  and flat softmax predictions.
  Naive concatenation is unaware of all semantic relations (b). 
  Partial merge is aware 
  of class equivalence (c).
  Our universal model is aware 
  both of class equivalence and
  class overlap (d).
  The two baselines (b-c) 
  learn ragged decision boundaries
  due to conflicting gradients
  in overlapping classes.
  Smooth decision landscape
  of the universal model (d)
  suggests that avoiding competition 
  between semantically related concepts
  improves generalization.
}
\label{fig:baselines}
\end{figure*}
\section{Experimental setup}

Our experiments validate the proposed method
against several baselines
and compare it against the previous work.
Our baselines involve 
strongly supervised learning
on two pseudo-taxonomies
and universal learning
with pseudo-relabeled ground truth.
We take special care in order 
to properly evaluate predictions 
that fall outside of the particular 
evaluation taxonomy.
Our experiments promote fair comparison
by pairing approaches
with the same architecture. 
We carefully describe 
the implementation details
in order to promote reproducibility
of our experiments.

\subsection{Baselines}
\label{ss:baselines}
Multi-dataset training can be conceived
by complementing shared features
with per-dataset segmentation heads
\cite{fourure17neucom,masaki21itsc}.
However, realistic applications require
inference in mixed-content images.
For instance, an autonomous vehicle
should be able to recognize
a COCO chair on a Vistas road.
This requirement could be addressed
by supplying an additional head 
for dense dataset recognition.
In this case, we can recover
the joint posterior of 
class $c$ and dataset $\set D$ 
in each pixel according to:
\begin{equation}
  P(c,\set D \mid \vec{x})=
    P(c \mid \set D, \vec{x})
    \cdot
    P(\set D \mid \vec{x})
  \label{eq:two-head}
\end{equation}
However, dataset recognition
does not make much sense
when training on datasets
from the same domain
(e.g.\ Vistas and WildDash 2).
Moreover, submissions 
with explicit dataset recognition
have been outright prohibited at major 
multi-domain recognition competitions
\cite{rvc22www}. 
Hence, we consider another two baselines
that outperform per-dataset heads 
in our multi-dataset experiments.

We now consider a related 
alternative approach
that we denote as naive concatenation.
Similarly to per-dataset heads, 
the naive concatenation also assigns 
a distinct training logit 
to each dataset-specific class.
The only difference is that 
here all logits get jointly activated
with a common softmax.
Such models have to discriminate 
semantically related classes 
from different datasets,
or, in other words, perform
a kind of implicit dataset recognition.
This promotes overfitting 
to dataset bias
instead of encouraging 
cross-dataset generalization. 
For instance, we can not hope 
to learn anything useful
from discriminating 
Vistas cars and WildDash cars.
Besides wasting the model capacity 
on dataset recognition, 
redundant logits hamper the training
due to increased memory footprint. 
Furthermore, the inferred semantics 
has to be post-processed
if we wish to consolidate 
related logits from different taxonomies.

\begin{figure*}[t]
\centering
\includegraphics[width=0.3\textwidth]{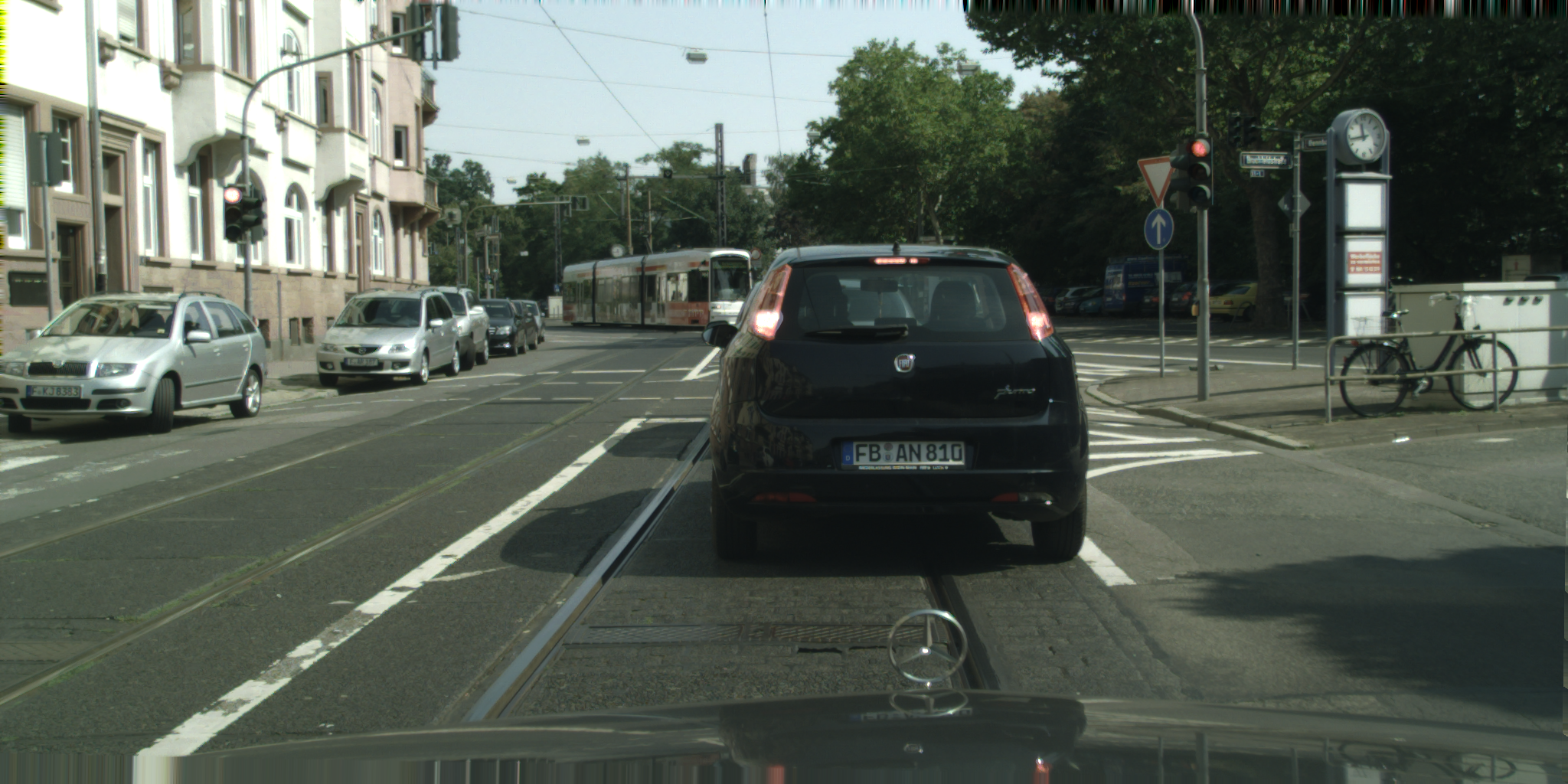}
\includegraphics[width=0.3\textwidth]{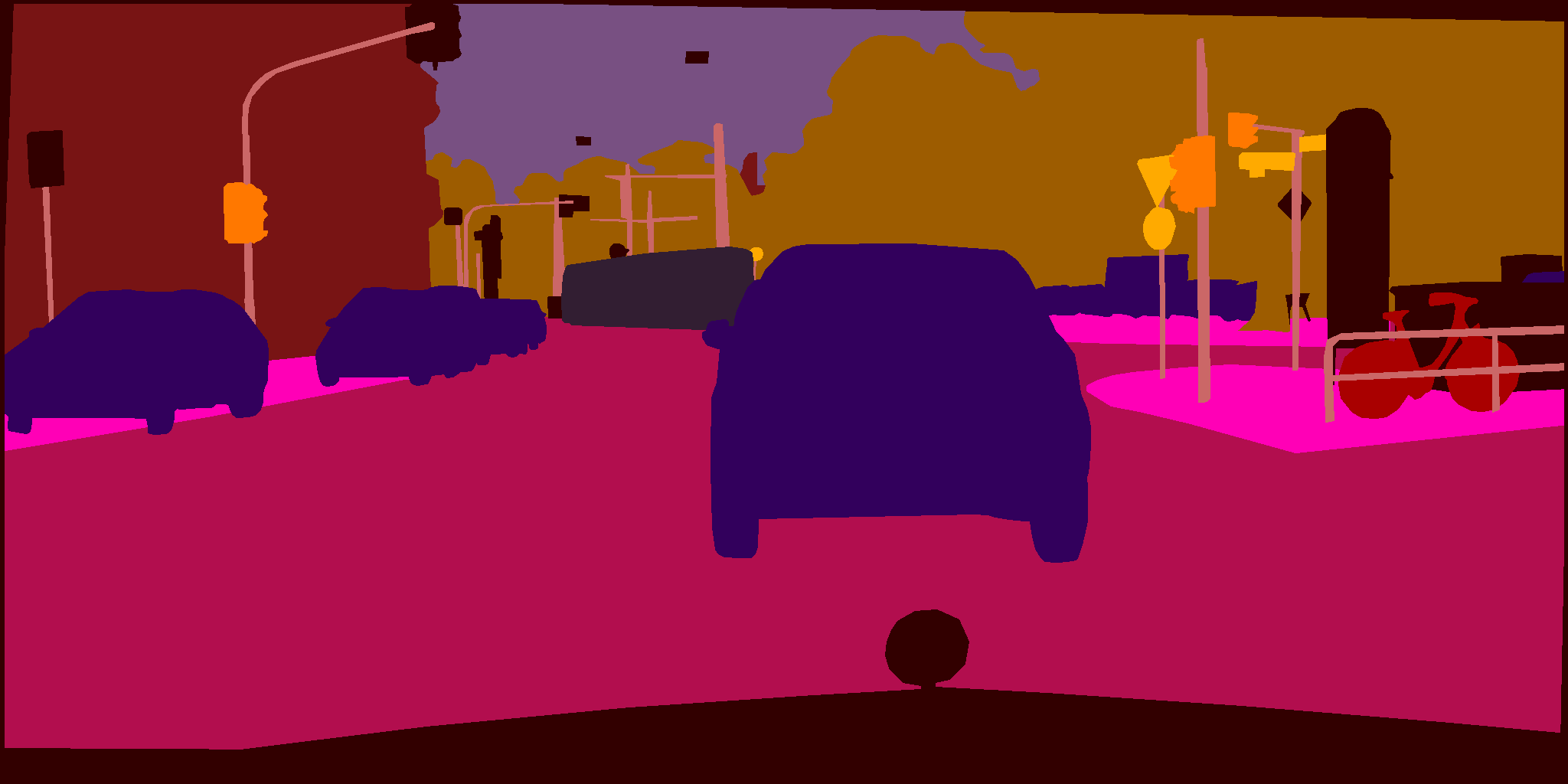}
\includegraphics[width=0.3\textwidth]{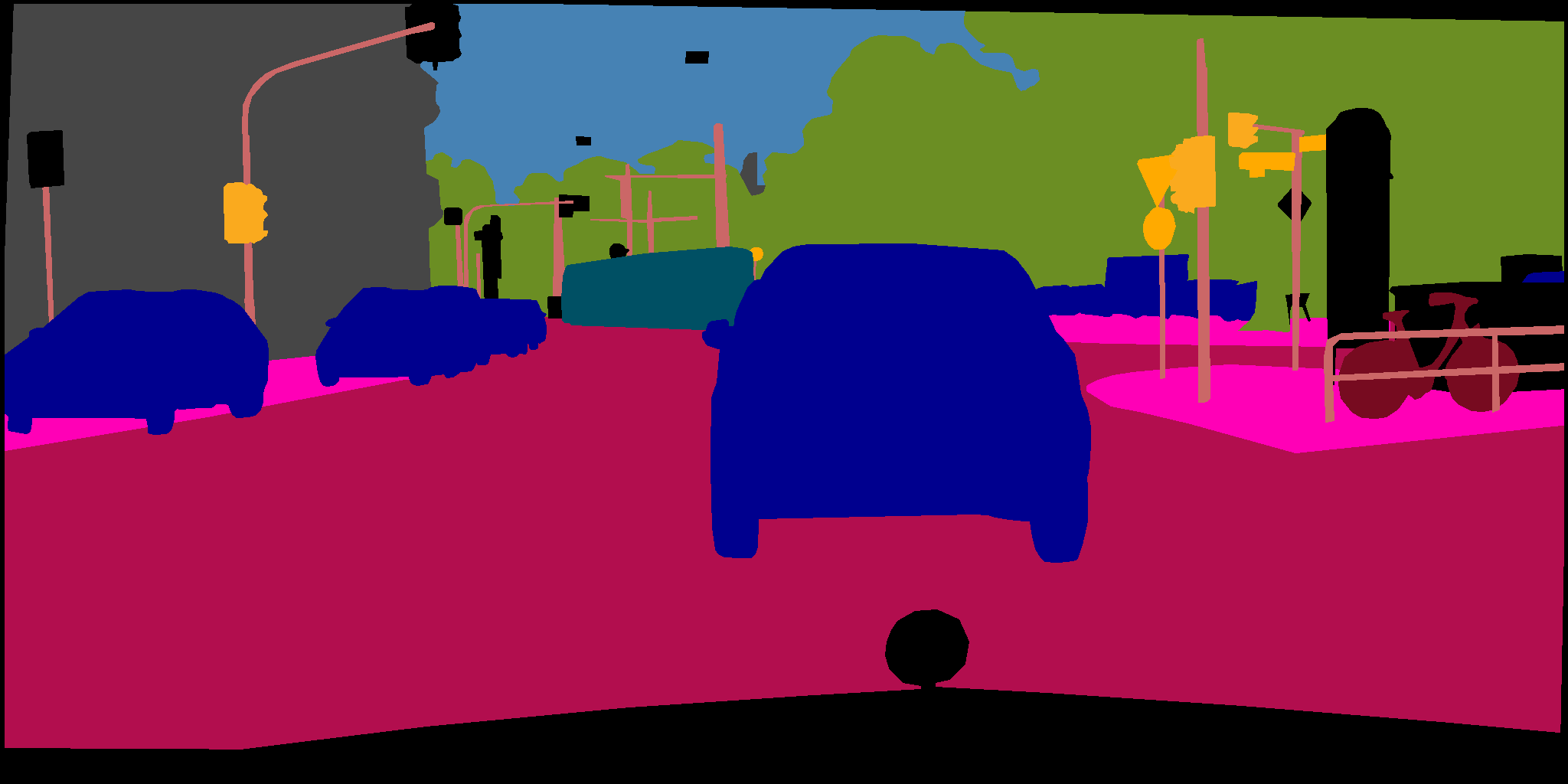}\\
\includegraphics[width=0.3\textwidth]{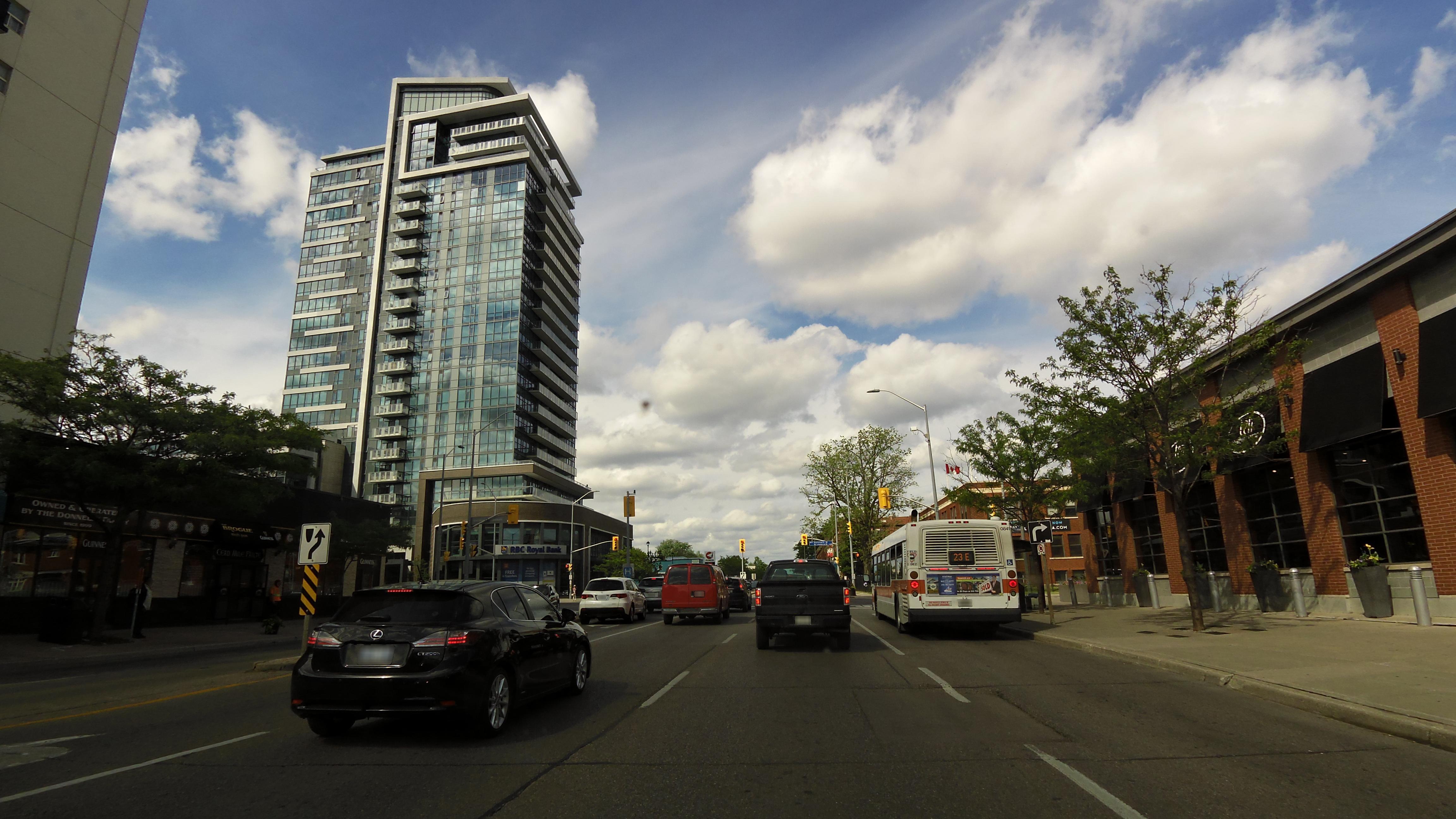}
\includegraphics[width=0.3\textwidth]{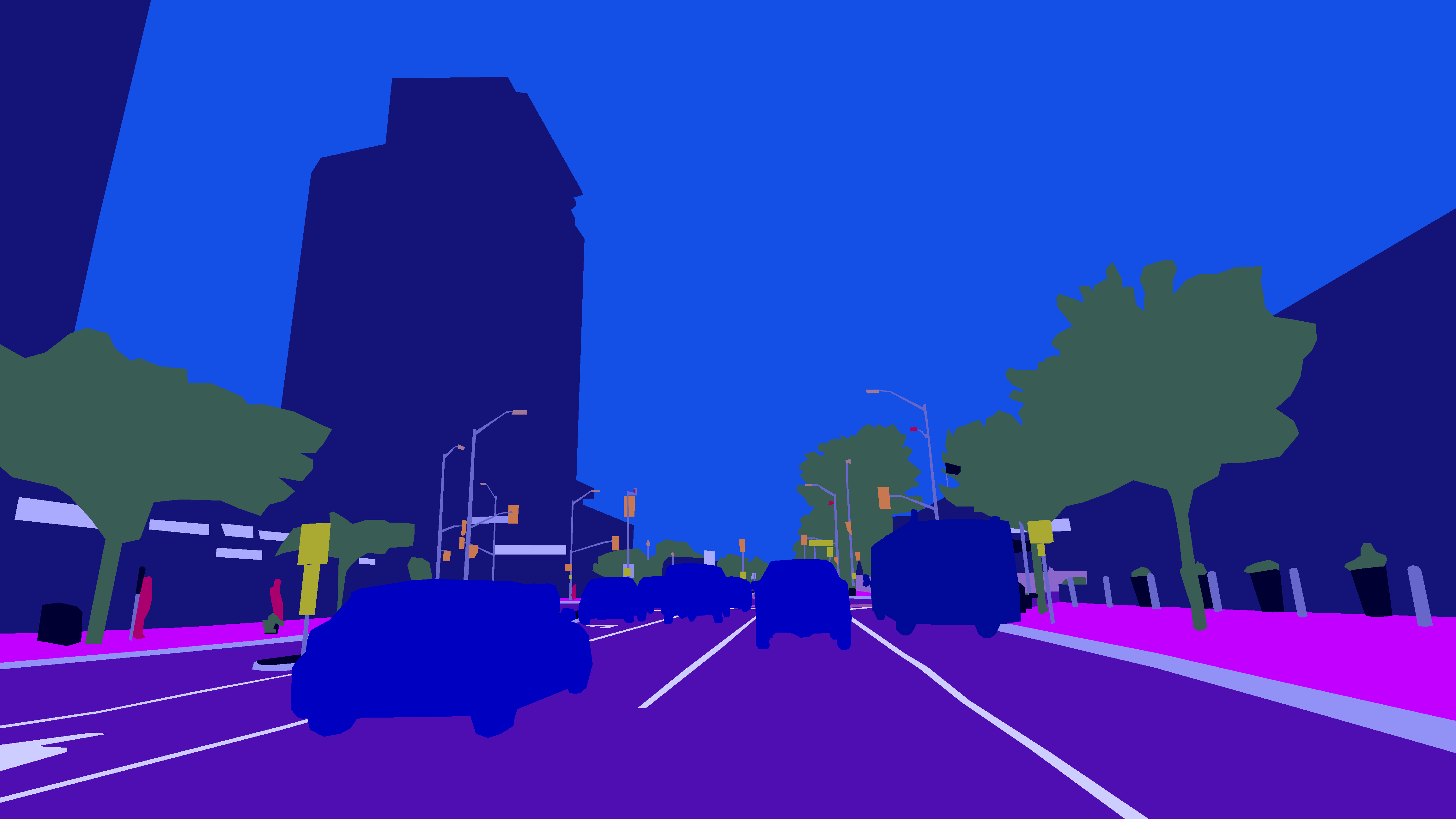}
\includegraphics[width=0.3\textwidth]{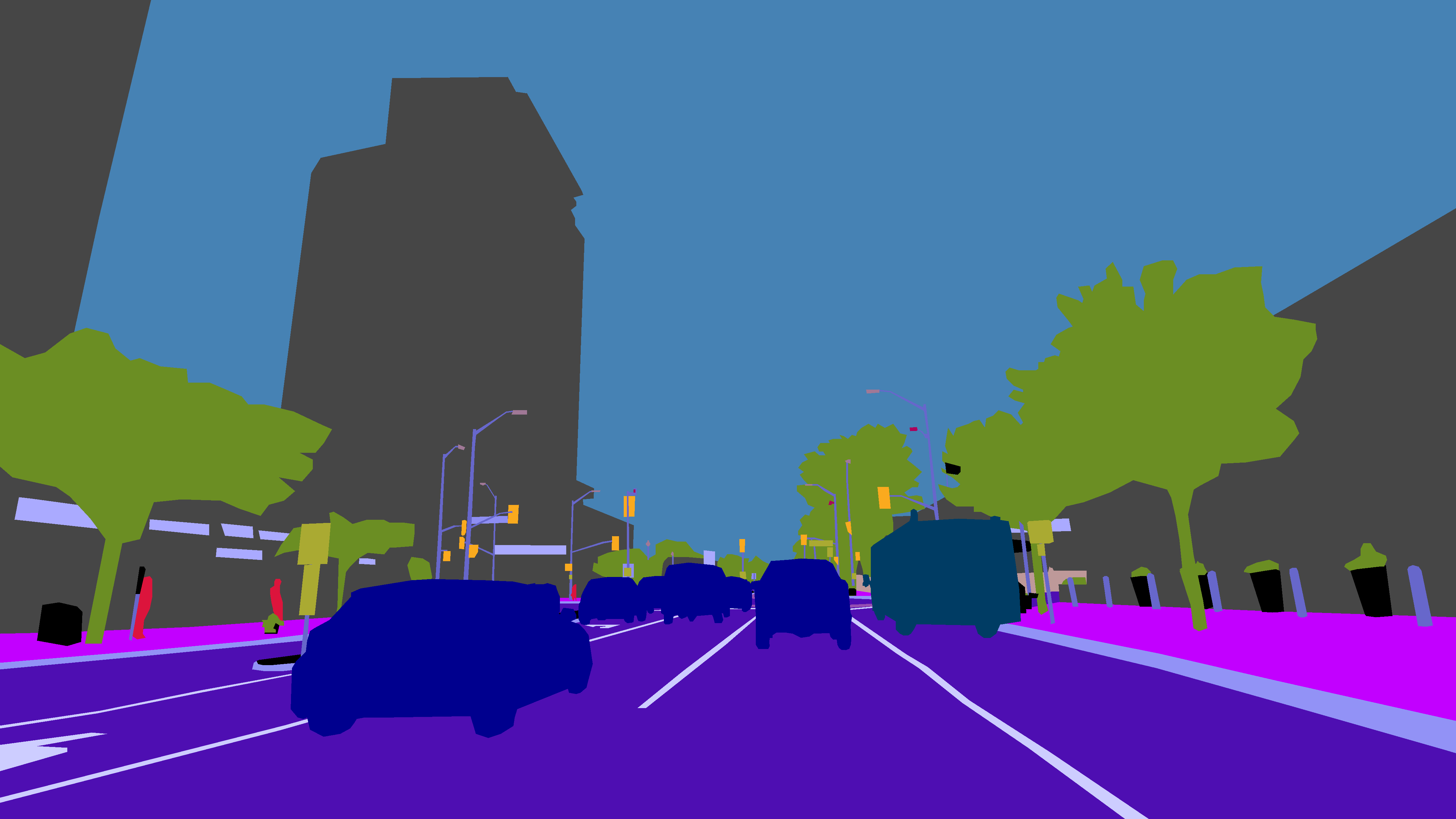}
\caption{We illustrate the training logits
  for our two baselines and joint training
  on Cityscapes and Vistas.
  The rows show validation images 
  from the two datasets.
  The columns show the input image, 
  the naive concatenation ground truth,
  and the partial merge ground truth.
  In naive concatenation all classes are
  considered semantically separate. 
  In partial merge identical classes 
  such as \mycls{sky}, \mycls{vegetation} 
  and \mycls{building}
  are merged into the same class, 
  while overlapping classes such as \mycls{road} and \mycls{road marking} remain separate.
}  
\label{fig:real-baselines}
\end{figure*}

Drawbacks of naive concatenation 
can be mitigated by merging classes 
with identical semantics.
We refer to the resulting approach
as partial merge, since it is unable
to consolidate overlapping classes.
In comparison with naive concatenation,
this approach reduces the waste of capacity
but it does not completely eliminate 
the competition between related logits.
For instance, a partial merge model
would have independent logits for 
\mydscls{WD}{road} and 
\mydscls{Vistas}{crosswalk}.
Such model would need to discriminate
WildDash crosswalks from Vistas crosswalks,
and that does not encourage 
generalization in the wild.
We conclude that partial merge models 
still have to discriminate datasets,
and that this problem can be addressed
only through a proper universal taxonomy.

Figure \ref{fig:baselines} compares 
naive concatenation and partial merge
with our universal approach
on a toy 2D problem.
Dataset $\set D_1$ consists of 4 classes
that we designate with blue hues.
Dataset $\set D_2$ consists of 3 classes
that we designated with yellow hues.
Subfigure (a) shows that
there is one 1:1 correspondence
and one 2:1 correspondence
between classes of $\set D_1$ 
and $\set D_2$.
Thus, naive concatenation has 7 logits,
partial merge has 6 logits, 
while our universal taxonomy has 5 logits.
We observe that our universal model
is likely to generalize better 
due to smoother decision surfaces.

Figure \ref{fig:real-baselines}
compares the training taxonomies 
of naive concatenation and partial merge 
for joint training on Cityscapes and Vistas.

\subsection{Universal pseudo-labels}

Training with the NLL+ loss
(\ref{eq:nllplusdef})
introduces some noise 
into the learning process,
since the model may overfit 
to any of the incorrect 
universal classes 
of the partial label.
This noise can be removed
by manually refining the labels
towards correct universal classes.
However, manual relabeling 
is slow and costly.
Hence, we consider to collect 
universal pseudo-labels
by considering predictions 
trained on particular datasets.

\begin{figure}[hb]
 \centering
 \includegraphics[width=0.45\textwidth]
  {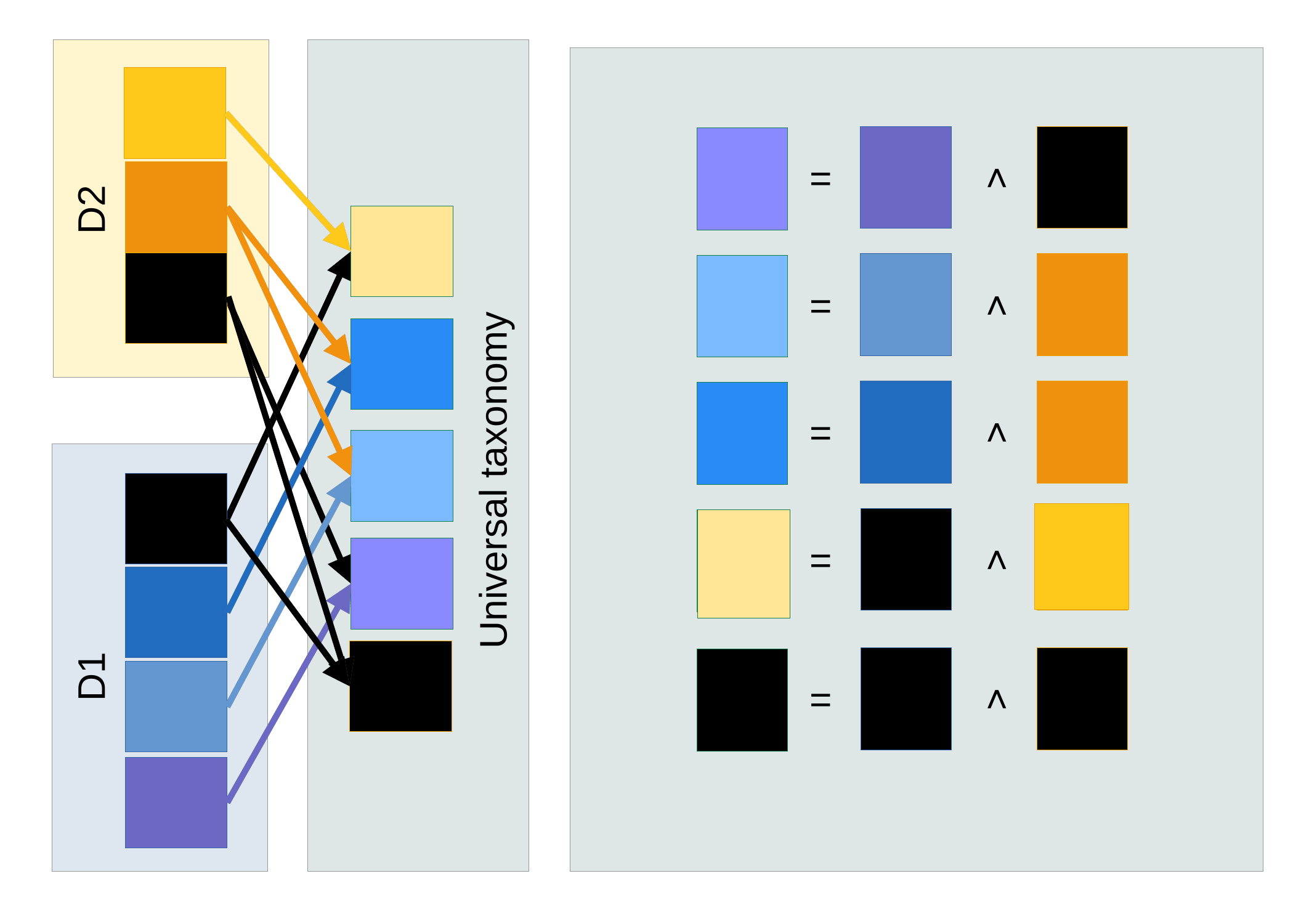}
 \caption{Universal classes may be defined
 as unique tuples of dataset-specific classes.
 Therefore, an ensemble of dataset-specific open-set 
 classifiers can deliver universal pseudo-labels.
 }
 \label{fig:pseudolabel-open-graph}
\end{figure}

We start by noting that each universal class 
can be expressed as intersection 
of the associated dataset-specific classes. 
Thus, we may represent each universal class 
with a unique set of labels. 
Consider the taxonomy 
from Figure \ref{fig:approach-taxonomy}.
Class  \myunicls{pickup} could be defined
as \mydscls{VIPER}{truck} $\cap$ \mydscls{Vistas}{car} $\cap$
\mydscls{ADE20k}{van}, 
while \myunicls{van}  could be defined as 
\mydscls{VIPER}{van} $\cap$ 
\mydscls{Vistas}{car} $\cap$
\mydscls{ADE20k}{van}.
Note that some universal classes 
can be orthogonal
with respect to a given dataset.
For example, 
\myunicls{cushion} would be defined as
\mydscls{ADE20k}{cushion}, 
since cushions are outliers 
for VIPER and Vistas datasets. 
This suggests that closed-set classifiers 
are not suitable for universal pseudo-labeling
as they produce false positives 
in outlier pixels.

Consequently, we consider open-set 
dataset-specific taxonomies 
that introduce a separate class 
for the unknown visual world.
In this case,
\myunicls{cushion} would be defined as 
\mydscls{ADE20k}{cushion} $\wedge$
\mydscls{VIPER}{outlier} $\wedge$ 
\mydscls{Vistas}{outlier}.
Consequently, one could perform 
such pseudo-labelling
with ensembles of dataset-specific models
as shown in Figure 
\ref{fig:pseudolabel-open-graph}.

However, the quality of pseudo-labels
could be further improved
by leveraging  the ground truth.
Suppose we are given an input image
and the corresponding dense labels  
$(\vec{x}^a,\vec{y}^a) \in \set{D}_a$.
Then we can express
dense pseudo-labels 
for a universal class $u$
as sum of conditional 
per-dataset scores:
\begin{equation}
  \mathbf{S}(u \mid \vec x^a, \vec y^a) = 
    \sum_{d \neq a} 
    \mathbf{S}^{d}
      (u \mid \vec x^a, \vec y^a).
  \label{eq:relabel}
\end{equation}
We formulate $\mathbf{S}^d$ 
in terms of predictive probabilities
of a model trained on $\set{D}_d$.
The score depends on whether
$\set{D}_d$ contains a class
$c_{i}^d \in \set{S}_{\set{D}_d}$
that maps to the universal class $u$
($c_{i}^d \mapsto u$).
Note that there can be 
at most one such class
since we assume that all datasets
have proper taxonomies.
If $c_{i}^d$ does not exist
or if the ground truth $\vec y^a$ 
in the particular pixel $(r,k)$
does not map to $u$, then 
$S^d_{rk}(u \mid \vec x^a, y^a_{rk})=0$.
Otherwise, the score approximates
the conditional probability 
of $c_{i}^d$ given 
the ground truth $y^a_{rk}$:
\begin{equation}
  {S}^d_{rk}(u \mid \vec x^a, y^a_{rk}) = 
  \frac
    {\P(Y_{rk}=c_{i}^d \mid \vec x^a)}
    {\sum_{c_j^d \not\perp y^a_{rk}}
     \P(Y_{rk}=c_j^d \mid \vec x^a)
    } 
    \;
\end{equation}

\begin{figure*}[t]
\includegraphics[width=0.24\textwidth]{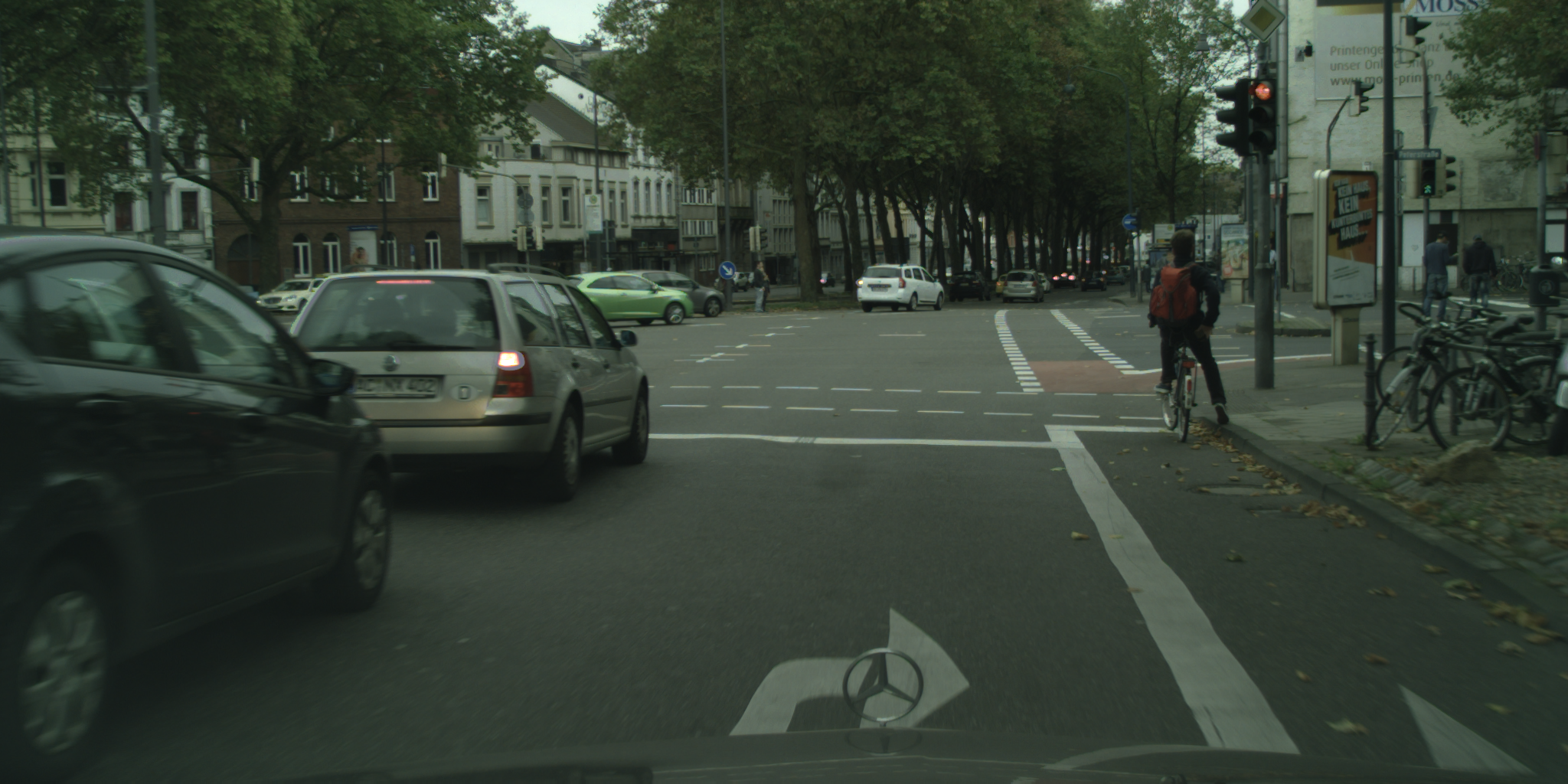}
\includegraphics[width=0.24\textwidth]{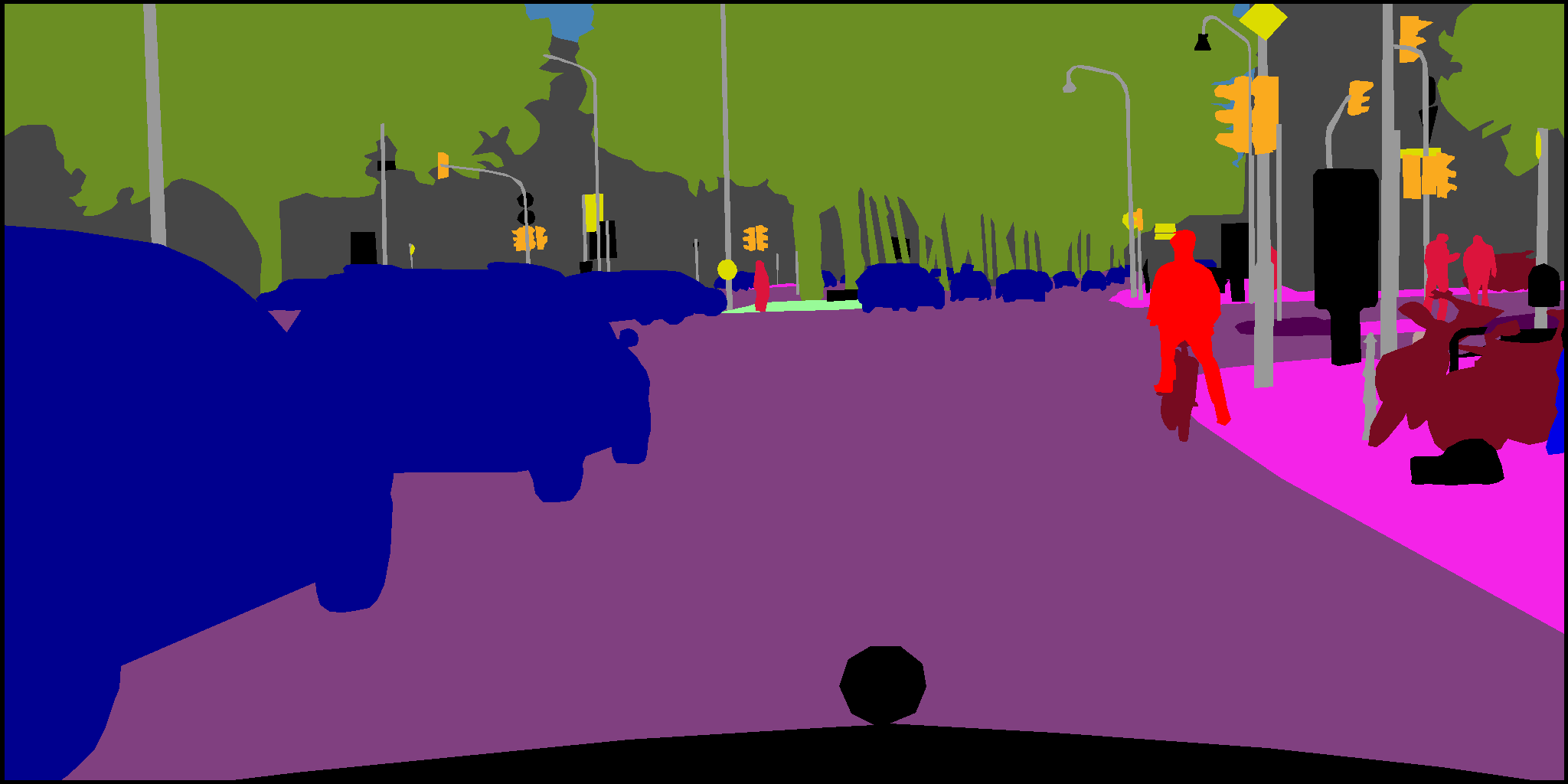}
\includegraphics[width=0.24\textwidth]{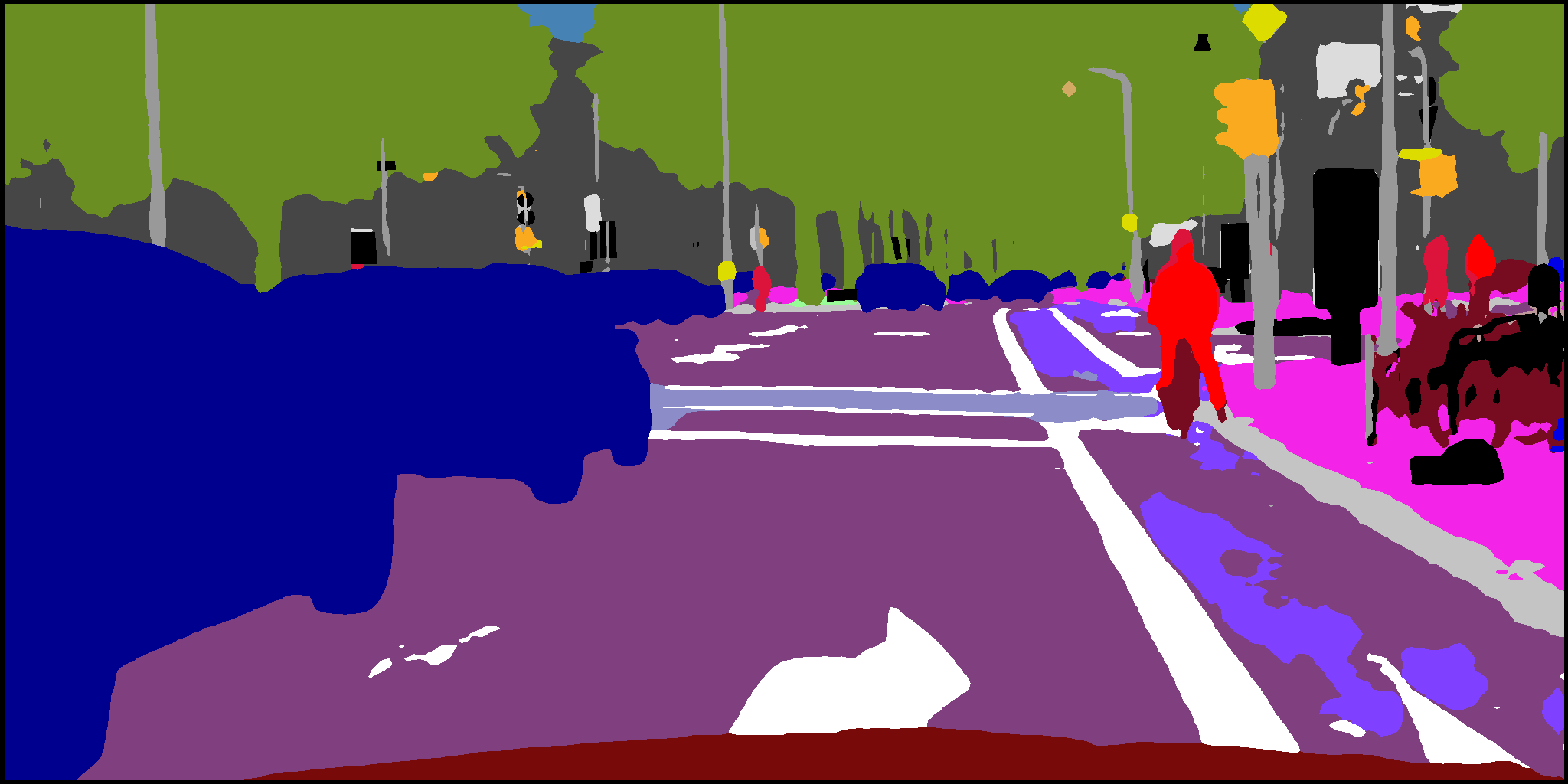}
\includegraphics[width=0.24\textwidth]{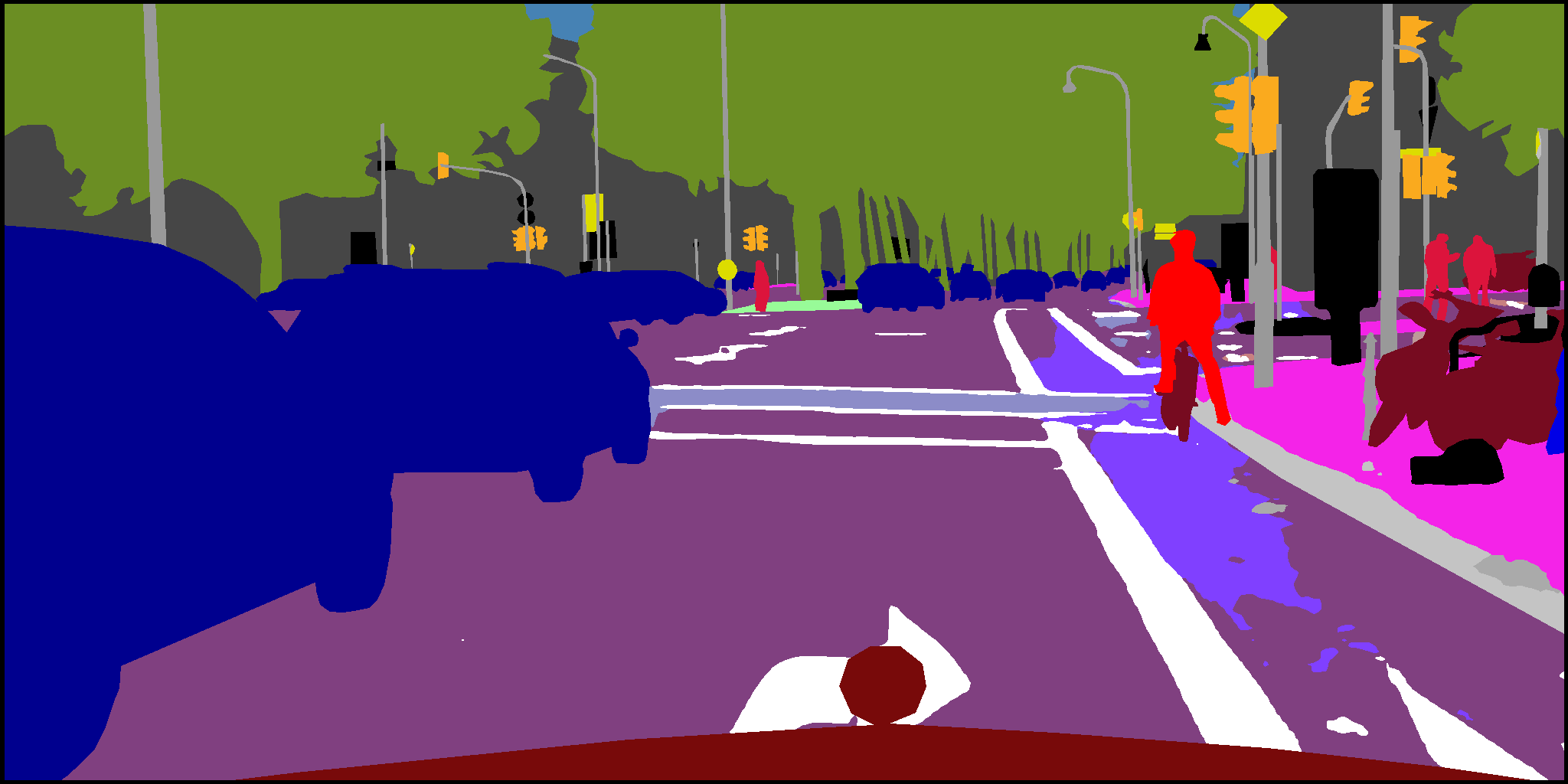}\\
\includegraphics[width=0.24\textwidth]{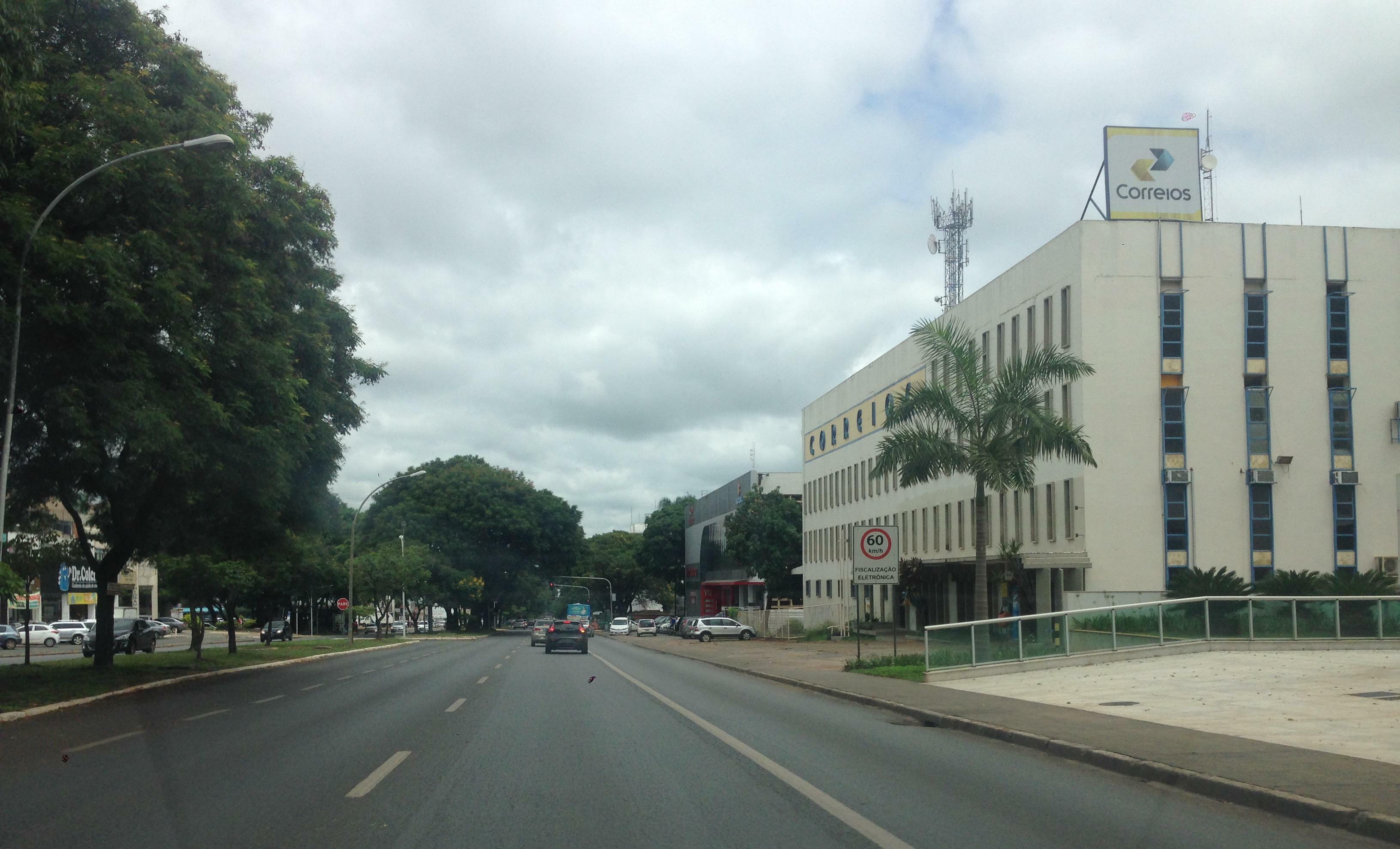}
\includegraphics[width=0.24\textwidth]{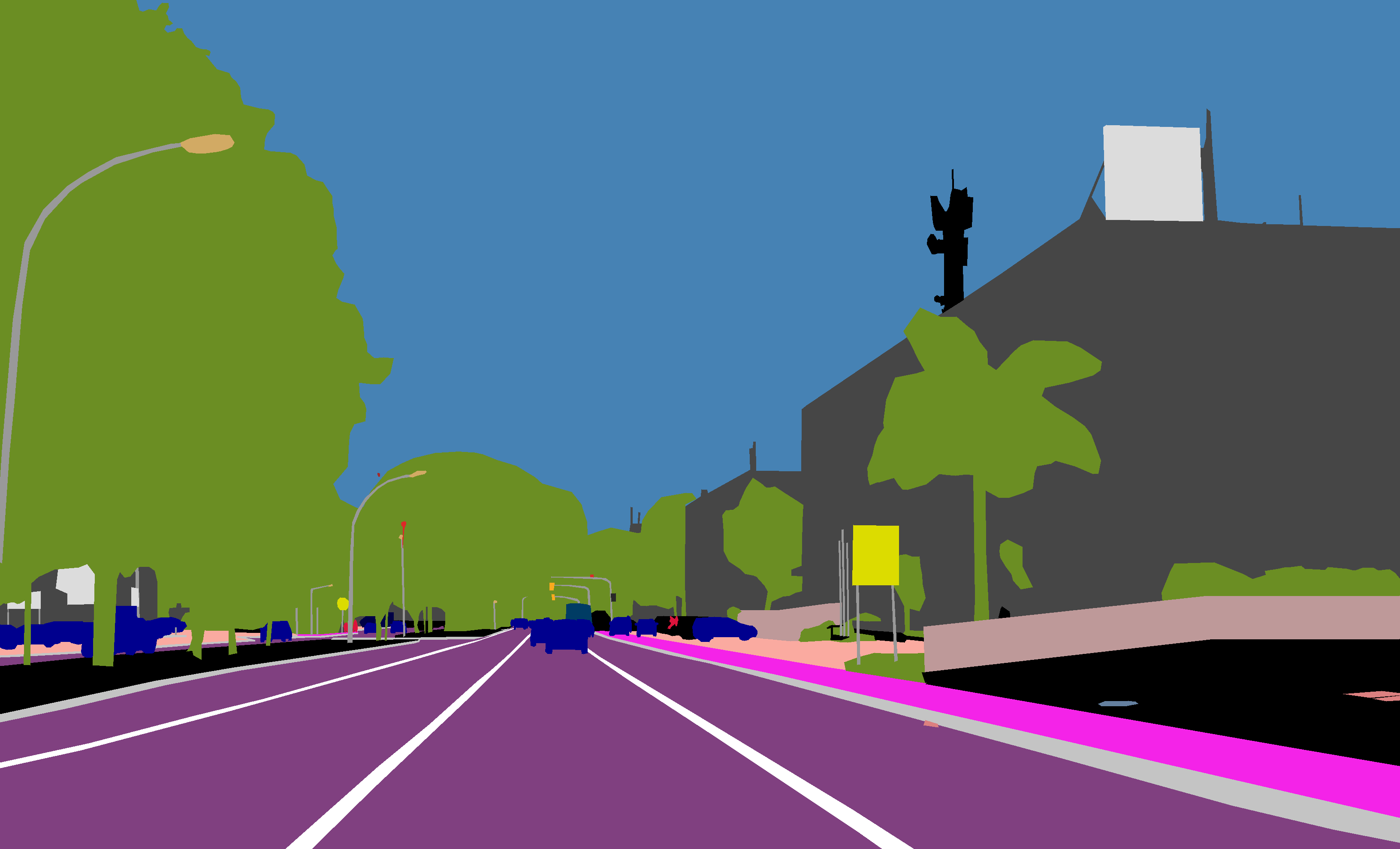}
\includegraphics[width=0.24\textwidth]{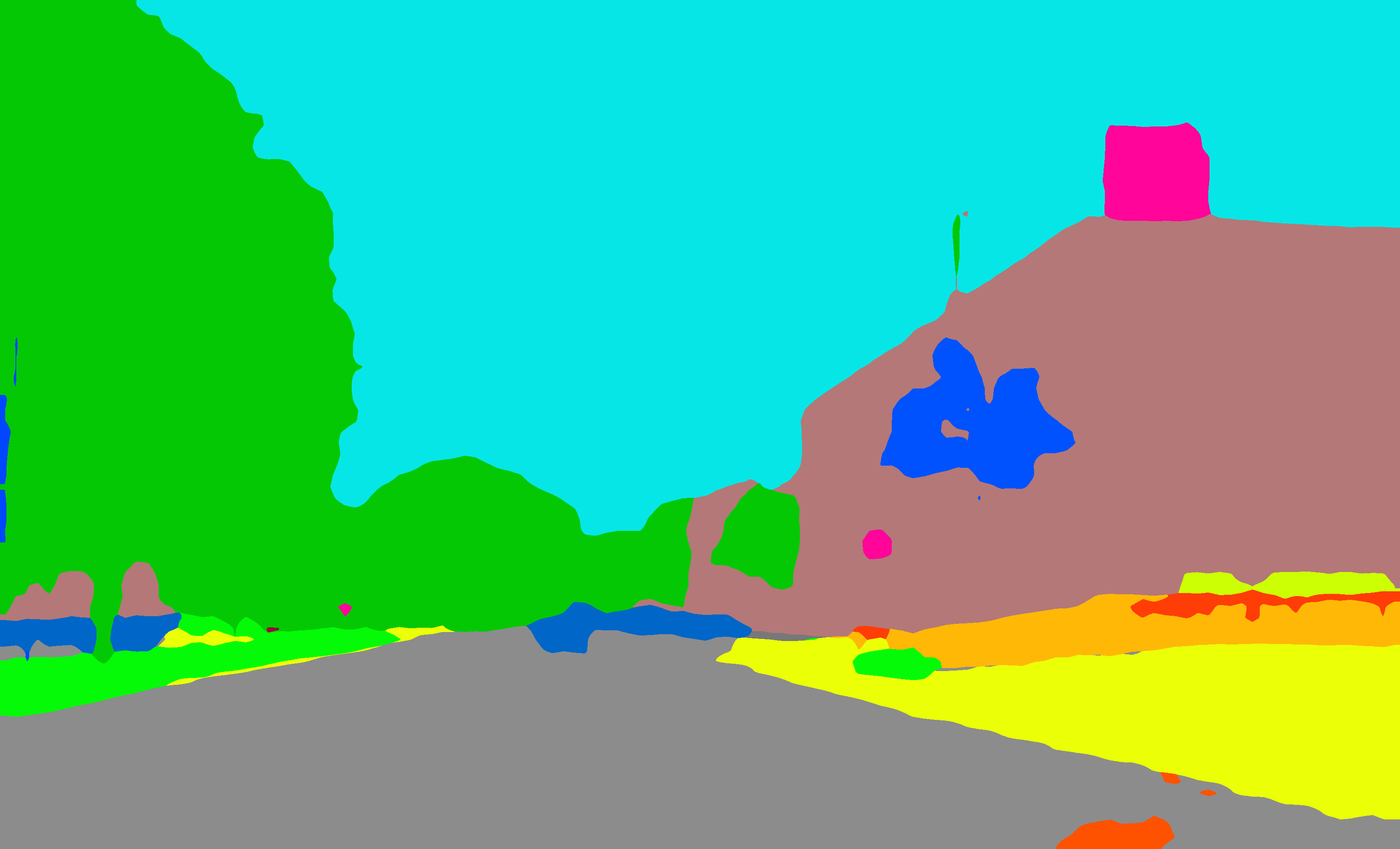}
\includegraphics[width=0.24\textwidth]{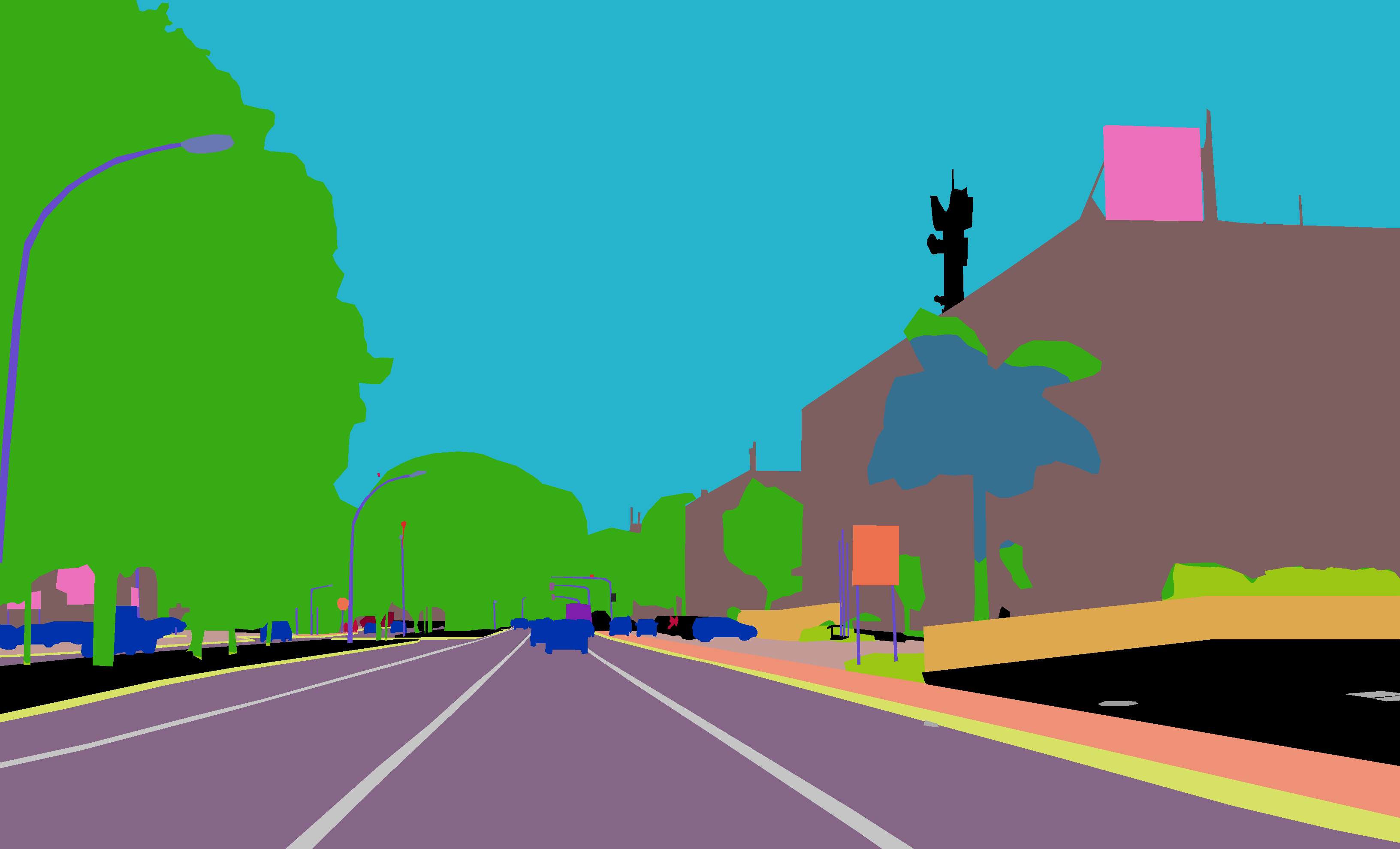}
\caption{We form universal pseudo-labels 
  by refining native labels 
  according to predictions 
  from foreign dataset-specific models.
  The two rows illustrate pseudo-labels
  Cityscapes $\rightarrow$ Cityscapes-Vistas, 
  and Vistas $\rightarrow$ ADE20K-Vistas.
  The columns show: the input image, 
  native ground-truth, 
  foreign predictions and 
  our universal pseudo-labels.
  Note that the prediction mistakes from column 3
  are reduced in column 4. 
  This is evident on edges of cars and distant poles 
  (top),
  and the palm trunk (bottom). 
}  
\label{fig:pseudo-labels}
\end{figure*}

\begin{figure}[ht]
 \centering
   \includegraphics[width=0.45\textwidth]
  {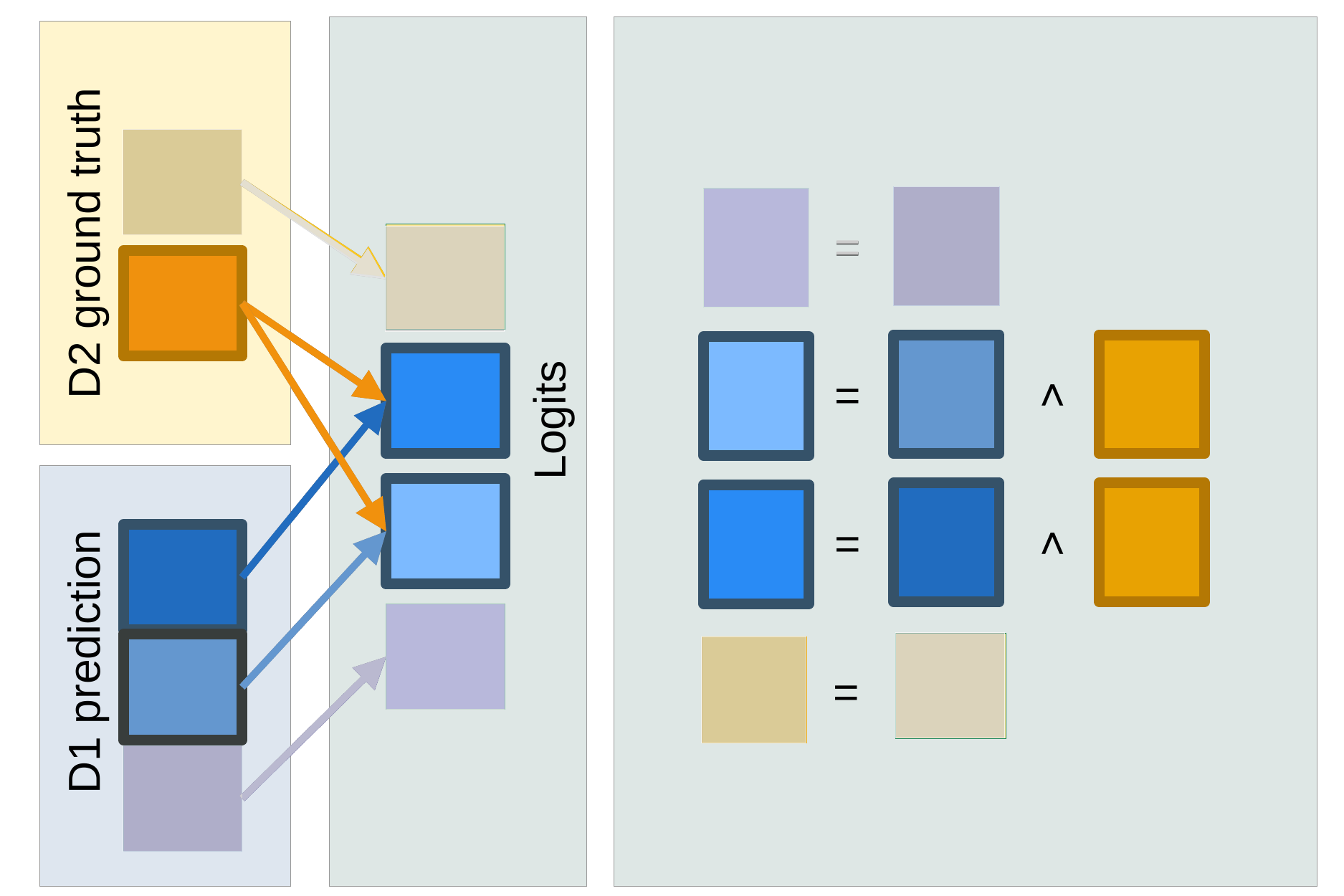}
 \caption{
   Universal pseudo-labels can be constructed
   by refining native ground-truth 
   according to dataset-specific predictions. 
   This improves the quality of pseudo-labels
   due to fewer conflicts with respect to
   a pure prediction-based approach 
   (cf.\ Fig.\ \ref{fig:pseudolabel-open-graph} and 
         Fig.\ \ref{fig:pseudo-labels}).
 }
 \label{fig:pseudolabel-gt-graph}
\end{figure}

Assume we wish to pseudo-relabel 
a pixel labeled as 
\mydscls{Vistas}{car}
according to a model 
trained on VIPER.
Figure \ref{fig:approach-taxonomy}
shows that the candidate universal classes
are \myunicls{pickup}, 
\myunicls{van} and \myunicls{car}.
We observe that the knowledge 
of the correct Vistas class
limits the possible VIPER predictions to 
\mydscls{VIPER}{truck}, 
\mydscls{VIPER}{van} and 
\mydscls{VIPER}{car}.
We therefore recover 
the desired pseudo-label
by only looking at
these three VIPER classes.
For example, the score 
$S^{\text{VIPER}}(\myunicls{pickup}
  \mid \vec x, \mydscls{Vistas}{car})$
would be determined 
by dividing the posterior  
of \mydscls{VIPER}{truck},
with the sum of posteriors 
for the three VIPER classes
that intersect the ground truth.
We illustrate the
pseudo-labeling based on 
dataset-specific ground-truth
in Figure \ref{fig:pseudolabel-gt-graph}.

Finally, we recover pseudo-labels 
by applying $\argmax$ 
over ensembled scores 
(\ref{eq:relabel})
of all universal classes 
that relate to the ground truth.
This allows to train universal models
through the standard NLL-loss
with respect to the pseudo-labels.
Note that this approach faces
the following two shortcomings: 
i) two-step training, and 
ii) pseudo-labels could be inaccurate
due to domain shift.
We compare pseudo-relabeled universal models
with our weakly supervised models
in section \ref{sec:results}.

\begin{figure*}[t]]
 \centering
 \begin{subfigure}{0.32\textwidth}
  \centering
  \includegraphics[width=0.95\textwidth]{figs/graphs/concat_model.pdf}
  \caption{naive concatenation (all classes)}
  \label{fig:eval-concat}
 \end{subfigure}
 \centering
 \begin{subfigure}{0.32\textwidth}
  \centering
  \includegraphics[width=0.95\textwidth]{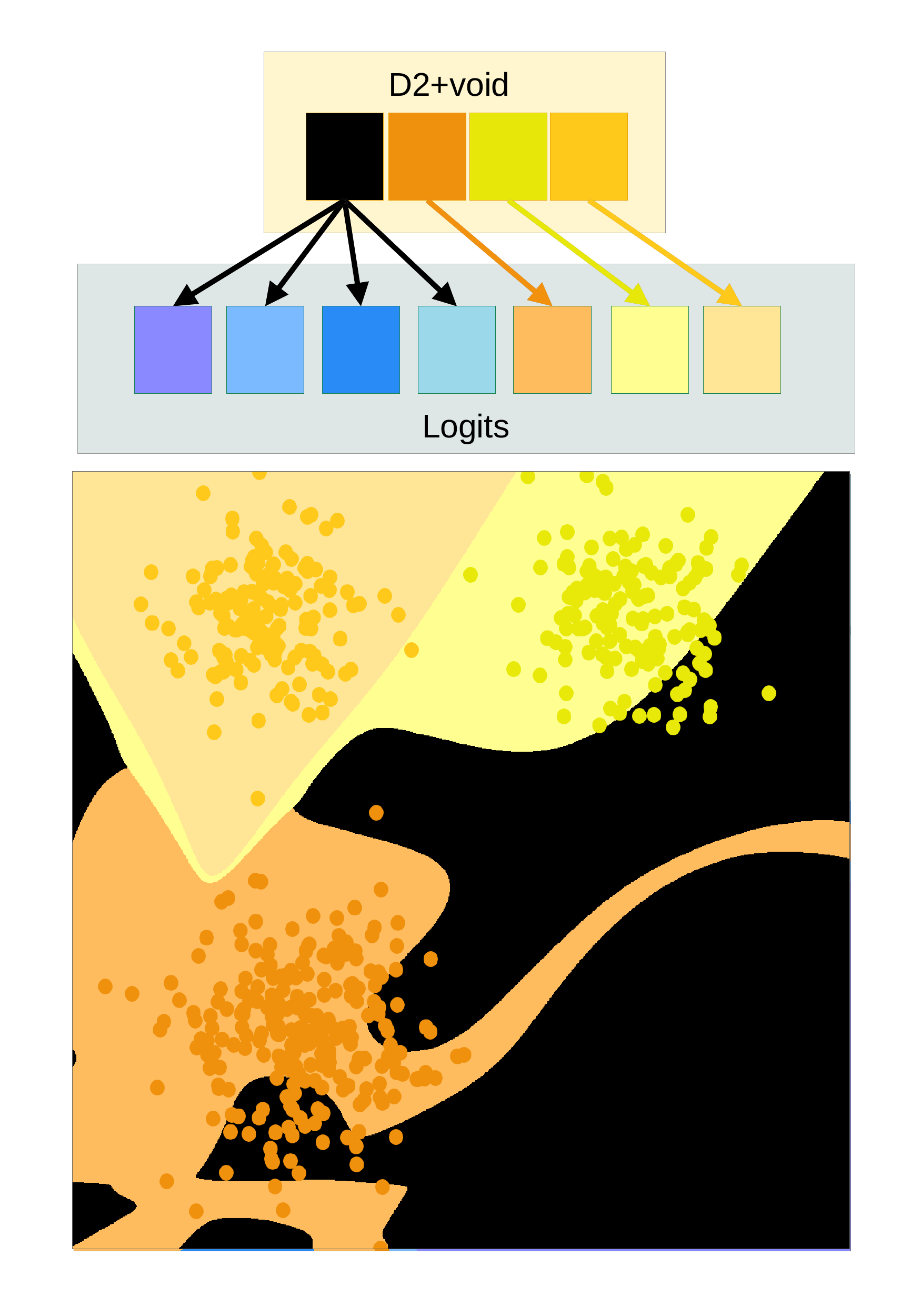}
  \caption{default scoring (D2)}
  \label{fig:eval-concat-sag}
 \end{subfigure}
 \begin{subfigure}{0.32\textwidth}
  \centering
  \includegraphics[width=0.95\textwidth]{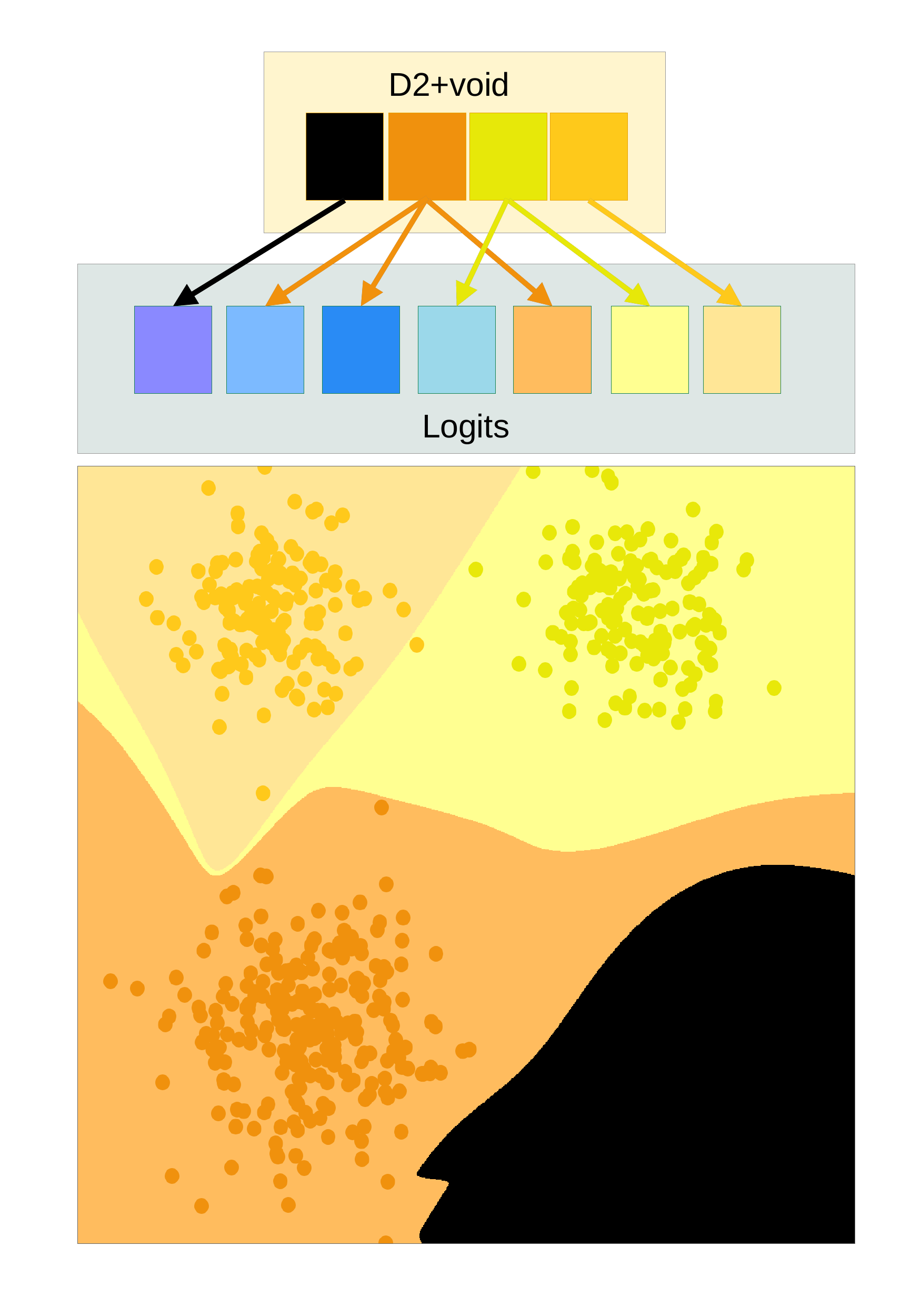}
  \caption{post-inference mapping (D2)}
  \label{fig:eval-concat-saw}
 \end{subfigure}
\caption{We train 
  the naive concatenation model 
  for the same toy problem
  as in Figure \ref{fig:baselines} (a),
  and subsequently evaluate on D2 (b-c).
  Default scoring assigns all foreign logits
  to the void class and thus triggers
  many undesired void predictions  
  that we show in black (b).
  A large proportion 
  of these void predictions
  swings to the correct class 
  when we score with 
  post-inference resolution (c)
  according to (\ref{eq:evalmap}).  
}  
\label{fig:eval-map}
\end{figure*}

\subsection{Evaluation of foreign predictions}
\label{ss:evalvoid}
Universal models are not easily evaluated
since the ground truth 
is almost never expressed  
in the universal domain.
We therefore evaluate 
our universal models 
by converting our universal predictions
to dataset-specific taxonomies
according to Equation (\ref{eq:proby}).
This conversion may get complicated,
since some universal logits 
may not correspond to any class 
of the considered evaluation dataset.
We shall refer to such universal logits
as foreign or extra-distribution logits.

We propose to deal with this issue
by extending each dataset-specific taxonomy
with a special void class
that maps to all universal classes
that are foreign to this particular taxonomy.
Note that such practice is compatible with
major multi-domain recognition competitions,
since they explicitly allow
post-processing of universal predictions
before submissions to individual benchmarks 
\cite{rvc22www}. 
We determine the posterior of the void class
according to the equation (\ref{eq:proby}),
just the same as we would do 
for any regular class.
Many semantic segmentation benchmarks 
accept void predictions 
and penalize them less strictly 
than incorrect within-domain predictions.
More precisely, void predictions
increase only the count of false negatives
while not affecting 
the count of false positives
for a given ground-truth class. 
We adopt the same convention
in our validation experiments.

\subsection{Post-inference mapping}
\label{ss:evalpostmap}

Different from universal models,
the baseline approaches may deliver
foreign predictions even 
in cases of good recognition.
For instance, a naive concatenation model
often recognizes \mydscls{COCO}{sky} as
\mydscls{Vistas}{sky}.
Partial merge models experience
similar failure modes, 
although less often.
We see that the baselines could improve 
through post-inference agreement 
with semantic relations.
We therefore propose to score 
each evaluation class $c^a_i$
by summing its posterior 
with posteriors of all intersecting 
foreign classes $c^b_j$:
\begin{equation}
  \mathbf{S}(c^a_i\mid\vec{x}^a) = 
    \mathbf{P}(c^a_i\mid\vec{x}^a) + 
    \sum_{c^a_i \not\perp c^b_j} 
      \mathbf{P}(c^b_j\mid\vec{x}^a)\;.
  \label{eq:evalmap}
\end{equation}
Figure \ref{fig:eval-map}
evaluates naive concatenation
according to post-inference mapping 
\eqref{eq:evalmap}
on the same toy problem
as in Figure \ref{fig:baselines},
and compares it to default scoring. 
We observe that post-inference mapping
leads to smoother decision surfaces
that promise better generalization.

\begin{table*}[htb]
  \centering
  \begin{tabular}{lcccccccccc}
    \multirow{2}{*}{Taxonomy} 
    & post-inference 
    && \multicolumn{2}{c}{(Vistas-City)} 
    && \multicolumn{2}{c}{(Vistas-WD2)} 
    && \multicolumn{2}{c}{(Vistas-ADE20k)}
    \\
    & mapping
    && City  & Vistas
    && WD2   & Vistas
    && ADE20k & Vistas
    \\
    \toprule
    single dataset & 
      N/A && 
      76.7  & 46.4  &&
      50.6  & 46.4 &&
      35.7 & 46.4
    \\
    [0.5em]
    \midrule
    
    per-dataset heads & no && 
      77.1 & 43.3 &&
      55.0 & 42.2 &&
      36.5 & 39.3
    \\
    per-dataset heads & 
      (\ref{eq:evalmap}) && 
      76.9  & 43.4  &&
      54.7  & 42.5 &&
      36.6 & 40.1
    \\
    [0.5em]
    naive concat & no && 
      76.8 & 44.4&&
      54.8 & 42.8 &&
      36.8 & 41.0
    \\
    naive concat & 
      (\ref{eq:evalmap}) && 
      76.8 & 44.4&&
      55.3 & 43.1&&
      36.8 & 42.2
    \\
    [0.5em]
    M2F naive concat & no && 
      78.5 & 52.8 &&
      60.3 & 51.5 &&
      N/A & N/A 
    \\
    M2F naive concat & 
      (\ref{eq:evalmap}) && 
      78.5 & 52.8 &&
      61.1 & 51.5 &&
      N/A & N/A
    \\
    [0.5em]
    partial merge & 
      no && 
      77.1 & 44.5&&
      54.5 & 44.0 &&
      37.3 & 41.1
    \\
    partial merge & 
      (\ref{eq:evalmap}) && 
      77.1 & 44.5&&
      54.7 & 44.1 &&
      37.4 & 41.8
    \\
    [0.5em]
    universal - pseudo & 
      (\ref{eq:proby}) && 
      76.9 & 44.9 &&
      55.5 & 45.5  && 
      34.1 & 43.7
    \\
    universal - NLL+ & 
      (\ref{eq:proby}) && 
      77.0 & 44.9&&
      56.2 & 44.4 &&
      37.4 & 42.8
      \\
    M2F universal & 
      (\ref{eq:nll-max}) && 
      77.4 & 52.9 &&
      60.5 & 52.1 &&
      N/A & N/A 
      \\
    \mytabsep
\end{tabular}
\caption{Evaluation of joint training 
  on Vistas-City,
  Vistas-WD2,
  and Vistas-ADE20k (mIoU).
  Post-inference mapping
  contributes noticeable improvement
  where dataset detection is difficult
  (Vistas-WD2, Vistas-ADE20k).
  Leveraging our universal taxonomy
  in different setups
  outperforms the baselines.
  Weakly supervised learning with NLL+
  generalizes slightly better 
  than two-step learning
  on pseudo-relabeled data
  in spite of much simpler training.
 }
 \label{table:concatvsnllplus}
\end{table*}

\subsection{Implementation details}

Our semantic segmentation experiments
reduce the computational complexity
of multi-domain training
by leveraging pyramidal 
SwiftNet models \cite{orsic20pr}.
Small experiments from 
\ref{ss:pairs-experiments} 
and \ref{ss:relabeled-city} 
involve a ResNet-18 \cite{he16cvpr} backbone.
Large experiments on 
multi-dataset collections from 
\ref{ss:manual-experiments}, 
\ref{ss:rvc-experiments} and 
\ref{ss:wd-experiments} involve 
a checkpointed DenseNet-161 backbone 
\cite{kreso20tits, huang19pami}.
We denote these two models
as SNp-RN18 and SNp-DN161.
All our M2F experiments 
use the ResNet-18 backbone.

We train our universal models
according to NLL+ loss 
(\ref{eq:nllplusdef}).
All other approaches
use the standard NLL loss.
Both losses prioritize 
pixels at semantic borders
\cite{zhen19aaai}. 
We use the Adam optimizer 
and attenuate the learning rate 
with cosine annealing
from  $5\cdot10^{-4}$ 
to $6\cdot10^{-6}$.
Our submissions to the RVC 2020 
benchmark collection \cite{rvc22www}
were trained 
for a fixed number of epochs.
All other experiments train 
only on the training splits 
of the involved datasets 
and use early stopping 
with respect to 
the average validation mIoU.
Please note that our 
validation experiments on WildDash 2 
split the training images and labels
into minitrain and minival according to
the alphabetical ordering. 
We place first 572 images into minival 
and the remaining images into minitrain. 

We augment training images 
with horizontal flipping, 
random scaling between 
0.5$\times$ to 2$\times$ 
and random cropping.
We use 768$\times$768 crops
except when training on 
the MSeg dataset collection 
where we use 512$\times$512 crops.
In experiments on ADE20k, Vistas
as well as MSeg and RVC collections,
we start augmentation by upsampling images 
so that the smaller side is 1080 pixels.
We apply the same preprocessing 
to the test images as well
and proceed by downsampling predictions 
to the input resolution. 
Most of our experiments 
use a single Tesla V100 32 GB
and set the common batch size 
with respect to the most 
memory inefficient model.
Our RVC submissions have been 
trained and evaluated on 6 such GPUs.
Our mini-batches prefer images with 
multiple class instances and rare classes, 
as well as encourage 
even representation of datasets.

\section{Results}
\label{sec:results}

We validate our universal taxonomies
by comparing them 
against the three baselines 
in \ref{ss:pairs-experiments},
and training on relabeled data 
in \ref{ss:manual-experiments}.
We demonstrate that NLL+ loss 
can learn unlabeled visual concepts 
in \ref{ss:relabeled-city}.
We evaluate our universal models 
on the RVC benchmark collections 
in \ref{ss:rvc-experiments},
and on the WildDash 2 benchmark 
in \ref{ss:wd-experiments}.

\subsection{Baselines}
\label{ss:pairs-experiments}

We consider joint training 
on three pairs of datasets:
i) Vistas - Cityscapes, 
ii) Vistas - WildDash 2 (WD2) 
and iii) Vistas - ADE20k.
In the first setup, 
all Vistas classes are either equivalent 
or subsets of their Cityscapes counterparts.
Hence, the universal taxonomy 
coincides with the Vistas taxonomy.
The second setup also pairs
two road-driving datasets, however 
here both datasets are quite diverse.
Consequently, we expect more competition 
between related logits 
from the two taxonomies. 
The third setup pairs datasets 
from different domains. 
The training batch sizes are 18
for road-driving setups
and 10 for Vistas-ADE20k.

Table \ref{table:concatvsnllplus}
compares our weakly supervised
universal approach (universal - NLL+)
with the three baselines 
and training on pseudo-labels
according to (\ref{eq:relabel}).
All approaches map foreign logits
to the void class, as explained in
section \ref{ss:evalvoid}.
The baseline with per-dataset heads  
\cite{masaki21itsc,fourure17neucom}
determines the joint posterior
of classes and datasets according to 
(\ref{eq:two-head}).
We observe that partial merge
outperforms naive concatenation,
and that naive concatenation
outperforms independent per-dataset heads.
Overall, all baselines profit from
post-inference mapping
(\ref{eq:evalmap}),
although we notice 
most improvement on 
Vistas-WD2 and Vistas-ADE20k.

The table indicates that our universal approaches 
outperform single-dataset training 
on all datasets except Vistas. 
This suggests that dense prediction
on Cityscapes, WD2 and ADE20k 
can succeed by trading off capacity
for additional supervision.
Furthermore, our
universal approaches outperform 
per-dataset heads, 
naive concatenation,
and partial merge.
The advantage is least evident 
in the Cityscapes-Vistas experiment. 
This effect is likely due to 
all Cityscapes images being acquired
with the same camera and 
in similar environment and weather.
This uniformity enables 
easy dataset detection and 
alleviates contention 
across related classes.
The table also indicates that NLL+ succeeds
to match and marginally  outperform
two-step training with universal pseudo-labels.
The advantage is especially prominent
in the case of Vistas-ADE20k, 
which is likely due to
the large domain shift
between the two taxonomies.
We note that pseudo-relabeling
would perform much worse without
having access to our universal taxonomy,
and that it would likely improve
if pseudo labels were provided
by our universal model.

Tables \ref{table:city-wd-cross} 
and \ref{table:wd-mvd-cross}
validate road-driving models from
Table \ref{table:concatvsnllplus}
on road-driving datasets
that were not seen during training:
CamVid (CV)
\cite{badrinarayanan17pami}, 
KITTI (KIT) \cite{geiger13ijrr}, 
BDD \cite{yu18bdd} and 
IDD \cite{varma19wacv}.
All baseline models were evaluated
with post-inference mapping.
All models map foreign predictions
to class void.
This comparison assesses
quality of the learnt features, 
and generalization potential 
of different approaches.

\begin{table}[t]
  \centering
  \begin{tabular}{lcccccc}
    Model & WD2 & CV & KIT & BDD & IDD \\
    \toprule
    single dataset & 50.6 & 73.9 & N/A & 58.7 & 59.5
    \\
     \midrule
    per-dataset & 42.8 & 74.4 & 55.3 & 58.0 & 41.6
    \\
    naive concat & 43.3 & 74.1 & 58.9 & 56.7 & 42.4
    \\
    partial merge & 43.8 & 73.9 & 59.4 & 57.0 & 43.0
    \\
    univ - pseudo & 42.4 & 73.3 & 58.1  & 57.5  &  42.6
    \\
    univ - NLL+ &  \textbf{43.9} & \textbf{75.3} & \textbf{60.5} & \textbf{58.0} & \textbf{42.8}
    \\
    \mytabsep
\end{tabular}
\caption{Cross-dataset evaluation 
  of joint training on Vistas-City
  (mIoU).
  We evaluate models from Table  \ref{table:concatvsnllplus}
  on WildDash 2 mini val, 
  CamVid test, KITTI val, BDD val, 
  and IDD val. 
 }
 \label{table:city-wd-cross}
\end{table}

Our universal approaches 
either outperform the baselines or are 
within the variance.
Weak supervision with NLL+
is consistently slightly better than 
two-step learning
on pseudo-relabeled data.
The contribution of our approach
is somewhat more prominent
in the Vistas-WD setup
where there is more competition
between semantically related logits. 
Note that KITTI performance
heavily depends on whether 
Cityscapes images 
were seen during training.
This is most likely due to significant
similarity between the two datasets.

\begin{table}[t]
  \centering
  \begin{tabular}{lcccccc}
    Model & City & CV & KIT & BDD & IDD \\
    \toprule
    single dataset & 76.7 & 73.9 & N/A & 58.7 & 59.5
    \\
    \midrule
    per-dataset & 69.3 & 72.8 & 52.6 & 58.1 & 41.5
    \\
    naive concat & 69.0 & 72.7 & \textbf{53.6} & 56.1 & 41.6
    \\
    partial merge& 69.8 & 72.4 & 53.5 & 57.1 & 41.9\\
    univ - pseudo & 71.2 & 74.5 & 52.6  & \textbf{59.2} & 42.4
    \\
    univ - NLL+  & \textbf{71.4} & \textbf{74.9} & 53.0  & 59.0 & \textbf{42.6}
    \mytabsep
\end{tabular}
\caption{Cross-dataset evaluation 
  of joint training on Vistas-WD2
  (mIoU).
  We evaluate models from Table  \ref{table:concatvsnllplus}
  on Cityscapes val, 
  CamVid test, KITTI val, 
  BDD val, and IDD val.  }
 \label{table:wd-mvd-cross}
\end{table}

\subsection{Mask-level recognition}

Table \ref{table:m2f} considers 
a Mask2Former architecture 
that was adapted for multi-dataset training 
according to Fig.\ \ref{fig:m2f}. 
We jointly train the proposed model
on Vistas and Cityscapes,
and present its performance 
along the NLL+ model 
from Table \ref{table:concatvsnllplus}.
Note that the two models 
are not comparable since  
the transformer decoder
has much more parameters
than ResNet-18.

\begin{table}[thb]
  \centering
  \begin{tabular}{@{\,}lc@{\:}cc@{\:}c@{\:}c@{\:}c@{\:}c@{\,}}
      
      & \multicolumn{2}{c}{within} & \multicolumn{5}{c}{cross-dataset} \\
      Model & City & Vistas & WD2 
      & CV & KITTI & BDD & IDD \\
    \toprule
    univ-NLL+
      & 77.0 & 44.9 & 43.9 & 75.3 & 60.5 & 58.0 & 42.8
    \\
    univ-M2F & 77.4 & 52.9 & 49.8 & 79.4 & 72.3  & 61.4 & 50.3
\end{tabular}
\caption{
  Multi-dataset training 
  of a mask-level model (M2F)
  for semantic segmentation.
  The M2F model succeeds to learn 
  a universal taxonomy and to deliver
  prominent cross-dataset generalization.
 }
 \label{table:m2f}
\end{table}

We observe that, qualitatively,
our universal M2F model 
behaves similarly to our pixel-based models
in previous experiments. 
Notably, within-domain performance 
suffers only a minor performance hit
with respect to single-dataset experiments
(0.2 pp on Cityscapes, 1.2 pp on Vistas).
As expected, cross-dataset performance 
exhibits sensitivity to domain shift, 
which is best observed on the IDD dataset. 
Nevertheless, our method succeeds to deliver 
reliable universal predictions 
after learning on dataset-specific labels.

\begin{table*}[thb]
  \centering
  \begin{tabular}{llccccccc}
    Eval. protocol & Taxonomy
      & ADE20k & BDD & City 
      & COCO & IDD & SUN & Vistas \\
    \toprule
    \multirow{2}{*}{Default}
      & MSeg & 28.4 & \textbf{61.9}
      & \textbf{77.0} & 34.3 & 47.4 
      & \textbf{47.3} & 28.7
    \\
      & universal - NLL+ & \textbf{35.6} & 60.4
      & 76.1 & \textbf{39.3} & \textbf{56.7} 
      & 46.9 & \textbf{44.2}
    \\
    [0.5em]
    \multirow{2}{*}{MSeg}
      & MSeg & 38.5 & \textbf{61.9} 
      & \textbf{77.0} & \textbf{40.9} 
      & \textbf{61.9} & \textbf{47.3} & 47.6
    \\
      & universal - NLL+ & \textbf{39.7} 
      & 60.4 & 76.1 & 40.3  
      & 58.1 &46.9  & \textbf{49.9}
    \mytabsep
\end{tabular}
\caption{Multi-domain experiments with
  SNp-DN161 on the seven MSeg datasets.
  We compare standard learning 
  on relabeled data (MSeg)
  with weakly supervised learning 
  on original labels (Universal - NLL+),
  according to two evaluation protocols
  (Default, MSeg).
  The models are evaluated 
  on validation subsets 
  of the seven datasets by mapping 
  all foreign predictions to class void.
 }
 \label{table:univ-mseg}
\end{table*}

\begin{table*}[b]
  \centering
  \begin{tabular}{lcccccccc}
    Taxonomy & mapping
      & ADE20k & BDD & Cityscapes 
      & COCO & IDD & SUN & Vistas \\
    \toprule
    naive concat 
      & no
      & 1.9 & 1.2
      & 0.0 & 6.7 & 0.1  
      & 1.4 & 0.2
    \\
    partial merge
      & no
      & 0.8 & 0.5
      & 0.0  & 2.3 & 0.9 & 0.2 & 0.0
    \\
    MSeg 
      & no 
      & 0.6 & 0.2
      & 0.0 & 0.3 & 0.5 & 0.3 & 0.5
    \\
    [0.5em]
    universal - NLL+ 
      & (\ref{eq:proby})
      & 0.2 & 0.3
      & 0.0 & 0.7 & 0.3 & 0.1 & 0.1
    \mytabsep
\end{tabular}
\caption{Percentage of foreign predictions 
  in multi-domain experiments with
  SNp-DN161 on the seven MSeg datasets
  according to the default protocol.
  We conjecture correlation
  between foreign predictions   
  and competition between overlapping classes.
  The overall incidence 
  of wrong predictions
  is around 15\%.
 }
 \label{table:mseg-void}
\end{table*}

\begin{figure*}[htb]
 \centering
\includegraphics[width=0.95\textwidth]{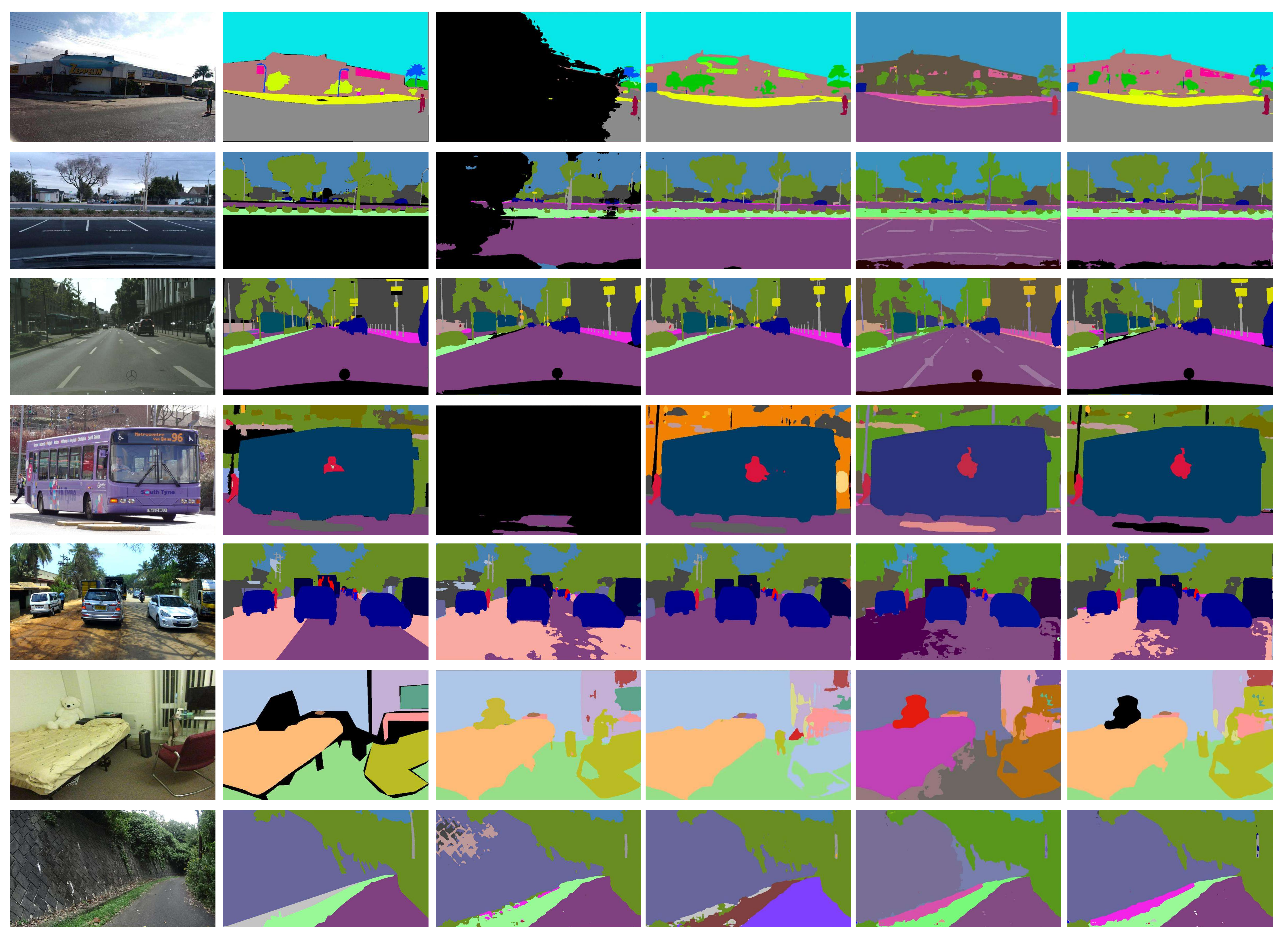}
 \caption{Qualitative evaluation
   of multi-domain models
   on MSeg images 
   (top to bottom): ADE20K, BDD, 
   Cityscapes, COCO, IDD,
   SUN RGBD, and Vistas.
   The columns present 
   the input image (column i),
   ground-truth labels (ii),
   as well as predictions of 
   naive concatenation (iii),
   the MSeg model (iv),
   and our model in universal (v)
   and native dataset-specific labels (vi). 
   Naive concatenation triggers
   void predictions (black)
   in contested classes.
   This is especially common 
   in images from diverse datasets
   (ADE20k, BDD, COCO).
   The MSeg model cannot recognize 
   some dataset-specific classes
   such as palm tree (ADE20k),
   tree (COCO) and parking lot (IDD).
   Our model recognizes universal classes
   and connects them with their
   dataset-specific counterparts such as
   marking-other $\rightarrow$ road 
   (BDD, Cityscapes) or
   curb $\rightarrow$ sidewalk 
   (ADE20k, Cityscapes).
   Multi-dataset training 
   enables outlier detection
   in dataset-specific predictions 
   (teddy bear, SUN RGBD).
}
\label{fig:mseg-images}
\end{figure*}

\subsection{MSeg collection}
\label{ss:manual-experiments}

We consider a large-scale collection
of the following seven datasets:
ADE20k \cite{zhou17cvpr},
BDD \cite{yu18bdd}, 
Cityscapes \cite{cordts16cvpr}, 
COCO \cite{lin14eccv}, 
IDD \cite{varma19wacv},
SUN RGBD \cite{song15cvpr}
and Vistas  \cite{neuhold17iccv}.
This collection is of particular interest
because it has been manually relabeled 
towards a custom taxonomy known as MSeg.
The MSeg taxonomy has 194 classes
that are consistently labeled
across all seven datasets
\cite{lambert20cvpr}.
However, in order to contain
the manual relabeling effort,
the MSeg taxonomy drops 61 
dataset-specific classes.
Pixels with these labels 
either get relabeled
 to a superset class
(e.g.\ \mydscls{COCO}{floor-wood} 
 is replaced with \mydscls{MSeg}{floor})
or outright ignored
(e.g.\ \mydscls{Vistas}{catch-basin}).

We construct our universal taxonomy
for the MSeg collection 
according to the procedure from
section \ref{ss:method-taxonomy}.
Our taxonomy consists of 
255 universal classes
that allow unambiguous prediction 
of all dataset-specific classes
in the entire collection.
We train our universal model
on original labels of the seven datasets
according to NLL+ loss and compare it
with standard learning on MSeg labels.

Please note that models 
trained on MSeg labels
are not in the same ballpark
with our universal models
who get to see only original labels.
Still, the comparison provides
some insight into the trade-off
between flexibility of weak supervision
and noiseless learning on 
opportunistically relabeled data.
We train all models on a single Tesla V100 
for 20 epochs with batch size 10.
We consider two evaluation protocols
since the MSeg taxonomy is not able
to recognize all evaluation classes
from the seven datasets.
The MSeg protocol only considers 
the 194 classes that are retained 
in the MSeg taxonomy \cite{lambert20cvpr},
while the default protocol 
considers all classes
from the particular datasets.
Note that experimental performance 
on Cityscapes, BDD and SUN RGBD
will not depend on the chosen protocol
since the MSeg taxonomy incorporates
all classes from these three datasets.

Table \ref{table:univ-mseg}
summarizes performance metrics
of the two models
according to the two protocols.
Our approach slightly underperforms
on the three datasets that are
fully preserved within the MSeg taxonomy.
This can be attributed either to noisy NLL+ training or to
depleted model capacity
due to learning 61 more classes.
Top section of the table 
presents evaluation 
according to the default protocol.
Our approach prevails by a wide margin
(from 5 to 16 percentage points) 
on all four datasets that are 
not fully represented 
by the MSeg taxonomy. 
Bottom section of the table
presents evaluation
according to the MSeg protocol.
Somewhat surprisingly,
our approach remains competitive
in spite of inferior supervision,
especially on the three datasets
with the largest individual taxonomies
- ADE20k (150), COCO (133), and Vistas (65).
Differences across the two dataset groups  
could be due to our training procedure 
allocating more model capacity 
to datasets with more rare classes.
This hypothesis is based on observation
that dropped classes
are comparatively rare.
Overall, the table seems to suggest 
that harnessing flexibility 
of weakly supervised universal taxonomies
may be more cost-effective than
strengthening supervision 
by manual relabeling,
especially when considering 
options for including 
future datasets.
Figure \ref{fig:mseg-images} 
presents a qualitative comparison
of different models trained on 
the MSeg collection.

Table \ref{table:mseg-void} studies 
the frequency of void predictions 
for several models trained 
on the MSeg dataset collection.
We observe that our universal approach 
and the MSeg approach tend to produce
less void predictions 
than the two baselines. 
The naive concatenation model 
performs the worst,
while the model with partially merged
classes sits somewhere in-between.
This suggests that foreign predictions
are related to the competition 
between related visual concepts
from different datasets.

The presented experiments also show
that the amount of void predictions 
is not uniform across datasets. 
This indicates that 
some of the datasets are easy to detect 
due to uniform image acquisition  
(camera, weather, location etc.).
Furthermore, batch creation 
and dataset sizes might introduce
various kinds of bias into the models.
In the end, no approach 
is uniformly better or
worse on all datasets.
Furthermore, limitations of
dataset-specific evaluation
can favour overfitting 
to specific scenery 
(e.g.\ KITTI) 
or noisy labels (e.g.\ BDD).
These challenges could be addressed
by collecting novel test images and 
annotating them with ground-truth labels
from several dataset-specific taxonomies.
Such evaluation would compel the models 
to recognize visual concepts
in unusual surroundings
and thus bring us closer
to universal computer vision.

\subsection{Novel concepts}
\label{ss:relabeled-city}

\begin{figure}[htb]
  \centering
  \includegraphics[width=0.45\textwidth]
    {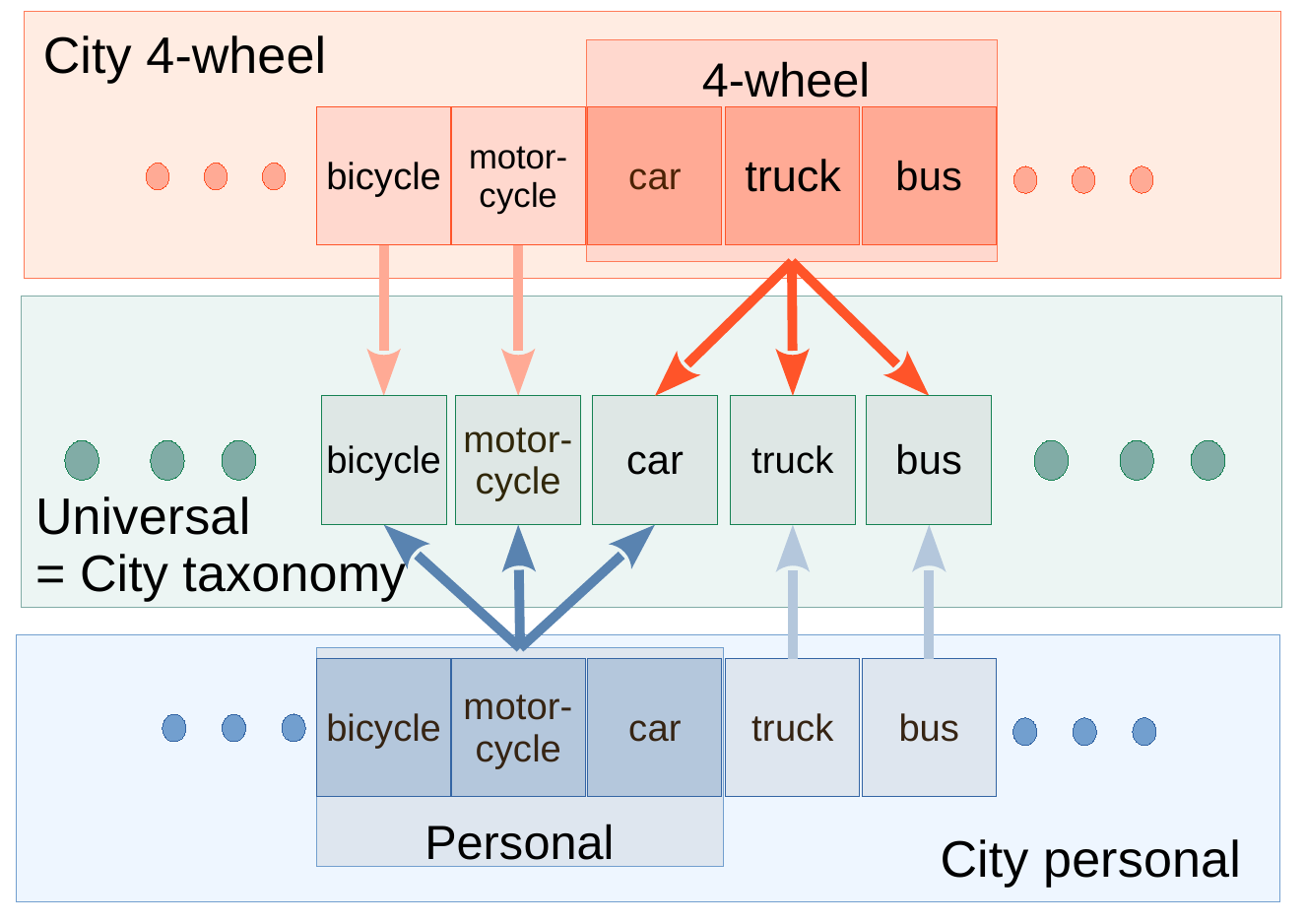}
  \caption{We relabel Cityscapes train
    into two subsets. 
    Subset "City 4-wheel" 
    groups cars, buses and trucks.
    Subset "City personal" groups 
    cars, bicycles and motorcycles. 
    Universal taxonomy for the two splits 
    includes all 19 Cityscapes classes.
  }
  \label{fig:novel-city}
\end{figure}

The toy example in 
Figure \ref{fig:toyoverlaps}
suggests that NLL+ 
should be able to learn novel concepts 
that are not explicitly labeled 
in any of the datasets. 
We test this hypothesis 
by splitting Cityscapes train 
and relabeling the two subsets
with overlapping taxonomies
according to Figure \ref{fig:novel-city}.
The two subsets have approximately
equal size and class distribution.
The subset "City-4wheel" includes images 
from cities between Aachen and Hanover,
and groups trucks, buses and cars into
the class
\mycls{four-wheels-vehicle}.
The subset "City-personal" includes 
the remaining images and groups 
cars, bicycles and motorcycles into 
the class
\mycls{personal-vehicle}.
Both subsets have 17 classes. 
Note that cars are never labeled 
as a standalone class, 
whereas buses, trucks, 
motorcycles and bicycles  
occur as standalone classes 
in only one of the two splits. 

In this setup, the universal taxonomy
consists of the standard 19 classes
of the Cityscapes dataset
as shown in Figure \ref{fig:novel-city}.
Each standalone class 
is mapped to itself.
The two composite classes are
mapped as follows:
\mycls{four-wheel-vehicle} $\mapsto$
\myunicls{car} $\cup$ 
\myunicls{bus} $\cup$  
\myunicls{truck}, and
\mycls{personal-vehicle} $\mapsto$
\myunicls{car} $\cup$ 
\myunicls{bicycle} $\cup$  
\myunicls{motorcycle}.

We validate our method (NLL+) against 
four other multi-dataset approaches.
The NLL baseline simply ignores 
all composite labels
and therefore should not be able
to recognize the class car.
Naive concatenation and partial merge
are the two baselines from section
\ref{ss:baselines}.
NLL-max is a modification of our approach
that replaces the summation 
in (\ref{eq:proby}) 
with the probability of 
the most likely universal class.
We also include the model trained 
on the standard Cityscapes train dataset,
and denote it as oracle since it receives 
more supervision than the other approaches.
The oracle exposes the handicap 
due to weak supervision 
and shows the upper bound
of multi-dataset training. 
We train all models for 250 epochs
with batch size 14 
on a single GTX1080
by oversampling images 
with instances of class 
\mycls{train}.

\begin{table}[htb]
\begin{center}
\begin{tabular}{l@{\quad}
    r@{\quad}r@{\quad}r@{\quad}
    r@{\quad}r@{\quad}r}
  Model & \faCar & \faBus & \faTruck & 
      \faMotorcycle & \faBicycle & mIoU 
  \\
  \toprule
  NLL baseline &
    0 & 54.2 & 43.9 & 32.5 & 60.3 & 58.7
  \\
  NLL-max  &
    0    &  9.6 & 40.8 &  1.6 & 75.7 & 61.8
  \\
  naive concat & 91.1 & 61.4 & 42.2  & 39.0 & 72.3 & 67.6
  \\
  partial merge & 92.4 & 54.3 & 55.1 & 39.0 & 74.3 & 71.4
  \\
  NLL+  &
    \textbf{93.6} & \textbf{73.3} & \textbf{66.6} & \textbf{46.4} & 75.4 & \textbf{74.3}
  \\
  \midrule
  M2F concat & 92.9  & 53.1  & 60.7  & 35.4  & 76.7  & 72.5 \\
  M2F universal  & 95.1 & 77.7 & 75.4 & 54.7 & 77.6 & 77.7 \\
  \midrule
  NLL oracle  &
    94.4 & 82.9 & 72.9 & 62.2 & 76.5 & 76.2 \\
  M2F oracle & 95.3 & 84.2 & 76.5 & 61.0 & 78.8 & 77.6
  \mytabsep
\end{tabular}
\caption{Experimental validation 
  of multi-dataset training 
  on relabeled Cityscapes
  according to the setup from 
  Figure \ref{fig:novel-city}.
  NLL baseline ignores all composite labels,
  while NLL oracle trains 
  on original Cityscapes labels 
  in all images.
  Our universal taxonomy 
  becomes even more advantageous
  in M2F experiments where it performs 
  au pair with the oracle.
}
\label{table:novelconcepts}
\end{center}
\end{table}

Table \ref{table:novelconcepts} shows
that NLL-max delivers poor performance
and that it can not detect cars.
A closer look revealed that NLL-max
is prone to overfitting to universal classes
that receive direct supervision 
through the other dataset.
More concretely,
46\% of training pixels at cars 
were recognized as buses 
while the rest 
were recognized as motorcycles. 
Note that this effect can not arise
in setups without overlapping classes
\cite{zhao20eccv}. 
The two baselines 
succeed to recognize cars 
due to post-inference mapping
(\ref{eq:evalmap}),
however they underperform 
with respect to NLL+.
We attribute the success of our method
to learning with partial labels,
as well as to principled formulation
of the weakly supervised objective. 
We observe most improvement on classes 
with only half standalone labels
(\mycls{bus}, \mycls{truck}, 
 \mycls{motorcycle}, \mycls{bicycle}). 
These improvements arise
due to contribution 
of learning with partial labels,
and absence of competition 
between related logits. 
This competition is especially influential 
in this particular setup
since there is very little domain shift 
between the two training splits.

\begin{figure*}[htb]
 \centering

\includegraphics[width=0.95\textwidth]{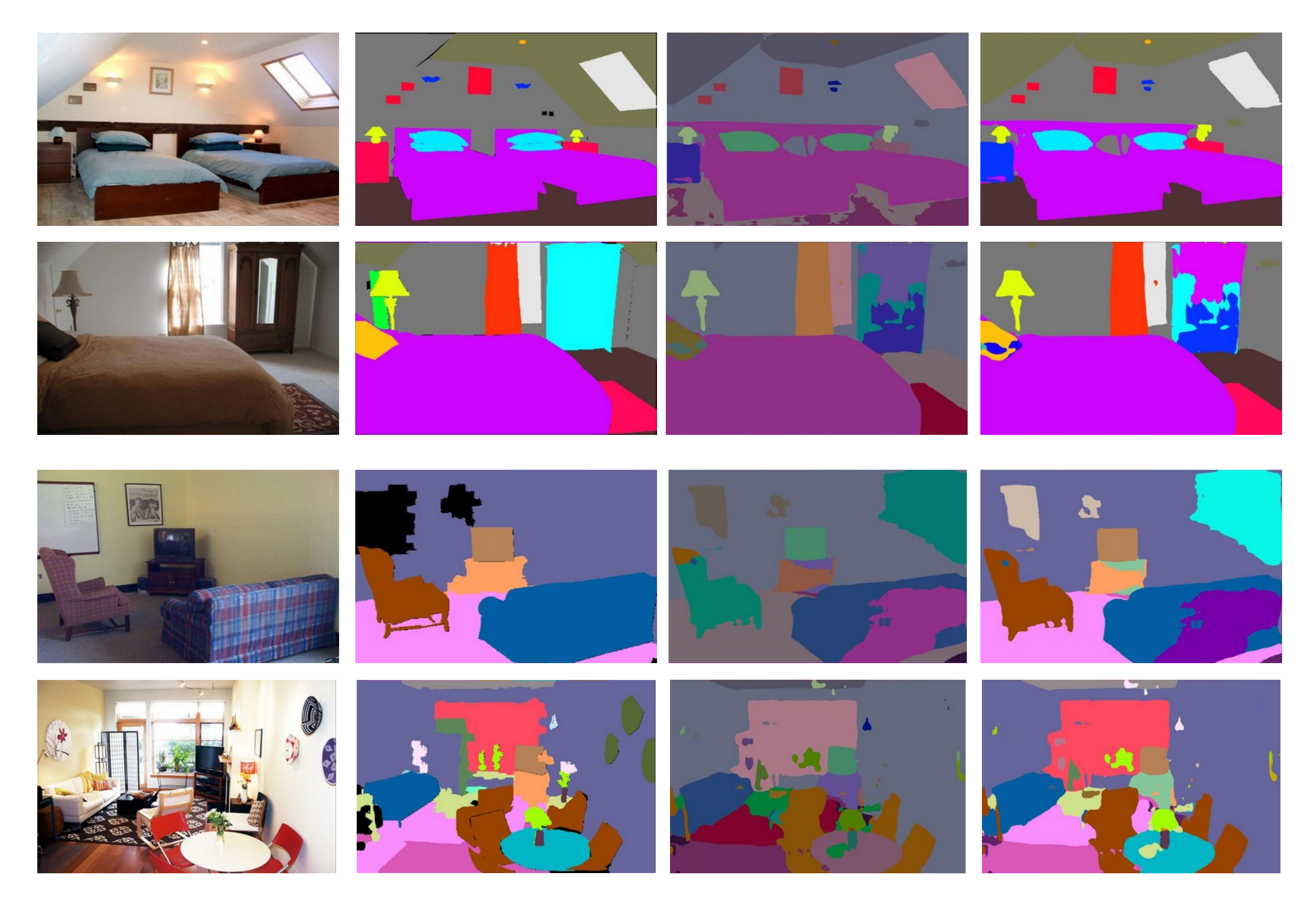}
 \caption{We demonstrate the ability of our universal models to learn
 novel concepts on ADE20K (top two images) 
 and COCO (bottom two images). The columns show the input
 image, the dataset-specific ground truths, the universal
 predictions and the dataset-specific prediction.
 Both ADE20K and COCO 
 contain classes \mycls{floor} (dark brown in ADE20K, dark purple in COCO)
 and \mycls{carpet} (red in ADE20k, pink in COCO). The 
 difference between the two datasets is that
 ADE labels carpeted floors with the class \mycls{floor} (dark brown, row 2)
 while COCO labels carpeted floors with the class \mycls{carpet} (pink, row 3).
 Our universal taxonomy contains classes \myunicls{floor} (purple),
 \myunicls{carpeted floor} (gray) and
 \myunicls{carpet} (dark red). The mappings are: \mycls{ADE20k-floor} $\mapsto$
\myunicls{floor} $\cup$ 
\myunicls{carpeted floor},
\mycls{ADE20k-carpet} $\mapsto$
\myunicls{carpet}, \mycls{COCO-floor} $\mapsto$
\myunicls{floor}, and
\mycls{COCO-carpet} $\mapsto$
\myunicls{carpet} $\cup$ 
\myunicls{carpeted floor}. Using our approach,
 we are able to train the model to recognize carpeted floors as 
 a standalone visual concept that can be connected to the
 correct dataset-specific class.
}
\label{fig:novel-concepts}
\end{figure*}

Qualitative experiments from Figure \ref{fig:novel-concepts}.
demonstrate the ability of our universal models 
to learn novel concepts in ADE20K and COCO.

\subsection{RVC challenge}
\label{ss:rvc-experiments}

Robust Vision Challenge is 
a prominent recent competition
in multi-domain computer vision   
\cite{rvc22www}.
The challenge promotes 
real-world usability
by requiring the submitted models 
to perform well on multiple benchmarks
akin to combined events in athletics.
The challenge considers several tasks 
in dense reconstruction and recognition,
however here we consider only
the semantic segmentation track
that requires submissions to
ADE20k, 
Cityscapes, 
KITTI \cite{geiger13ijrr} 
(only in 2020), 
Vistas,
Scannet \cite{dai17cvpr}, 
Viper \cite{richter17iccv}
and WildDash 2 \cite{zendel18eccv}.


RVC submission rules aim 
to reward cross-domain competence
and to discourage brute-force solutions 
that thrive by overfitting to dataset bias.
Thus, all benchmark submissions 
have to be inferred by a single model 
with less than 300 training logits.
The predictions must reside 
in a universal label space
that is dataset-agnostic.
In other words, explicit dataset recognition 
and dataset-specific sub-solutions 
are outright prohibited.
These requirements disqualify 
naive concatenation and per-dataset heads.
Note that the universal predictions 
have to be projected to 
the particular dataset-specific taxonomy
for each benchmark submission.
This can be implemented
by multiplying universal predictions
with a matrix whose rows correspond 
to dataset-specific classes.

The challenge allows training on 
any publicly available data.
Still we choose to train 
only on the seven training subsets 
due to overwhelming 
computational complexity.
Table \ref{tbl:rvctrain} presents
the basic RVC training setup
that involves quarter trillion 
labeled pixels, 35$\times$ 
more than in Cityscapes.

\begin{table}[htb]
\centering
\begin{tabular}{llrr@{ - }lr@{$\pm$}l}
Dataset & content 
  & \multicolumn{1}{c}{images}
  & \multicolumn{2}{c}{classes}
  & \multicolumn{2}{c}
    {$\sqrt{\mathrm{pixels}}$} 
  \\
\hline
ADE20k      & photos   &  22210
  & 150 & 150& 460  & 154\\
City   & driving  &   3475 
  & 28 & 19  & 1448 & 0   \\
KITTI        & driving  &    200 
  & 28 & 19  & 682  & 1   \\
VIPER        & GTA-V &  18326 
  & 32 & 19  & 1440 & 0  \\
ScanNet      & interior &  24902 
  & 40 & 20  & 1109 & 78  \\
Vistas       & driving  &  20000 
  &  65 & 65 & 2908 & 608 \\
WD2          & driving  &   4256 
  & 26 & 20  & 1440 & 0   
\end{tabular}
\caption{Basic RVC training setup.
  The columns show the number 
  of annotated non-test images,
  the number of training 
  and test classes,
  as well as mean and standard deviation 
  of the square root 
  of the number of pixels
  ($\sqrt{HW}$) 
  across images.
  }
\label{tbl:rvctrain}
\end{table}

We construct the universal taxonomy
for the basic RVC collection
according to the procedure
from section \ref{ss:method-taxonomy}.
Our source code for mapping 
each dataset-specific class
to the corresponding subset 
of 192 universal classes
is available online \cite{bevandic22github}.
There is one case where we stray 
from our procedure 
in order to prevent 
proliferation of twin classes. 
Vistas, KITTI and Cityscapes 
label vehicle windows as vehicles, 
while VIPER labels those pixels with 
what is seen through the glass. 
Consistent application of rule 3 
would require forking each VIPER class
(e.g \mycls{person-through-glass}
or \mycls{vegetation-through-glass}). 
Instead, we introduce 
simplifying assumptions such as
\mydscls{VIPER}{car} = \mydscls{Vistas}{car} 
and 
\mydscls{Vistas}{car} $\perp$ 
\mydscls{VIPER}{person} 
in order to reduce the footprint 
of the universal models.

We increase the model capacity by setting 
the upsampling width to 384 channels.
We leverage checkpointed backbones 
\cite{bulo18cvpr,kreso21tits}
and custom backprop through NLL+ loss 
in order to allow training on batches 
of 8$\times$768$^2$ crops per 32GB GPU.
We perform the training on 6 
V100 32 GB GPUs for 100 epochs. 
We minimize boundary modulation
\cite{zhen19aaai}
on ScanNet images in order to 
alleviate noisy labels.
The total complexity 
of our training setup 
is around 4 exaFLOP
or four days on our hardware.
Our benchmark submissions are
ensembled predictions  
on original images and 
their horizontal reflections
across six scales.
Inference took one day 
on our hardware.

\begin{table}[htb]
  \centering
  \begin{tabular}{l@{\;\;}
      c@{\;\;}c@{\;\;}c@{\;\;}
      c@{\;\;}c@{\;\;}c@{\;\;}c}
    Model & 
    ADE & City & KIT & MV & SN & VIP & WD 
    \\
    \toprule
     MSeg 
     \cite{lambert20cvpr}
     &
       \textbf{33.2} & \textbf{80.7} & 
       62.6 & 34.2 & 48.5 & 40.7 & 35.2 
     \\ 
    [0.5em]
    SNp\_rn152 &
     31.1 & 74.7 & 63.9 & 
     40.4&
     \textbf{54.6}& 
     62.5&
     45.4
    \\
    \midrule
      SNp\_dn161f &  
      30.8 & 77.9 & \textbf{68.9} & \textbf{44.6} & \textbf{53.9} & \textbf{64.6} &\textbf{46.8} \\
  \end{tabular}
  \caption{Performance evaluation on the
    RVC 2020 semantic segmentation track. 
    We submit the same model to 
    the seven benchmarks:
    ADE20k (ADE), Cityscapes (City), 
    KITTI (KIT), Vistas (MV), 
    ScanNet (SN), VIPER (VIP) and
    WildDash 2 (WD).
    The model from the bottom section
    has been submitted 
    after the challenge deadline.
  }
  \label{table:rvc-results}
\end{table}
\setlength{\tabcolsep}{4pt}
\begin{table*}[b]
\begin{center}

\resizebox{\textwidth}{!}{%
\begin{tabular}{lcccccccccc}
  \multirow{2}{*}{Model}
  & \multirow{2}{*}{Meta Avg} &
  & \multicolumn{4}{c}{Classic} 
  && \multirow{2}{*}{Negative}
  && \multirow{2}{*}{Hazards}
\\
\cmidrule{4-7}
  & mIoU cla
  &
  & mIoU cla
  & iIoU cla
  & mIoU cat
  & iIoU cat
  &
  & mIoU cla
  &
  & avg. impact\\
  
\toprule
  \multicolumn{1}{l}{EffPS\_b1bs4sem\_RVC \cite{mohan2021ijcv}} & 32.2 && 35.7 & 24.4 & 63.8 & 56.0 && 20.4 && -8\%\\
  \multicolumn{1}{l}{MSeg\_1080 \cite{lambert20cvpr}} & 35.2 && 38.7 & 35.4 & 65.1 & 50.7 && 24.7 && -12\%\\
  \multicolumn{1}{l}{seamseg\_rvcsubset \cite{porzi19cvpr}} & 37.9 && 41.2 & 37.2 & 63.1 & 58.1 && 30.5 && -13\%\\
  \multicolumn{1}{l}{UniSeg \cite{kim22eccv}} & 39.4 && 41.7 & 35.3 & 65.8 & 57.4 && 34.8 && -13\%\\
  \multicolumn{1}{l}{FAN\_NV\_RVC \cite{xiao22arxiv}} & 
  47.5 && 50.8 & 44.0 & \textbf{74.2} & 67.5 && 34.4 && -8\%\\
  \multicolumn{1}{l}{MIX6D\_RVC \cite{liu22arxiv}} & 
  \textbf{48.5} && 51.2 & \textbf{46.5} & 72.4 & 66.1 && \textbf{40.8} && -8\%\\
\midrule
  \multicolumn{1}{l}{SNp\_RN152pyr\_RVC
  \cite{orsic20arxiv}} & 
  45.4 && 48.9 & 42.7 & 70.1 & 64.8 && 32.5 && -7\% \\
  \multicolumn{1}{l}{SN\_DN161\_fat\_pyrx8 \cite{Bevandic_2022_WACV}}& 
  46.8 && 51.0 & 43.9 & 71.4 & 65.5 && 32.6 && -8\% \\
  \multicolumn{1}{l}{UNIV\_CNP\_RVC\_UE} & 46.9 && \textbf{51.6} & 45.9 & 72.8 & 67.5 && 29.0 && \textbf{-6\%}\\
\bottomrule
\end{tabular}
}
\caption{Current WildDash 2 leaderboard.
  All submissions have been trained 
  on multiple datasets.
  The bottom section presents our RVC models.
  The first two entries
  correspond to our models from 
  Table \ref{table:rvc-results}
  (we had to shorten them there
   in order to fit the tight layout).
  We also include two unpublished
  submissions to RVC 2022 
  (FAN\_NV, MIX6D).
  FAN\_NV was trained
  with 10$\times$ hardware
  with respect to our models.
}
\label{table:wd_bench_results}
\end{center}
\end{table*}

Table \ref{table:rvc-results}
presents performance evaluation on RVC 2020.
The top section shows 
the two valid submissions
to RVC 2020 semantic segmentation track
\cite{rvc22www}.
Our submission outperforms the model 
trained on the MSeg taxonomy 
\cite{lambert20cvpr}
due to being able to recognize all classes
from the particular benchmarks. 
Our approach succeeded due to being
able to adapt to the task at hand 
without requiring any manual relabeling. 
The bottom section shows 
our improved model that was 
trained as explained 
in the previous paragraphs.
We show it separately because
it was submitted 
to the seven benchmarks
after the deadline for RVC 2020.

\begin{figure*}[b]
    \centering 
    \includegraphics[width=0.3\textwidth]{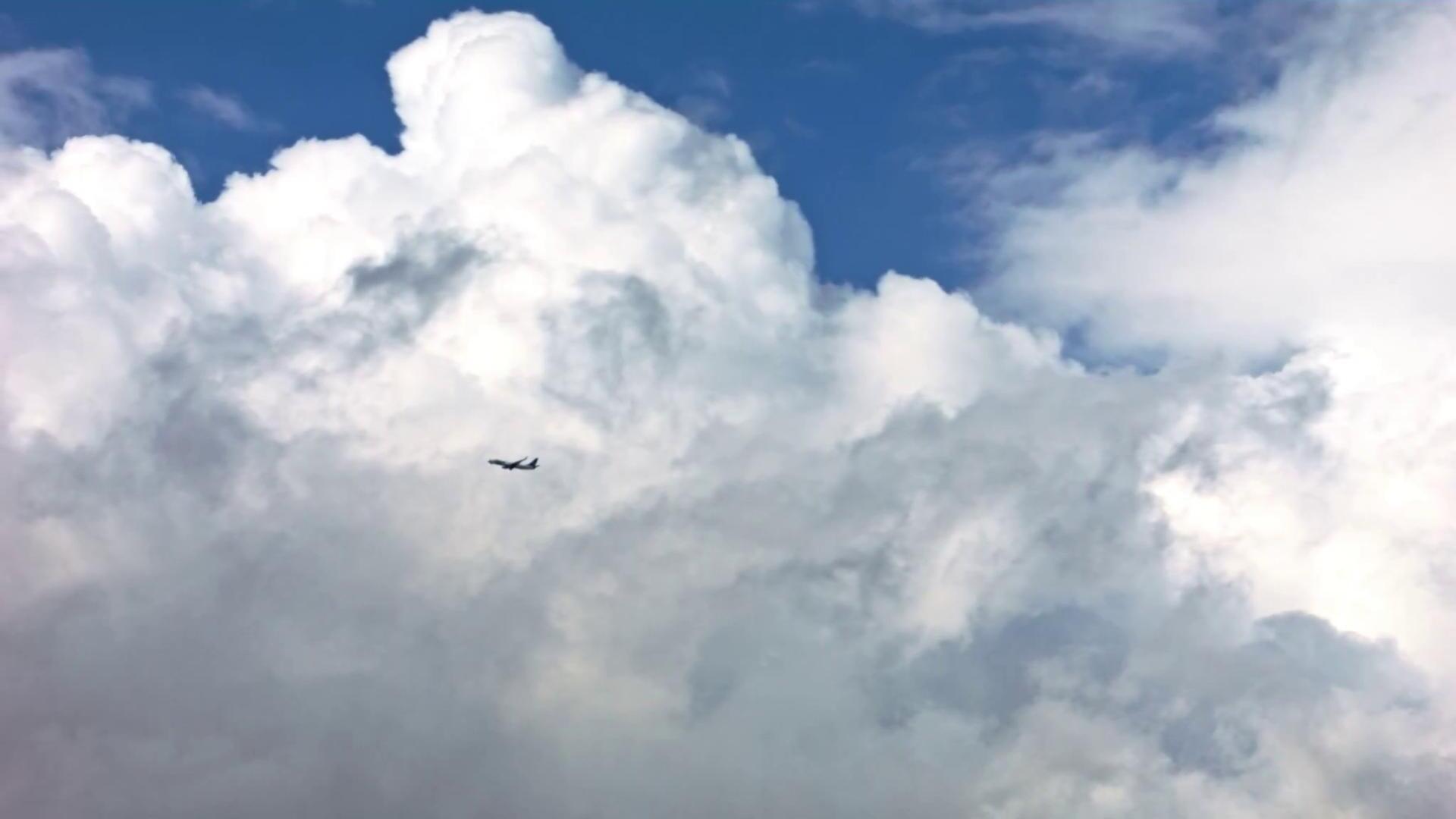}
    \includegraphics[width=0.3\textwidth]{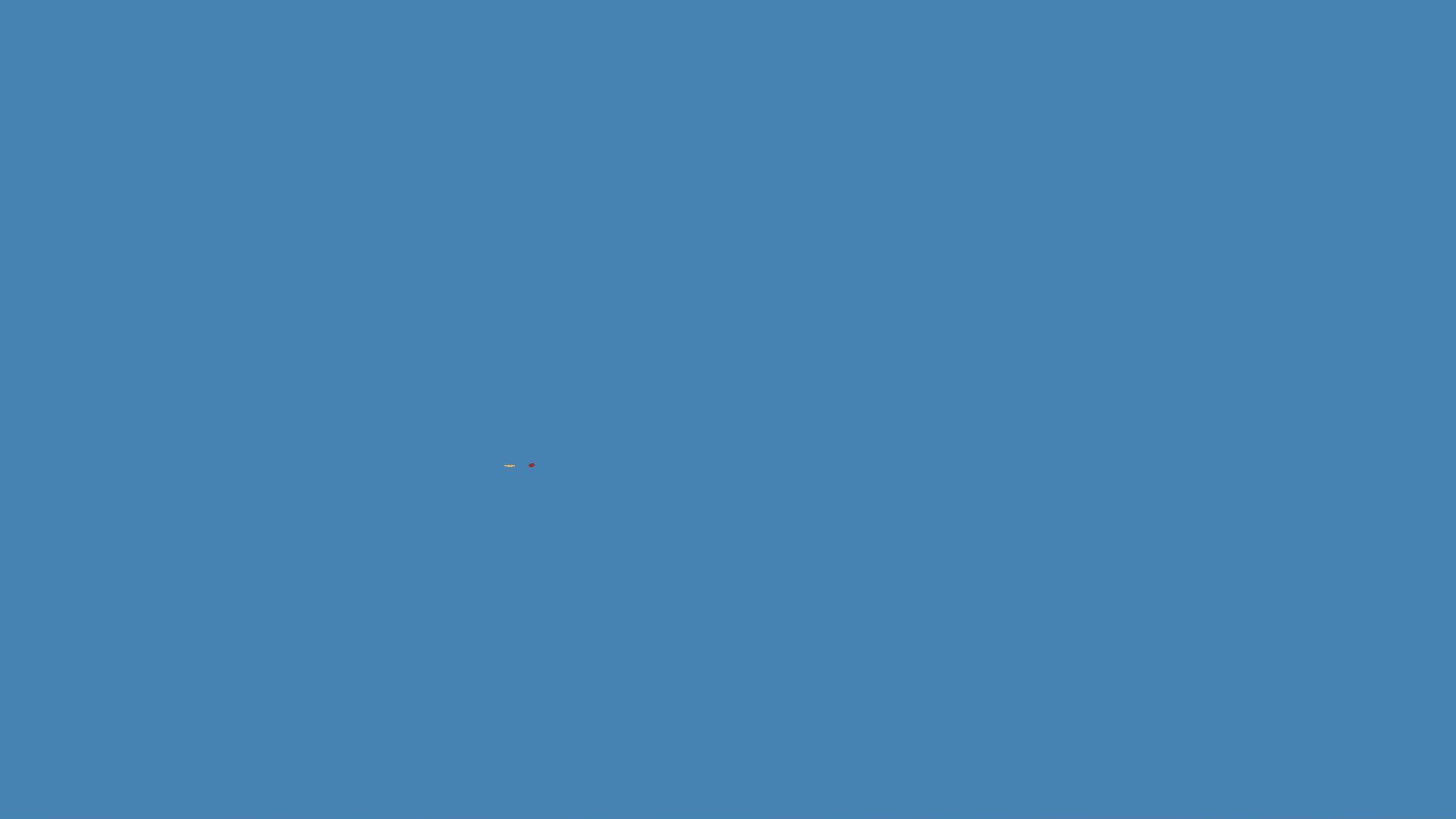}
    \includegraphics[width=0.3\textwidth]{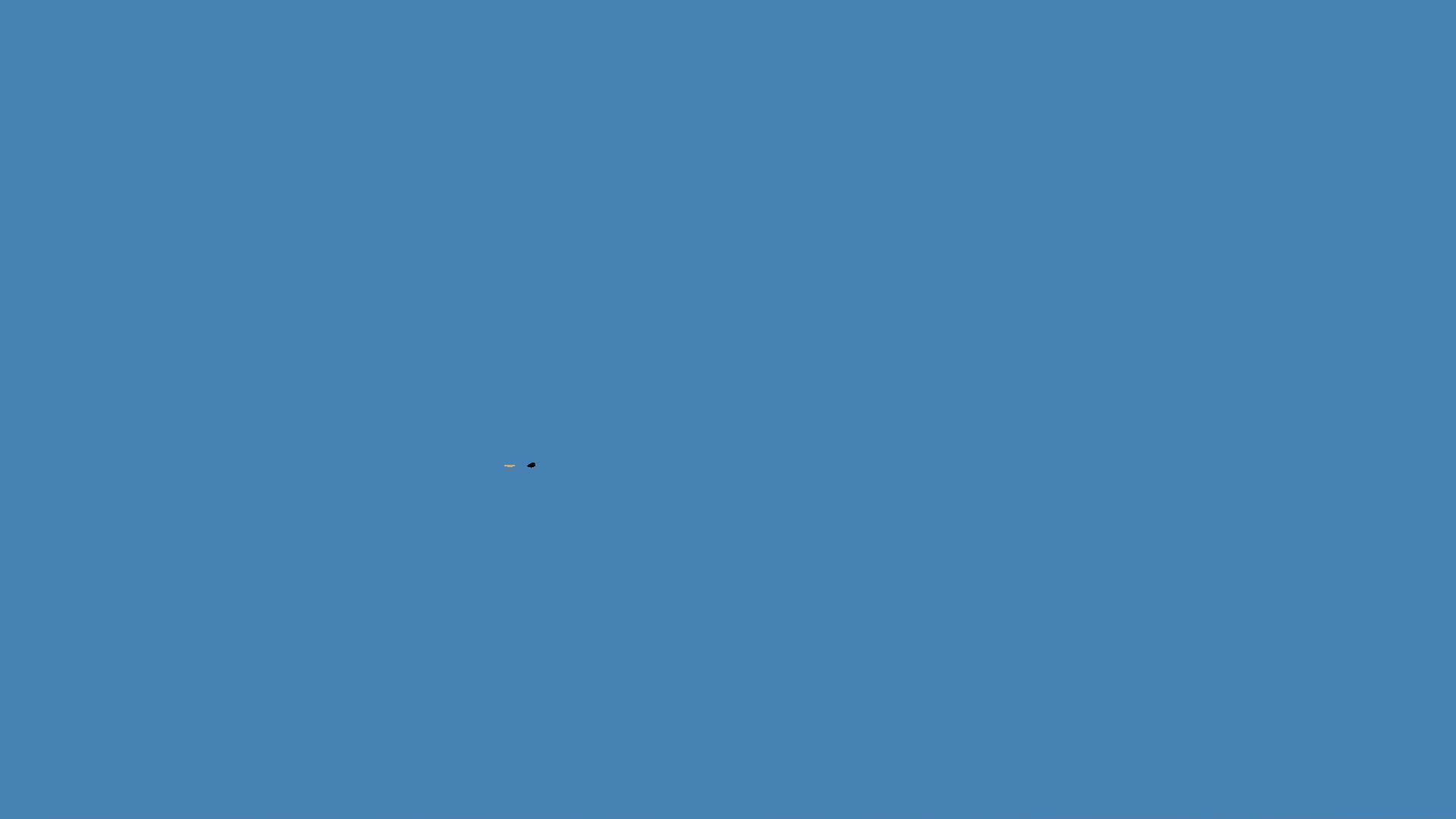}
    \\
    \includegraphics[width=0.3\textwidth]{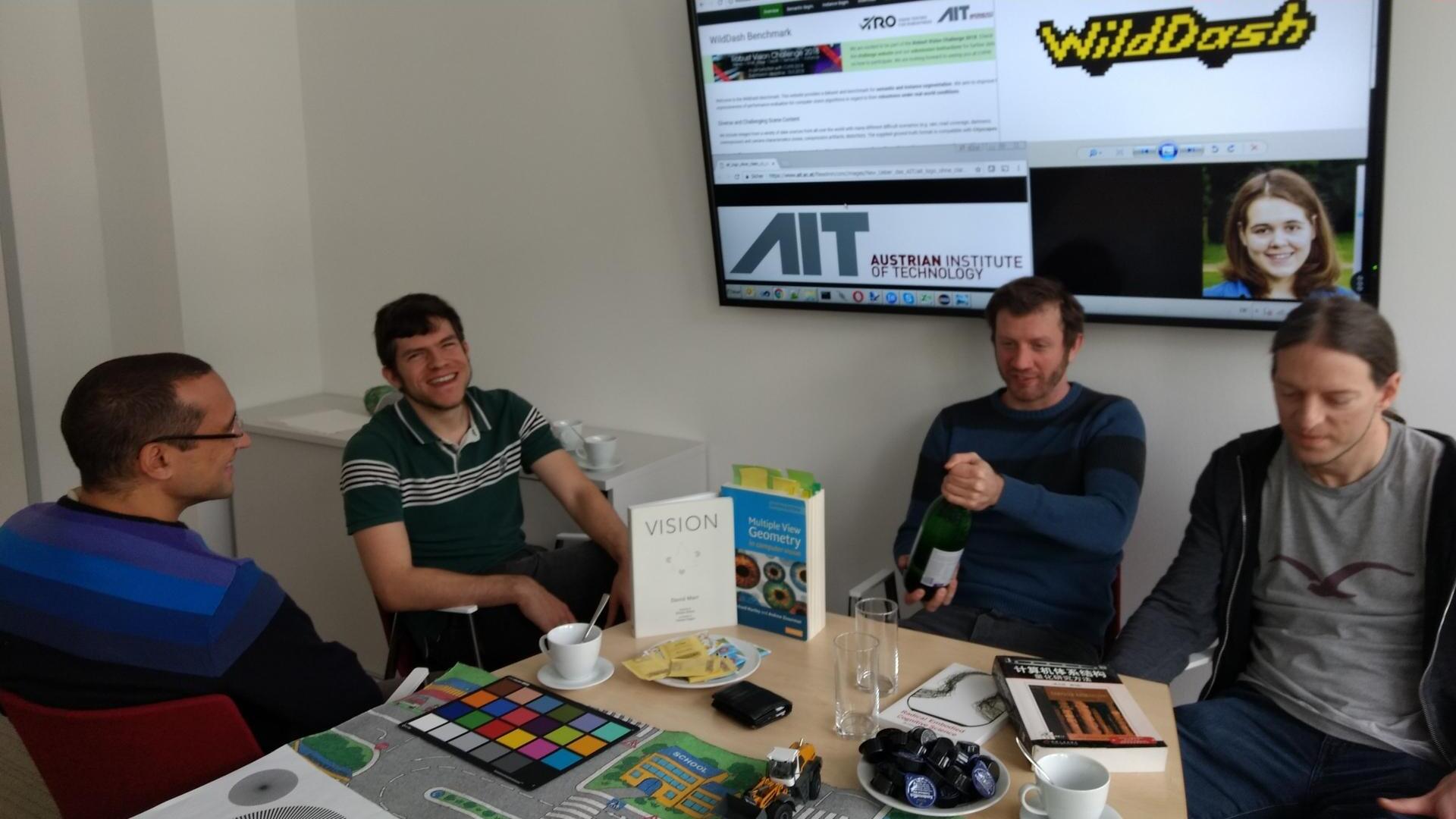}
    \includegraphics[width=0.3\textwidth]{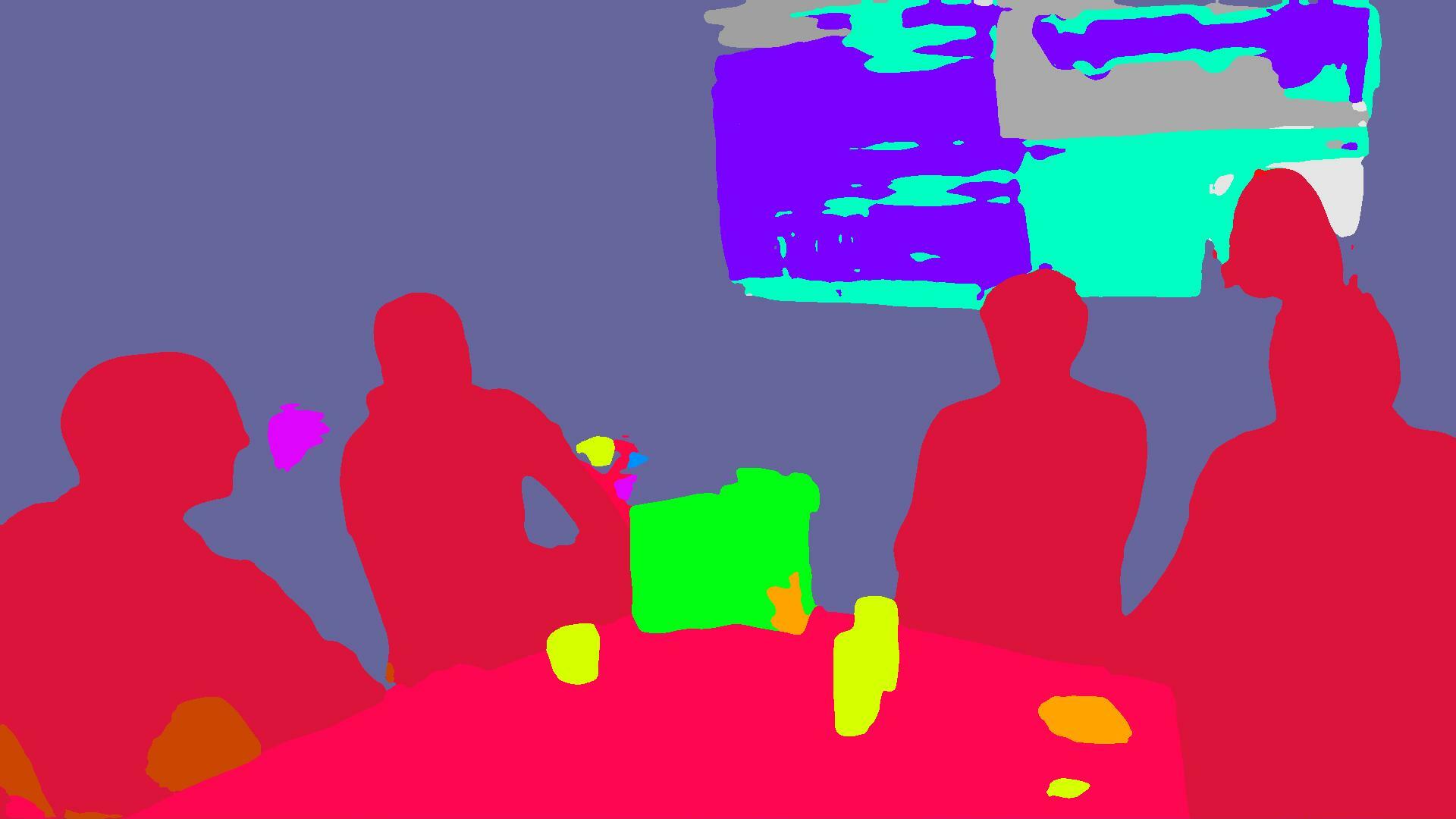}
    \includegraphics[width=0.3\textwidth]{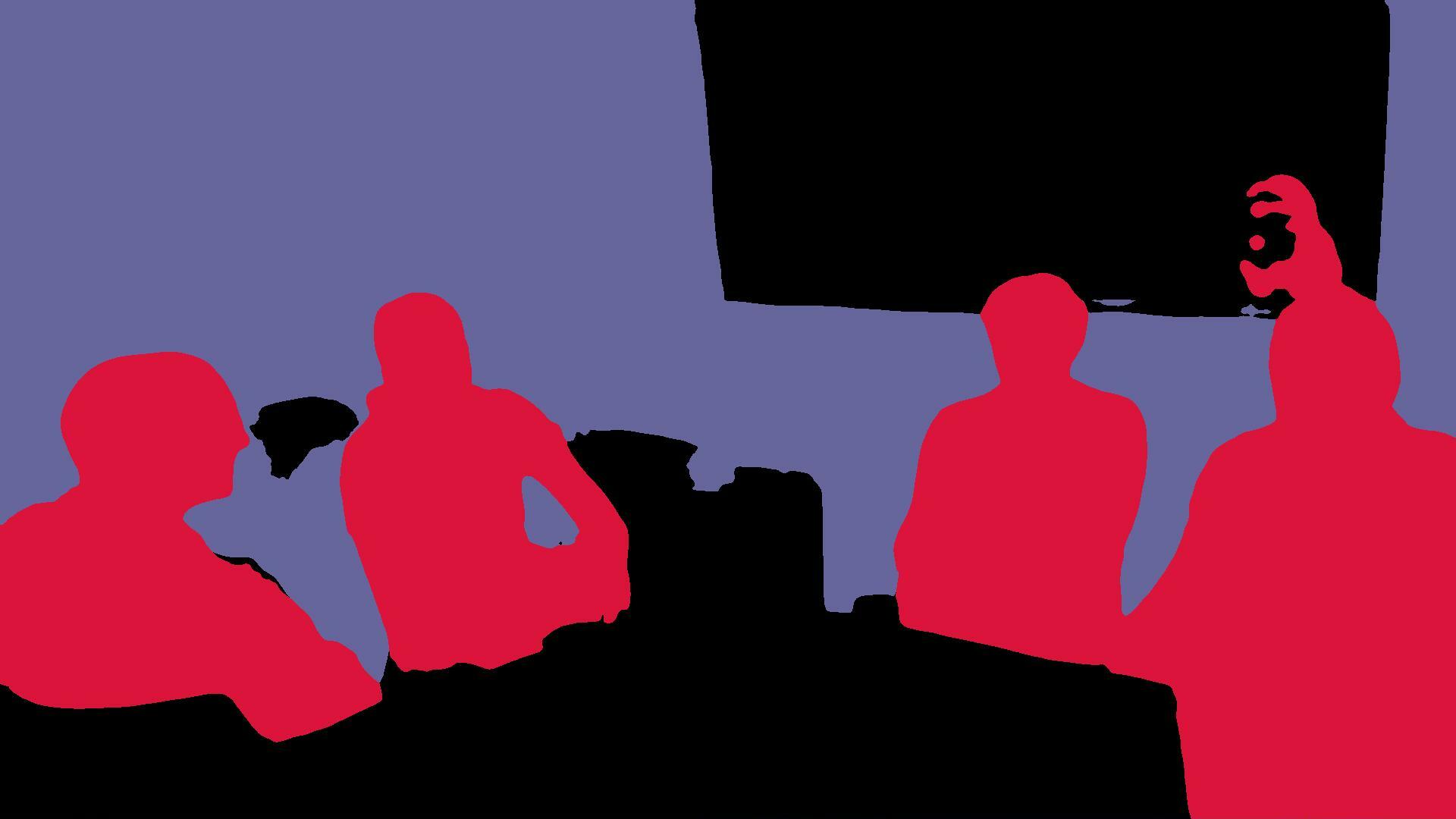}
    \\
    \includegraphics[width=0.3\textwidth]{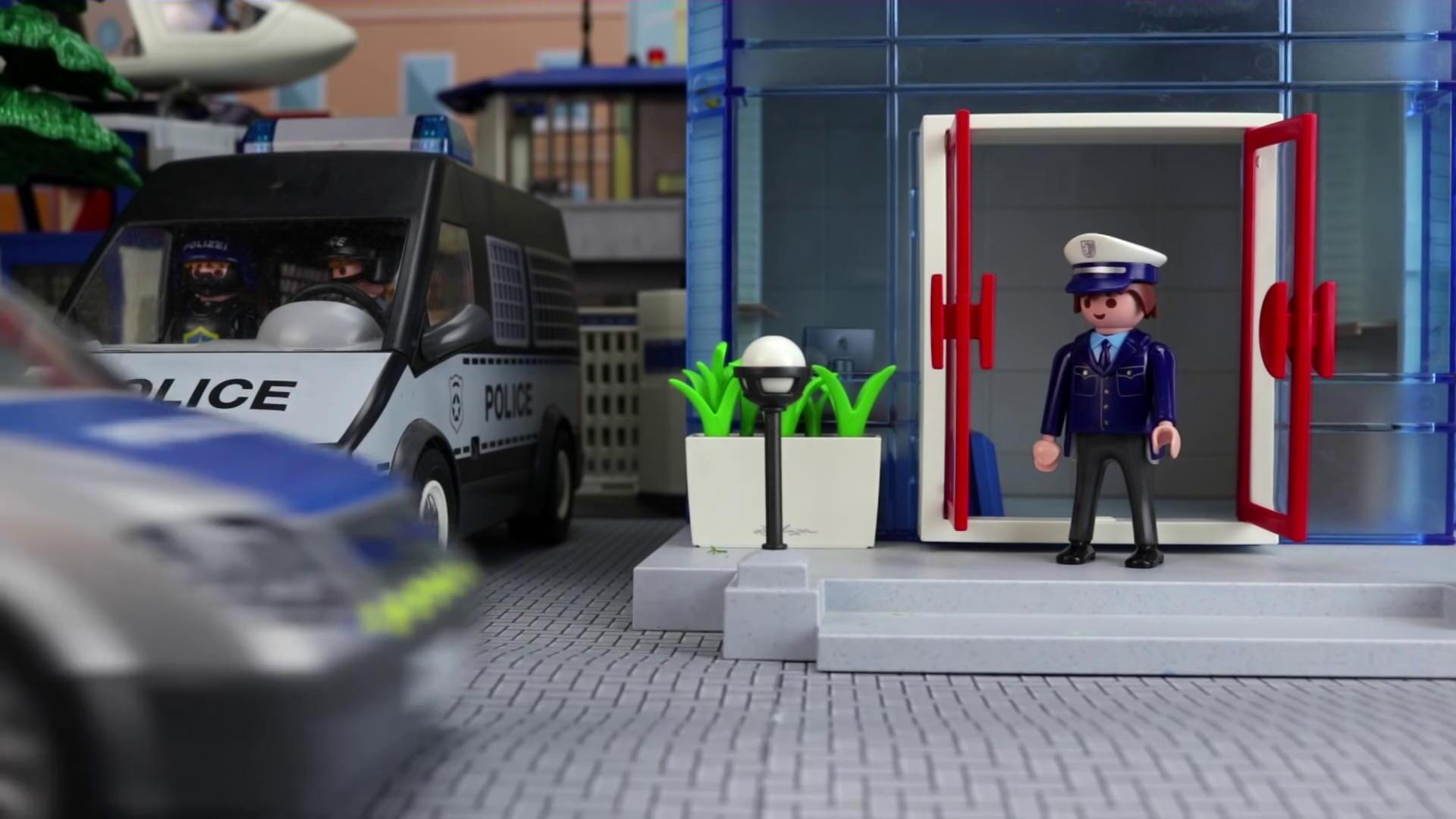}
    \includegraphics[width=0.3\textwidth]{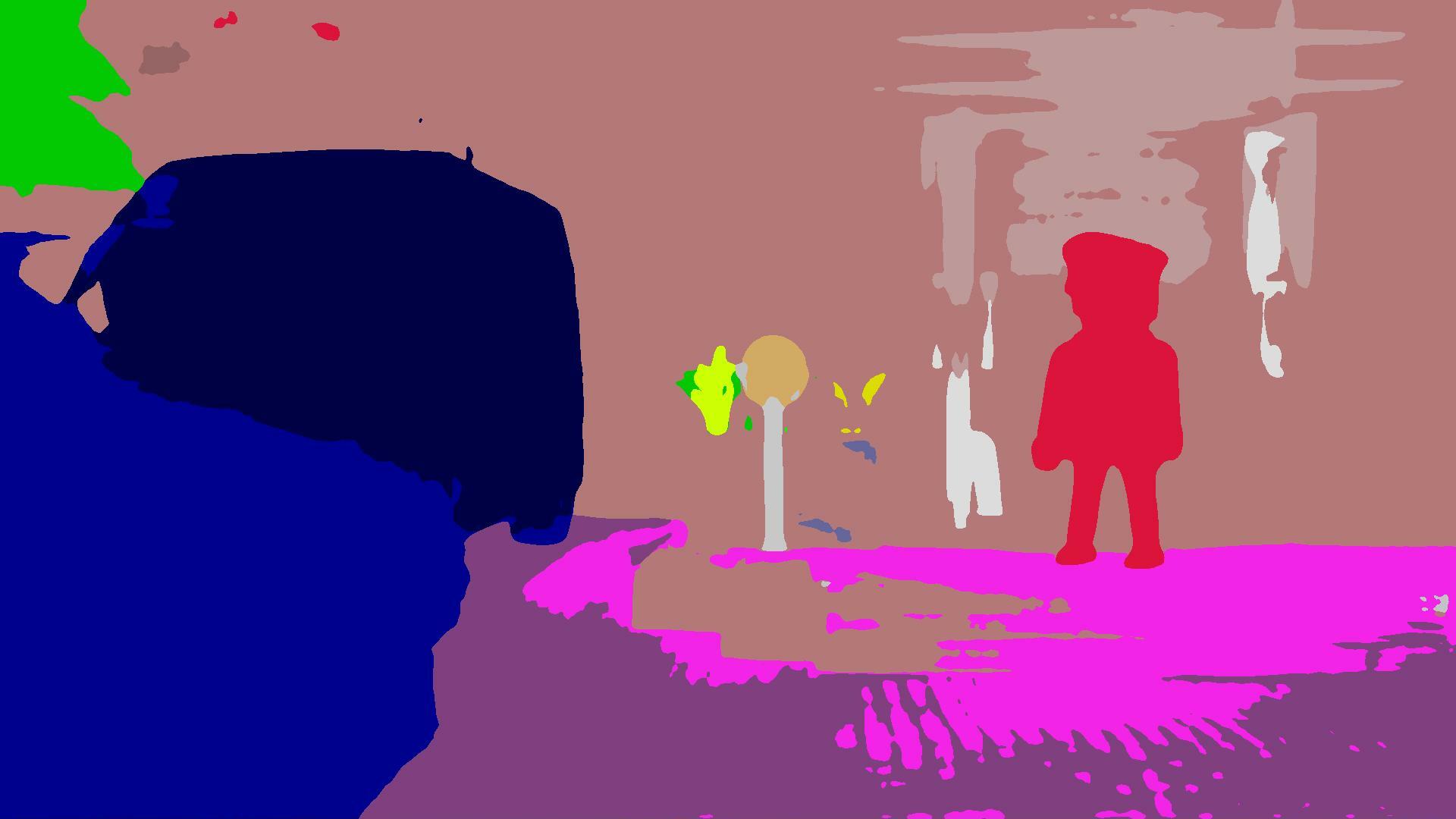}
    \includegraphics[width=0.3\textwidth]{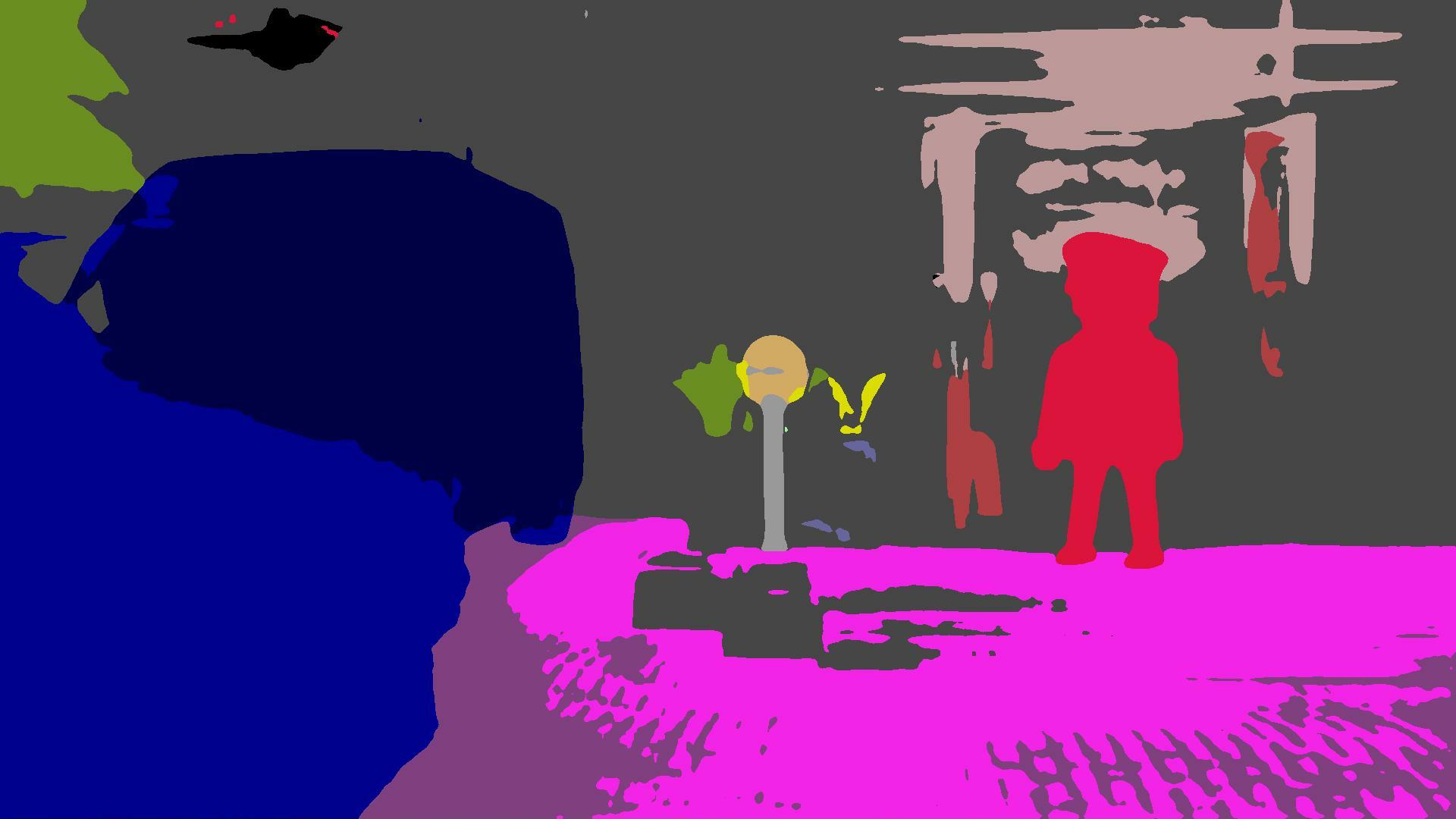}
    \\
    \includegraphics[width=0.3\textwidth]{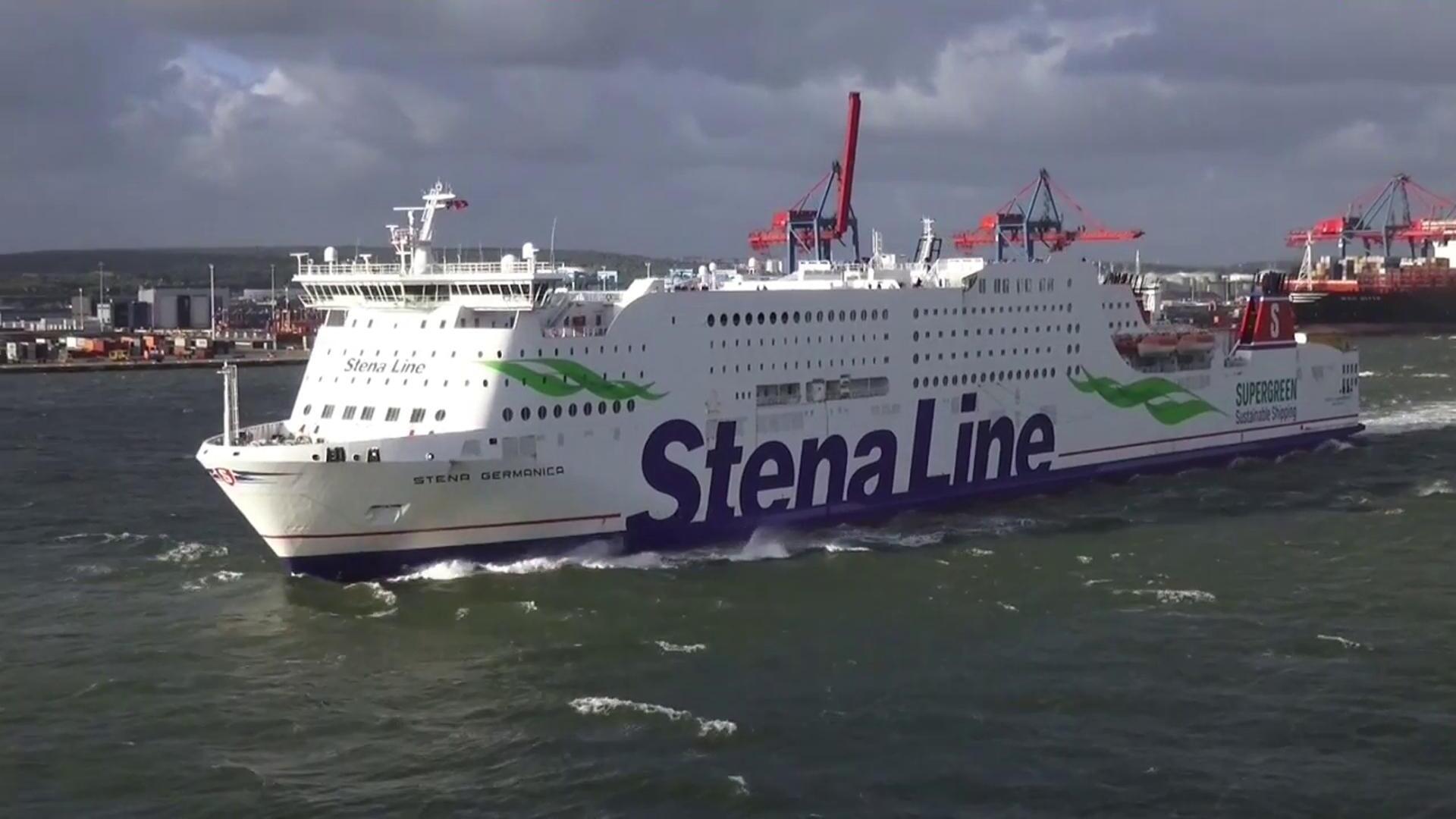}
    \includegraphics[width=0.3\textwidth]{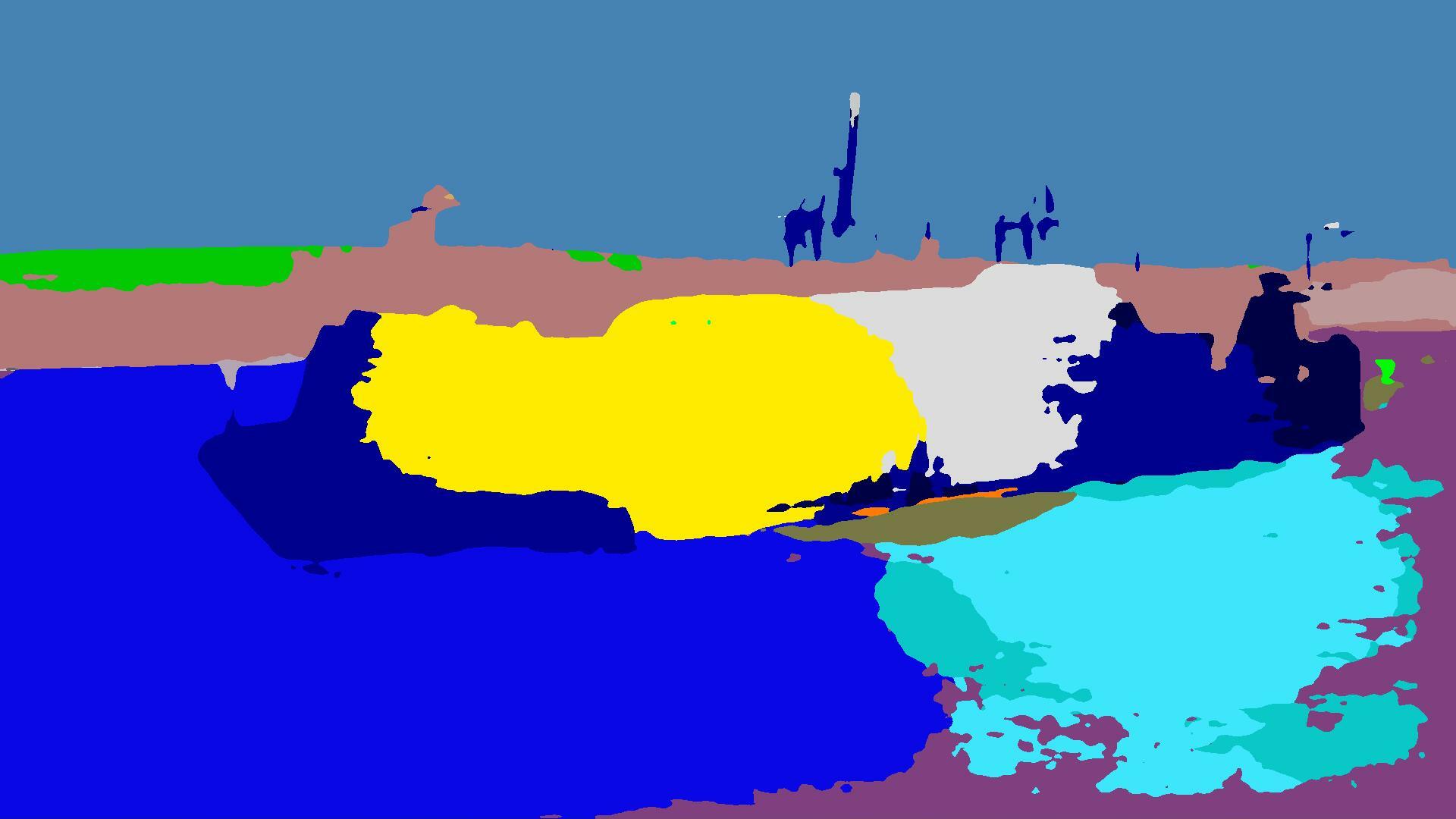}
    \includegraphics[width=0.3\textwidth]{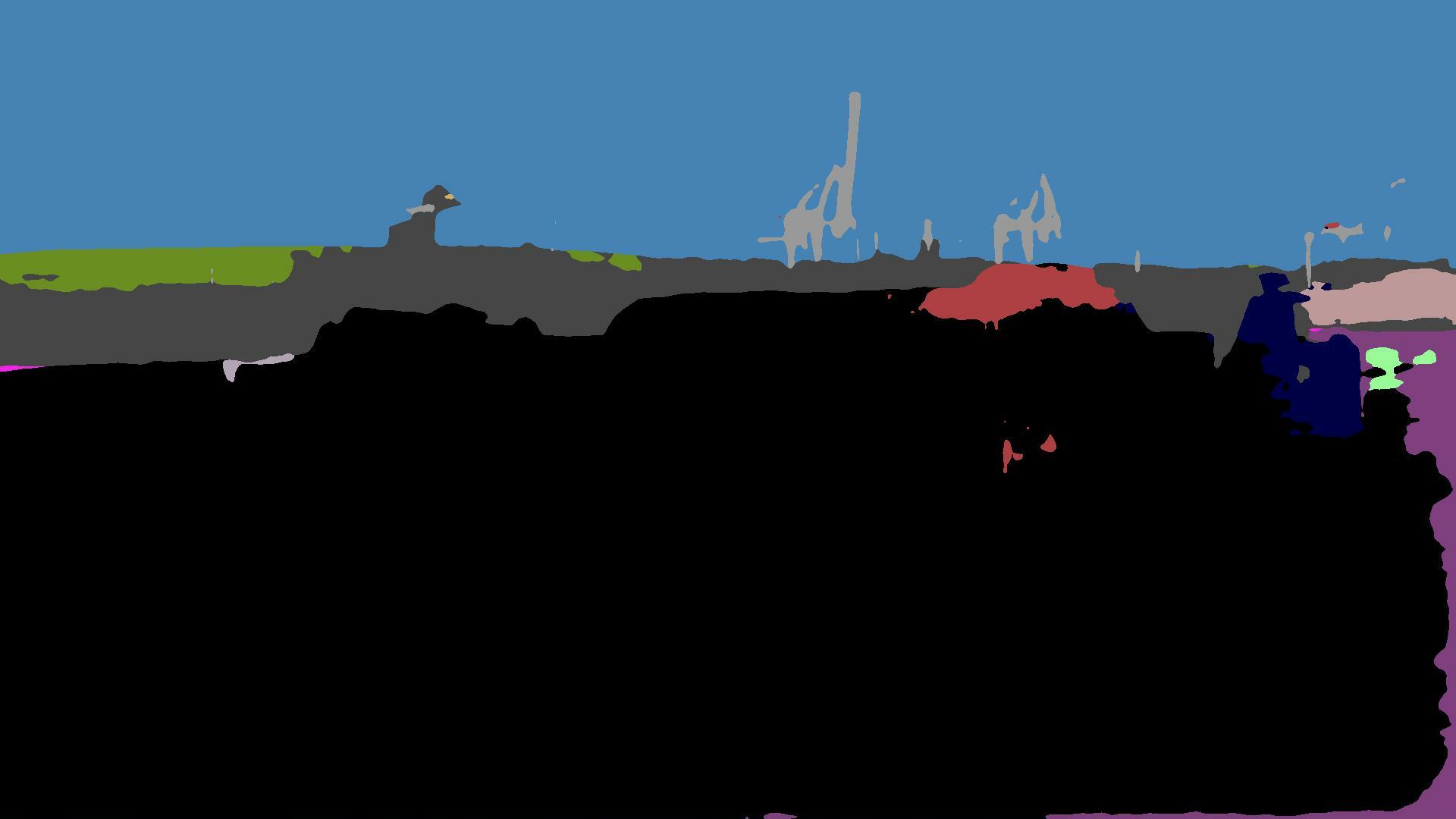}
    \\
    \includegraphics[width=0.3\textwidth]{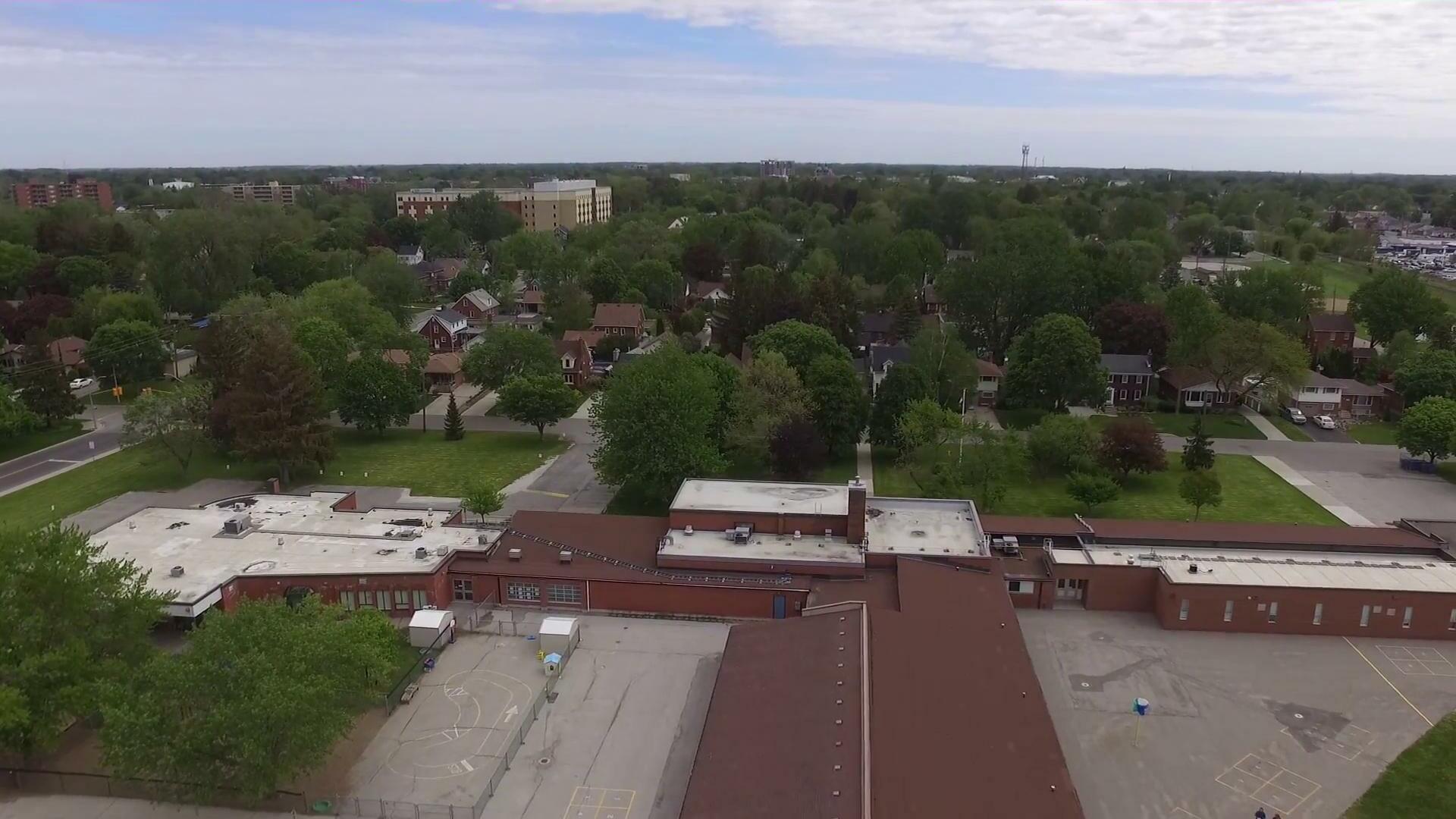}
    \includegraphics[width=0.3\textwidth]{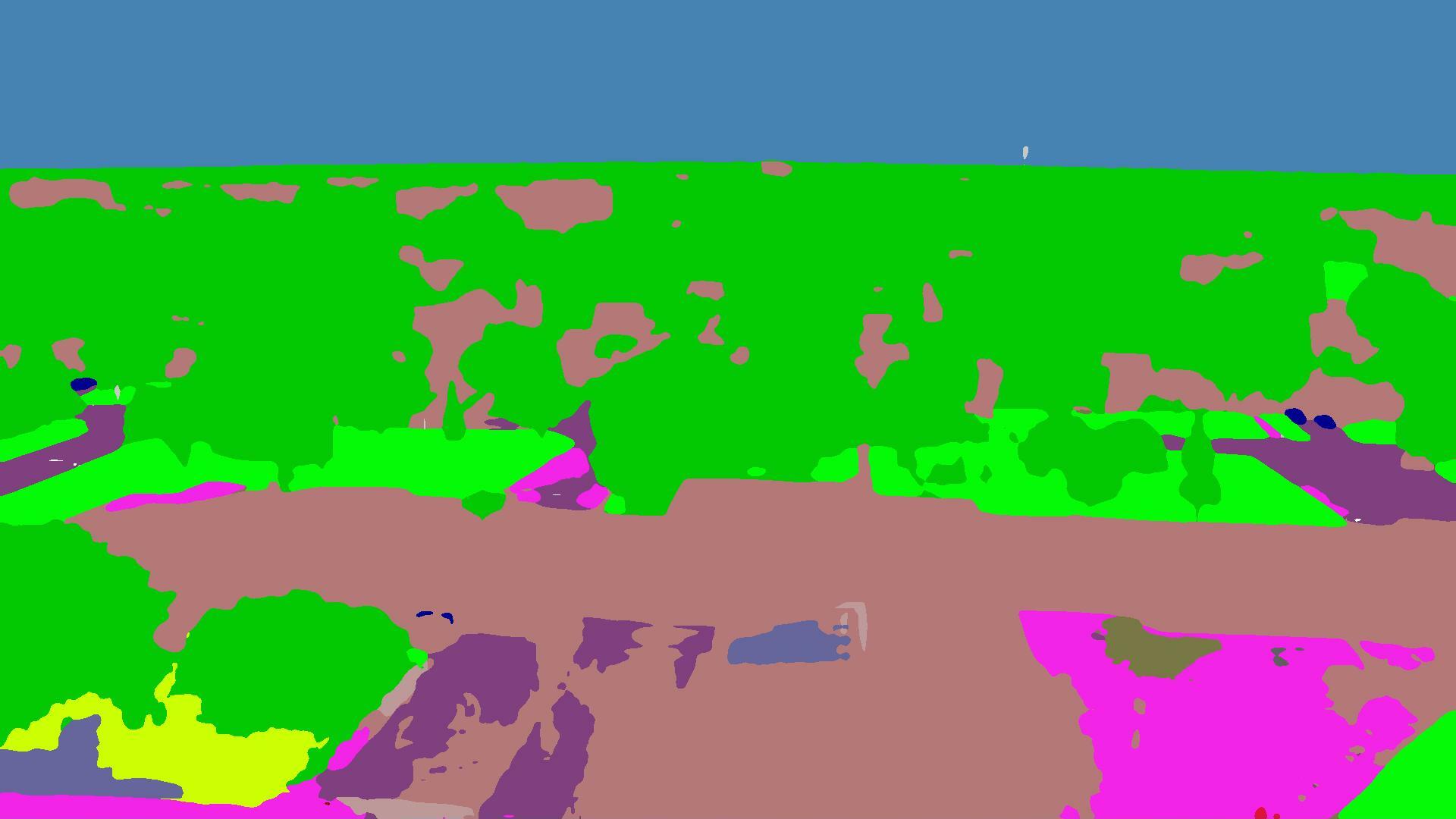}
    \includegraphics[width=0.3\textwidth]{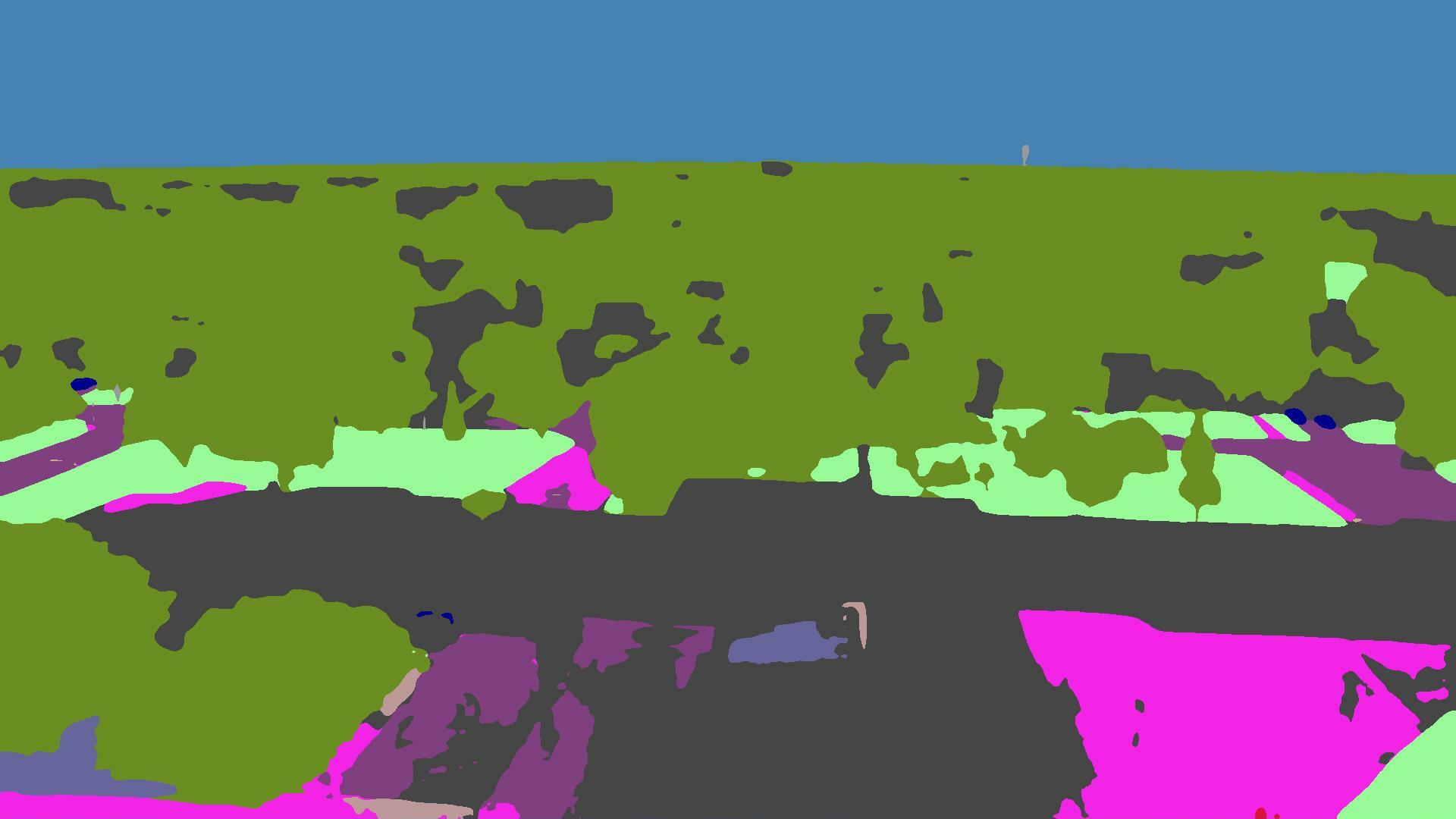}
  \caption{Performance 
    of our universal model 
    UNIV\_CNP\_RVC\_UE
    on four negative test images
    from WildDash 2.
    The columns show the input, 
    universal prediction, 
    and prediction in 
    the WildDash label space
    where class \mycls{void} is shown in black.
    Our model successfully recognizes 
    some non-road-driving classes,
    e.g.\ \mycls{table}, \mycls{chair}, \mycls{book}, \mycls{glass} 
    and \mycls{cabinet} (row 2), 
    or \mycls{boat} and \mycls{water} (row 4). 
    The model exhibits fair performance in 
    presence of large domain-shift (row 3)
    and robustness to changed perspective  (row 5).
  }
    \label{fig:wd-negative}
\end{figure*}

The winners of the RVC 2022 
semantic segmentation track
\cite{xiao22arxiv}
use a taxonomy obtained
by the partial merge approach.
Our experiments are not directly 
comparable with that work 
since their computing budget
is 10 times larger than ours
(64 GPUs vs 6 GPUs).
This makes a large difference 
in speed of training as well as
in available model capacity
since they can train 
on batches of 1 crop per GPU.
The runners-up of the RVC 2022 
semantic segmentation track 
\cite{liu22arxiv} 
construct their universal taxonomy
according to an early account
of our procedure from the section 
\ref{ss:method-taxonomy} \cite{orsic20arxiv}.
Their method showcases 
a great generalization performance
in spite of being trained 
on modest hardware (4 V100 GPU).
In comparison with our RVC 2020 model,
our submission to RVC 2022 featured 
a more ambitious training setup 
(we included BDD and COCO 
 to the training collection)
and a stronger backbone (ConvNeXt Large).
Yet, the performance was 
only slightly better than at RVC 2020.
Post-challenge validation has shown 
that ladder-style upsampling \cite{orsic20pr}
is not an effective design for large taxonomies.


\subsection{WildDash 2}
\label{ss:wd-experiments}

Multi-domain training discourages 
overfitting to dataset bias
and anticipates occurrence
of outliers during inference.
This makes it a prominent approach
towards robust performance 
in the real world.
We explore this idea in more depth
by analyzing the performance of our RVC model 
on the WildDash 2 benchmark.

The WildDash dataset collects 
challenging road-driving imagery
from across the whole world.
This sounds similar to Vistas
but there is one important difference.
Instead of aiming at a typical distribution
of world-wide road-driving scenes,
WildDash aims at edge cases
that are likely to break 
image uderstanding algorithms.
Just as Vistas is a step up from Cityscapes,
so WildDash further raises the bar
in several important aspects.

Wilddash is the first dataset 
to specifically target
expected points of failure 
of dense prediction models.
It explicitly enumerates visual hazards
such as underexposure, 
motion blur or particles 
\cite{zendel18eccv},
and quantifies their impact
to the prediction quality.
Furthermore, it includes
a negative test dataset 
from various non-driving contexts
that contains semantic anomalies 
with respect to typical 
road-driving taxonomies.
In all pixels of negative test images,
the model is allowed to predict
either a void prediction 
or a best-case ground truth.
Thus, WildDash could be seen 
as a precursor to recent dataset
for open-set segmentation
and dense anomaly detection
\cite{chan21neurips,blum21ijcv}.

Table \ref{table:wd_bench_results}
presents the actual
WildDash2 leaderboard.
Our RVC 2022 model achieves the best classic
segmentation score and a slightly better
robustness to common hazards.
We achieve a slightly lower overall score
due to worse performance on negative images. 
Available qualitative results 
\cite{rvc22www}
suggest that we find more positive content
in negative images than 
the competing approaches.

Figure \ref{fig:wd-negative} 
shows qualitative performance
of our model on negative images.
Columns show (left to right):
the input image, the prediction
in the universal taxonomy,
and the prediction 
in the WildDash 2 taxonomy. 
We designate void pixels 
with black colour.
The negative images were taken 
in non-road driving 
contexts (rows 1-4) or 
from an unusal perspective (row 5).
These images may contain pixels 
which conform to WildDash semantics  
such as people or walls (row 2).
Correct predictions of these pixels
are counted as true positives.

\subsection{Discussion}
This work, proposes manual universal taxonomies 
for encoding visual relations between 
classes in different datasets. 
Our approach encourages knowledge exchange
between datasets during either training or inference.
The proposed NLL+ loss includes this knowledge 
into standard multi-class training, 
and has the ability to transfer information 
down the label hierarchy.
We note that our framework is applicable
to different kinds of architectures.
For example, prototype-based approaches
could have prototypes for universal classes,
which could be associated 
with their dataset-specific counterparts.
We demonstrate the flexibility 
of our universal taxonomies
by applying them for multi-class training
of the recent M2F architecture.

Our concurrent work \cite{bevandic22bmvc} 
shows that it is possible to automate 
universal taxonomy construction. 
Still, we consider manual universal taxonomies 
as a useful baseline for measuring progress
of the automatic methods. 
Furthermore, manual universal taxonomies present 
an opportunity to evaluate other computer vision tasks. 
For example, a concurrent work 
on automatic discovery 
of visual relations between datasets
\cite{uijlings22eccv} 
performs experiments on the MSeg 
taxonomy that omits 61 classes 
and thus cannot properly encode 
all possible class relations. 
Furthermore, the Mseg taxonomy is limited 
to the original seven datasets 
and inflexible to further extensions
as it requires pixel-level relabeling.
Manual construction of universal taxonomies 
does require human effort 
but requires much less time than
pixel-level relabeling. 
Our universal taxonomy for the 10 
popular semantic segmentation datasets
has been constructed 
by the first author of this manuscript 
during less than two days.

We note that multi-dataset training
comes with additional challenges 
(e.g.\ domain shift)
that are not connected
to incompatible labelling policies. 
Multi-dataset training is likely
to increase resilience to these issues
even though this may not be evident 
when considering performance 
on particular datasets that reward overfitting.

\section{Conclusion}

This paper introduces a novel method
for training semantic segmentation
on a collection of datasets
with overlapping classes.
We express dataset-specific labels
as sets of disjoint universal classes
that correspond to distinct visual concepts.
Thus, the standard dataset-specific loss
can be formulated as negative logarithm
of aggregated universal posteriors
which we succinctly denote as NLL+.
We showcase the flexibility of our approach 
by implementing it alongside a recent
mask-level dense prediction model,
where we apply NLL-max loss over 
universal pixel-assignment masks.
In both cases, our approach succeeds to
learn universal classes from
the original ground-truth
in spite of incompatible taxonomies.

Practical implementations of our method
require construction 
of a flat universal taxonomy
that spans the desired dataset collection.
This calls for recovering
a set of disjoint universal classes
as well as mapping each dataset-specific class
to the corresponding universal counterparts.
We propose to solve this problem
by considering labels
as sets of all possible pixels
that should be annotated
with the particular label.
We hope to encourage 
future research in the field
by publishing  the source code \cite{bevandic22github}
for all universal
taxonomies from this paper.

Our experiments consider several baselines
that map semantically related labels
to distinct strongly supervised logits.
We show that their performance improves
if we consider class relationships
during post-inference processing.
However, our method outperforms
the baseline performance both in
within-dataset and cross-dataset contexts.
This suggests that it pays off to resolve
semantic incompatibilities before the training.
We observe the largest advantage
while training on non-biased datasets
such as WildDash and Vistas,
where implicit dataset detection
becomes increasingly difficult
and requires considerable capacity.
We even show that there are instances
where our method can go beyond
the semantics of individual datasets
by learning a novel concept
that does not exist as a distinct class
in any of the input taxonomies.

We also compare our approach with
the related MSeg taxonomy
that supports standard learning
with strong supervision
on relabeled images.
Recall that MSeg taxonomy
does not span all classes
from the seven involved taxonomies,
since it drops 61 fine-grained classes
in order to contain the relabeling effort.
Empirical comparison on 
the seven validation splits
reveals performance advantage
of our universal models.
Our models remain competitive
even when the evaluation
considers only the 194 semantic classes 
that have been kept or relabeled
in the MSeg taxonomy.
This suggests that learning on more classes
may compensate for weak supervision.
Our approach is much more versatile than MSeg
since it can be applied to new problems
without any relabeling.

To conclude, our work shows
that multi-dataset training
profits from resolving
semantic relationships
between individual taxonomies.
Our method delivers versatile and robust models
that can afford large-scale training
on collections of heterogeneous taxonomies.
Future work could aim towards
advanced evaluation datasets,
automatic recovery of universal taxonomies,
and applying the proposed framework
for open-set recognition.

\backmatter

\bmhead{Acknowledgments}

This work 
has been supported by 
Croatian Science Foundation
grant IP-2020-02-5851 ADEPT,
by NVIDIA Academic Hardware Grant Program,
by European Regional Development Fund
grant KK.01.1.1.01.0009 DATACROSS
and by VSITE College for 
Information Technologies
who provided access to 6 GPU Tesla-V100 32GB.

\bmhead{Data Availability Statement}
We perform our experiments on the following
publically available datasets:
ADE20k \cite{zhou17cvpr},
BDD \cite{yu18bdd}, 
Camvid \cite{badrinarayanan17pami}, 
Cityscapes \cite{cordts16cvpr}, 
COCO \cite{lin14eccv}, 
IDD \cite{varma19wacv},
KITTI \cite{geiger13ijrr},
MSeg \cite{lambert20cvpr},
SUN RGBD \cite{song15cvpr},
Scannet \cite{dai17cvpr}, 
Viper \cite{richter17iccv},
Vistas  \cite{neuhold17iccv}, and
WildDash 2 \cite{zendel18eccv}.

Our universal taxonomy for these
datasets is available online \cite{bevandic22github}.

\bmhead{Published Paper}
This manuscript has been accepted for publication
International Journal of Computer Vision,
after peer review and is subject to their terms of use,
but is not the Version of Record and does not 
reflect post-acceptance improvements, or any corrections.
The Version of Record is available here:
https://www.springerprofessional.de/international-journal-of-computer-vision/11065872.

\begin{appendices}

\section{Gradients for NLL+ loss}\label{secA1}
\begin{align*}
  &\frac{\partial \mathcal{L}}
      {\partial s_c} = \frac{\partial}
      {\partial s_c} ( - \ln \sum_{\cls u' \in m_{\set S_d}(\cls y)}\exp{s_{u'}}
      + \ln \sum_{u} \exp{s_{u}})\\
  &=
  -\enbbracket{\cls c \in m_{\set S_d}(\cls y)}
  \frac{\exp{s_c}}{\sum_{\cls u' \in m_{\set S_d}(\cls y)}\exp{s_{u'}}}
  + \frac{\exp{s_c}}{\sum_{\cls u}\exp{s_{u}}}\\
  &=
    -
  \frac{\enbbracket{\cls c \in m_{\set S_d}(\cls y)}\frac{\exp{s_c}}{\sum_{\cls u}\exp{s_{u}}}}
  {\sum_{\cls u' \in m_{\set S_d(\cls y)}}\frac{\exp{s_{u'}}}{\sum_{\cls u}\exp{s_{u}}}}
  + \frac{\exp{s_c}}{\sum_{\cls u}\exp{s_{u}}}\\
  &=
  - \frac{\P(\rvar Y = \cls y \mid \rvar U =  \cls c )    \P(\rvar U \cls = c\mid \vec x)}{ \sum_{\cls u' \in m_{\set S_d}(\cls y)}\P(\rvar U
      = \cls u'\mid \vec x)}
    +
    P(\rvar U = \cls c\mid\vec x)\\
    &=
  - \frac{\P(\rvar Y = \cls y, \rvar U \cls = c \mid \vec x)}{ \P(\rvar Y = \cls y\mid \vec x)}
    +
    P(\rvar U = \cls c\mid\vec x)\\
  &=
  - P(\rvar U = \cls c\mid\rvar Y = \cls y, \vec x)
    +
    P(\rvar U = \cls c\mid\vec x)\;.
  \label{eq:partial-derivative}
\end{align*}




\end{appendices}
\bibliographystyle{sn-vancouver}
\bibliography{mybib}


\end{document}